\documentclass{article}
\usepackage[a4paper, total={6in, 8in}]{geometry}
\usepackage[utf8]{inputenc}
\usepackage{amsmath}
\usepackage{amsfonts}
\usepackage{amssymb}
\usepackage{authblk}
\usepackage{bbm}
\usepackage{booktabs}
\usepackage{csquotes}
\usepackage{float}
\usepackage{graphicx}
\usepackage{hyperref}
\usepackage{longtable}
\usepackage{multirow}
\usepackage{natbib}
\usepackage{placeins}
\usepackage{siunitx}
\usepackage{svg}
\usepackage{tcolorbox}
\usepackage{xcolor}

\usepackage{fancyhdr}
\usepackage[skip=10pt plus1pt, indent=20pt]{parskip}

\hypersetup{
  colorlinks,
  citecolor=violet,
  linkcolor=red,
  urlcolor=blue}

\usepackage{tikz}
\usetikzlibrary{arrows,arrows.meta,decorations.shapes,decorations.pathreplacing,fit,positioning,trees,quotes}

\graphicspath{ {./images/} }

\def\Xset{\mathbb{X}}
\def\Zset{\mathbb{Z}}
\def\vcld{v_{c,l}\cdot}
\def\vcl{v_{c,l}}

\def\xp{\Xset^+_{\leq T}}
\def\xm{\Xset^-_{\leq T}}
\def\xmj{\Xset^-_{\leq T,j}}
\def\zp{\Zset^+_{\leq T,l}}

\def\zmj{\Zset^-_{\leq T,l,j}}

\def\st{\textrm{such that}}

\newcommand{\ct}[1]{\texttt{#1}}
 
\title{Bridging the Human--AI Knowledge Gap: \\ Concept Discovery and Transfer in AlphaZero}

\author[1,*]{Lisa Schut}
\author[2]{Nenad Toma\v{s}ev}
\author[2]{Tom McGrath}
\author[2]{\\ Demis Hassabis}
\author[2]{Ulrich Paquet}
\author[2]{Been Kim}

\affil[1]{\footnotesize OATML, Dept. of Computer Science, University of Oxford}
\affil[2]{\footnotesize Google DeepMind}
\affil[*]{\footnotesize Work done at Google DeepMind}

\begin{document}
\maketitle

\begin{abstract}
Artificial Intelligence (AI) systems have made remarkable progress, attaining super-human performance across various domains. 
This presents us with an opportunity to further human knowledge and improve human expert performance by leveraging the hidden knowledge encoded within these highly performant AI systems. Yet, this knowledge is often hard to extract, and may be hard to understand or learn from. Here, we show that this is possible by proposing a new method that allows us to extract new chess concepts in AlphaZero, an AI system that mastered the game of chess via self-play without human supervision. Our analysis indicates that AlphaZero may encode knowledge that extends beyond the existing human knowledge, but knowledge that is ultimately not beyond human grasp, and can be successfully learned from. In a human study, we show that these concepts are learnable by top human experts, as four top chess grandmasters show improvements in solving the presented concept prototype positions. 
This marks an important first milestone in advancing the frontier of human knowledge by leveraging AI; a development that could bear profound implications and help us shape how we interact with AI systems across many AI applications. 
\end{abstract}

\section{Introduction}
Artificial Intelligence (AI) systems are typically treated as problem-solving machines; they can carry out the jobs humans are already capable of but more efficiently or with less effort, which brings clear benefits in several domains. 
In this paper, we pursue a different goal: treat AI systems as learning machines and \textit{demand} from them to teach us the fundamental principles behind their decisions to
extend upon and complement our knowledge. We can imagine many benefits of learning from machines. For example, while a system capable of producing a more accurate cancer diagnosis or effective personalised treatment than human experts is useful, 
transferring the rationale behind their decisions to human doctors could not only bring advances in medicine but also leverage human doctors' strength and generalisation ability to enable new breakthroughs.
There is a tremendous untapped opportunity across various domains where the capabilities of AI systems are reaching or exceeding those of human experts (super-human AI systems).
This work is one of the very first steps towards the development of tools and methods that allow us to uncover hidden knowledge in highly capable AI systems, and empower human experts by helping them further improve their skills and understanding. 

The super-human ability of AI systems may arise in a few different ways: pure computational power of machines, a new way of reasoning over existing knowledge, or super-human knowledge we do not yet possess. This work focuses on the last two cases. For simplicity, we refer to both as super-human knowledge from now on.  

\begin{figure}[ht] 
\centering
\caption{Learning from machine-unique knowledge.}
\includegraphics[width=0.8\textwidth]{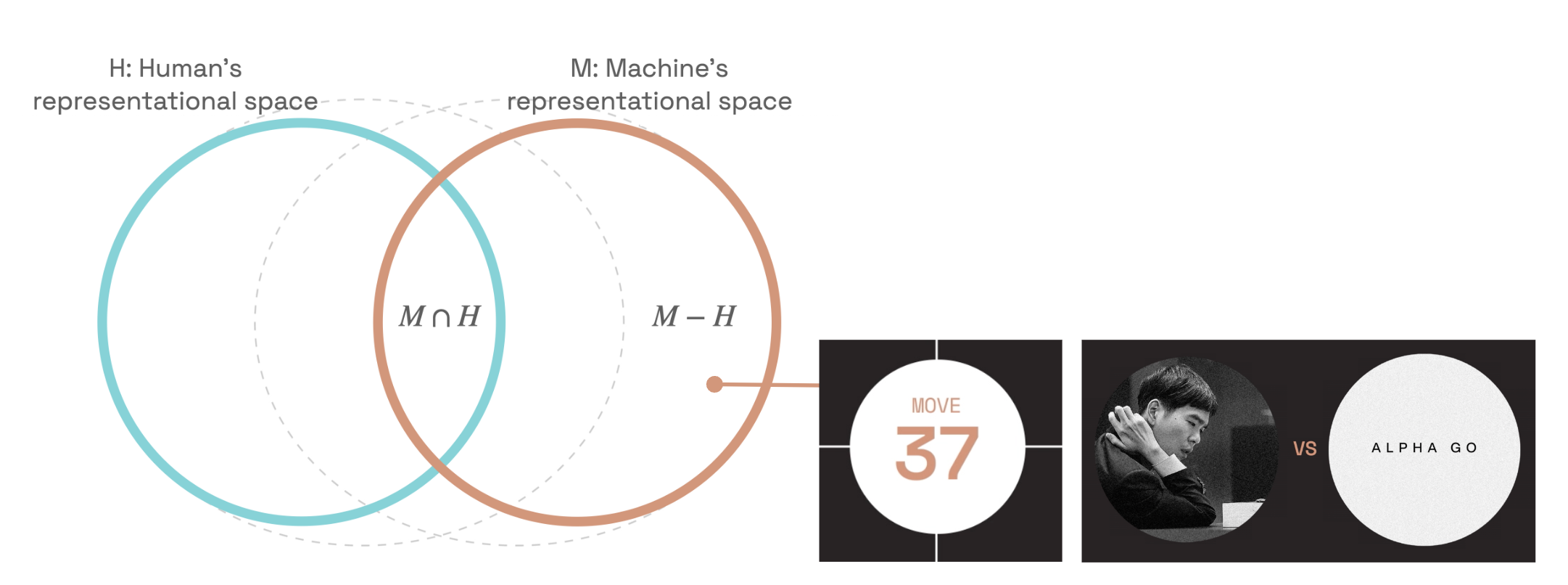}
\label{fig:m_minus_h}
\end{figure}

What does this mean from a research standpoint? The human representational space ($H$) has some overlap with the machine representational space ($M$) (see Figure~\ref{fig:m_minus_h}~\citep{iclrkeynote_been_2022}). A \textit{representational space } forms the basis of and gives rise to knowledge and abilities, which we are ultimately interested in. Thus, we use representational space and knowledge interchangeably -- roughly speaking, $H$ to represent what humans know and $M$ to represent what a machine knows. 
There are things that both AI and humans know ($M \cap H$), things that only humans know ($H - M$), and things only machines know ($M - H$). Most existing research efforts only focus on ($M \cap H$), e.g., interpretability has tried to shoehorn $M$ into ($M \cap H$), with limited success~\citep{adebayo2018sanity, nie2018,  bilodeau22impossibility}. We believe that the knowledge gap represented by $(M - H)$ holds the crucial key to empowering humans by identifying new concepts and new connections between existing concepts within highly performant AI systems. We already have evidence of cases when certain AI generations captivated the human imagination with ideas that were initially hard to grasp. One prominent example in the history of AI is the move 37 that AlphaGo made in a match with Lee Sedol. This move came as a complete surprise to the commentators and the player, and is still discussed to this day as an example of machine-unique knowledge. 
The vision to pursue super-human knowledge is ultimately for human-centered AI, and a world where human agency and capability do not come second. 
However, the question is--is this even possible?  

This work is the first step towards discovering super-human knowledge and new connections of existing knowledge in $(M - H)$. 
We focus on a domain that has inspired AI practitioners for decades, and captivated human imagination for centuries: the game of chess. Chess is an excellent playground to validate the existence and usefulness of set $(M-H)$ for many reasons: chess knowledge has been developed over a long period of time, and the ground truth is much easier to validate compared to the frontiers of other fields, such as science or medicine. We also have a quantitative measure of the quality of play, both for human experts as well as machines, known as the Elo rating~\citep{wiki:elo}. 

Chess engines have performed at a super-human level for a long time, ever since DeepBlue's match against Garry Kasparov. While early engines were based on human knowledge, the advent of AlphaZero~\citep{silver2017mastering} (AZ) showed a self-taught deep learning model achieve a super-human capability in chess without any human knowledge. 
However, as humans, we have not yet been able to tap into their knowledge fully. Through analysis of AZ's games, humans manually distilled patterns, such as its proclivity for playing on the flanks with moves like a4 or h4~\citep{game_changer}. However, this still analyses $M$ through the lens of $H$, a bias that limits what we can find from $M \cap H$. 

In this work, we aim to take the first step to change that by facilitating learning from the super-human knowledge in the
$(M - H)$ set of AZ. We hypothesise that $(M - H)$ exists, and can be taught to humans. 

We validate our hypothesis by showing that we can teach new chess concepts to four top human grandmasters, the best chess players in the world. Also, due to their undeniable strength, and talent, $(M - H)$ may fall into their `proximal zone of development' in Vygotsky's education theory: ``the space between what a learner can do without assistance and what a learner can do with adult guidance or in collaboration with more capable peers''. While communicating $(M - H)$ may require new language~\citep{iclrkeynote_been_2022}, we bypass this need in this work by leveraging chess champions' ability to connect the dots and generalise from patterns that arise in chess positions. 

\begin{figure}[!ht]
\caption{Example of a concept prototype. Most chess players would opt for \ct{Rxh5}, however, AZ plays \ct{Qc1}, with the idea of regrouping the pieces to the queenside. Further details can be found in \S\ref{appx:proto_human}.}
\centering
\includegraphics[width=0.45\textwidth]{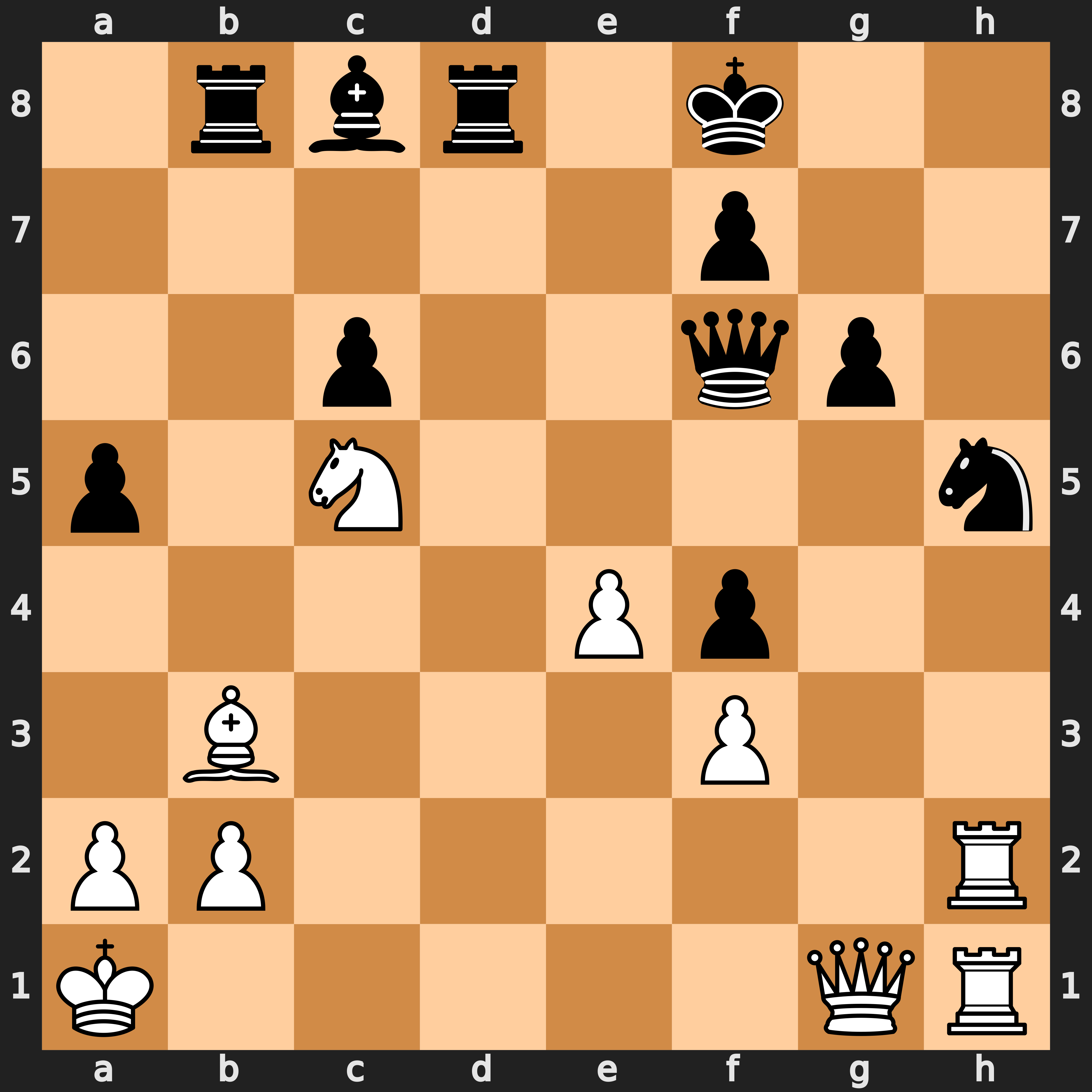} \\
\vspace{0.2cm}
\fbox{\begin{minipage}{\textwidth}
\begin{flushleft}
\textbf{Part of AZ's MCTS calculations:}
\ct{37.Qc1 Kg7 (37...Rb5 38.a4 Rb4 39.Ka2; 37...Qe5 38.Qc4 Be6 39.Nxe6+ fxe6 40.Qxc6) 38.Re1 Qe5 39.Rc2 Rb4 40.Ba4 Qd6 41.a3 Rd4 42.e5 Qd5 43.Bxc6 Qxc6 44.Nb3 White is better}
\end{flushleft}
\end{minipage}}
\label{fig:qc1_intro}
\end{figure}

We find evidence that suggests $(M-H)$ exists through analysing the dimension of the span of the latent representations of AZ's and human's games (\S\ref{sec:novelty}). 
Next, we develop a new framework to search for concepts in $(M-H)$, i.e., unearth AZ's super-human knowledge.
In our framework, we: 
\begin{itemize}
    \item develop a \textbf{new method for finding unsupervised concepts} in the latent space.
    By using the full AZ machinery, both the policy value network and MCTS tree,
    our method discovers \textit{dynamic} concepts 
    that motivate a sequence of actions in chess. We show that our method can find vector representations of concepts in a data-efficient manner (\S\ref{sec:convex_opt}).
    \item  ensure concepts are novel.
    Through spectral analysis, our framework only select concepts that contain information unique to the vector space of AZ's games compared to that of human games.\footnote{This difference is not due to variance, as the human games have a larger span in input space.}
    \item ensure that concepts are teachable. We develop a new metric that evaluates whether concepts are \textbf{teachable} to another AI agent with no prior knowledge of the concept (\S\ref{sec:teachability}). Through this metric, we select concepts based on their informativeness (i.e., useful for the AI agent in a downstream task).  
    \item provide insight into the meaning of the new concepts via graph analysis to reveal new concepts' relations to human-labelled concepts.
\end{itemize}
After finding these concepts in $(M-H)$, we analyse whether we can expand the human representational space $(H)$ to include these new concepts (\S\ref{sec:human}). We collaborate with four world top chess grandmasters and former World Champions to test whether they could learn and apply the concepts, by learning from prototypical examples. Figure~\ref{fig:qc1_intro} shows an example of a concept prototype. Here, most chess players would continue to play on the kingside with \ct{Rxh5}. However, AZ finds the only plan to maintain an advantage: \ct{Qc1} with the idea of re-manoeuvring the pieces to the queenside. 

The results of our study show an improvement in the grandmasters' ability to find concept-based moves aligned with AZ's choices, as compared to their performance prior to observing AZ's moves. Further, their qualitative feedback indicated an understanding and appreciation of AZ's plans. 
The discovered concepts often combine and apply chess concepts in a way that deviates from the traditional human principles of chess. We conjecture that the differences in humans' and AZ's play may stem from their differences in how position-concept relationships are built. 
While humans develop prior biases over which concepts may be relevant in given chess positions, AZ has developed its own unconstrained understanding of concepts and chess positions, enabling flexibility and creativity in its strategies. 

Our paper is structured as follows. First, we summarise related work in \S\ref{sec:related_work}. Next, we discuss the definition of concepts and how we operationalise it in \S\ref{sec:assumptions_defs}.
We introduce our method for finding concepts in \S\ref{sec:convex_opt} and filtering them in \S\ref{sec:filtering} to ensure concepts are informative, teachable, and novel.
We demonstrate the efficacy and performance of our method on supervised concepts in \S\ref{sec:method_eval}.
Finally, in \S\ref{sec:human}, we lay out the human experiment protocols and results and show how our framework can enable bridging the $(M-H)$ gap. We conclude the paper in \S\ref{sec:conclusion} by summarising our main findings, and discussing limitations and future work.

\section{Related work} \label{sec:related_work}

Here, we review relevant prior work on concept discovery, interpretability of reinforcement learning systems, and the intersection of AI and chess.

\subsection{Concept-based explanations}

In contrast to traditional feature or data-centric interpretability methods~\citep{ribeiro2016should,lundberg2017unified, integrated-gradients, koh2017understanding}, concept-based methods use high-level abstraction, concepts, with the goal of providing model explanations to inform human practitioners~\citep{bau2017network,kim2018interpretability,melis2018towards, koh2020concept, bai2022concept, achtibat2022where, NEURIPS2022_11a7f429}.
These types of explanations are shown to be useful in scientific and biomedical domains~\citep{graziani2018regression, sprague2018interpretable, clough2019global, bouchacourt2019educe, yeche2019ubs, sreedharan2020bridging, schwalbe2020concept, mincu2021concept, jia2022role}, where experts' concepts are highly relevant in decision making rather than individual low-level features.

More in line with the work presented in this paper, concept-based explanation methods have been studied in board game playing agents, including in Hex~\citep{forde2022concepts} and Go~\citep{tomlin2022understanding}. Establishing a causal link between concepts and predictions is non-trivial and a topic of ongoing research~\citep{goyal2019explaining, Bahadori2020, wu2023causal}.

Shortcomings of supervised concept-based methods have been studied. 
When leveraging a set of exemplars of a concept (a probe dataset), 
\cite{Ramaswamy_2023_CVPR} showed that different probe datasets can lead to inconsistent conclusions. Further, they showed that the number of concepts in probe datasets exceeds those used by humans. The linearity assumption has limitations \citep{chen2020concept, soni2020adversarial}, and the faithful alignment between the vector and humans' mental models of the concept was challenged in \cite{pitfalls21}. 

Going beyond supervised concepts and probe datasets has also been investigated~\citep{yeh2020completeness, ghorbani2019towards, ghandeharioun2021dissect} to discover concepts that a model represents without being limited to human labelled concepts. The concept is expressed using examples of training data~\citep{yeh2020completeness, ghorbani2019towards} or by generating new data~\citep{ghandeharioun2021dissect}. This work falls under methods to discover concepts but with a different goal of \textit{discovering} and \textit{teaching} humans new concepts rather than finding existing human concepts. 

\subsection{Generating explanations in Reinforcement Learning}
Generating explanations in Reinforcement Learning (RL) methods~\citep{alharin2020reinforcement, heuillet2020explainability, glanois2021survey, krajna2022explainability, vouros2022explainable, milani2022survey, dazeley2023explainable, omidshafiei2022beyond, das2023state2explanation} is of particular interest, as these methods are increasingly deployed in real-world applications, and the explanation requirements differ compared to the more traditional supervised learning setting. This is due to the temporal dependency between states, actions, and subsequent states, where an agent's historical, current, and future state-action sequences may relate to some long-term goal~\citep{dazeley2023explainable}. Explainability methods in RL can help identify issues with agents related to over-fitting the training data, poor out-of-distribution performance~\citep{annasamy2019towards} and inter-agent dynamics~\citep{omidshafiei2022beyond}.

Several efforts focused on designing more interpretable model architectures and training procedures in representation learning~\citep{raffin2019decoupling, raffin2018srl, 8852042, traore2019discorl, doncieux2020dream, openendedlearning} and symbolic and relational methods~\citep{sreedharan2020tldr, garnelo2016deep, garcez2018symbolic, zambaldi2018relational, hazra2023deep}, which may involve intermediate perceptual processing steps, like object recognition~\citep{goel2018unsupervised, li2018object}. 
Different RL methods (value-based, policy-based, model-based, fully or partially observable states)~\citep{alharin2020reinforcement} may lend themselves to different explainability approaches or variations thereof. 
Likewise, the explanations themselves may vary in scope, e.g., local explanations of individual agent actions or value assessments, or the overall high-level explanations of the agent policy~\citep{zrihem2016visualizing, sreedharan2020tldr, topin2021iterative}. 
The importance of treating explanations as a reward to ensure consistency is explored in ~\cite{yang2023leveraging}.

For trained RL systems, there is a pressing need for post-hoc RL interpretability methods. Input saliency maps ~\citep{wang2016dueling, Selvaraju_2019, greydanus2018visualizing, mundhenk2020efficient} and tree-based models~\citep{bastani2018verifiable, roth2019conservative, coppens2019distilling, liu2019toward, vasic2019moet, madumal2020distal} have been a common approach.
Saliency-based RL explainability approaches are not without issues, as they may suffer from unfalsifiability and be subject to cognitive bias~\citep{atrey2019exploratory} as well as provably wrong results~\citep{bilodeau2022impossibility}.
Visualizing the agent memory over trajectories~\citep{jaunet2020drlviz} or extracting finite-state models~\citep{koul2018learning} are explored to improve understanding of agents' behavior, as well as leveraging Markov decision processes~\citep{finkelstein2022explainable, zahavy2016graying} to generate explanations or detect sub-goals or emerging structures~\citep{rupprecht2019finding}. 
As RL methods may sometimes learn spurious correlations, interpretability methods were used to help identify and resolve the causal confusion~\citep{gajcin2022reccover} and further our understanding using counterfactuals~\citep{deshmukh2023counterfactual, olson2019counterfactual}. 

\subsection{Chess and Artificial Intelligence}
Chess has been a test bed for AI ideas for decades.
Early engines were based on human knowledge, and their super-human strength came from their computational capacity, which allowed them to consider a number of variations orders of magnitude above the abilities of human chess players. 
The introduction of neural network and RL-based approaches aimed to revitalise the field, which led to a surge of improvements in computer chess engines. These advances were in part inspired by the prominent results of AZ in chess and its variants~\citep{Silver1140, az_variants_preprint, CACM-paper, zahavy2023diversifying}, and Lc0~\citep{leela}, an open-source re-implementation of the original model, is still competing at the highest level of computer chess. 

As interactions with chess engines play a key role in chess players' preparation and training, interpretability helps chess players understand the underlying positional and tactical motifs. To this end, prior work has looked at piece saliency~\citep{puri2020explain}, tree-based explanations~\citep{10.1007/3-540-60364-6_40} and natural language~\citep{jhamtani2018acl}. At the intersection of chess and language, ChessGPT has recently been proposed~\citep{feng2023chessgpt} to bridge the modality of policy and language. DecodeChess is a software project aimed at deriving explanations from engine search trees~\citep{decodechess}. 

Recently, AZ has was shown to encode human-like concepts in its network~\citep{mcgrath2022acquisition}, and concept probing techniques have also been explored with the network-based Stockfish chess engine~\citep{palssonunveiling}. This prior investigation of concepts in AZ did not consider search and move sequences, and was largely restricted to identifying pre-existing human concepts. Preliminary questions have been raised regarding whether human players have been adopting AZ's ideas~\citep{gonzalez2022alphazero}, as some prominent motifs had been analysed in detail in Game Changer~\citep{game_changer}. Recently, it was also shown that AZ may be susceptible to adversarial perturbations~\citep{lan2022alphazerolike}, underscoring the need for a better understanding of the learned representations. 

\section{What are concepts?\label{sec:assumptions_defs}} \label{sec:assumptions}
There are several possible definitions of a concept -- varying from a human-understandable high-level feature to an abstract idea.
In this work, we define concepts as a unit of knowledge. 
There are two key properties we focus on. 
The first is that a concept contains \textit{knowledge}: information that is useful; in the context of machine learning, we take this to mean that it can be used to solve a task. 
For example, consider the concept of a beak. We can \textit{teach} an algorithm or person (transfer of the knowledge) what a beak is. If the person grasps the beak concept, they can use it to identify birds. 
Second, a \textit{unit} implies minimality; it is concise and irrelevant information has been removed. 

There are many ways to operationalise this definition and properties, and we choose one of them: showing a concept can be \textit{transferred} to another agent to help them solve a task (e.g., follow the strategy represented in a concept). Being able to do so implies that the concept is self-contained and useful for the task. 

How do we represent concepts? We leverage rich literature that assumes 
concepts are linearly encoded in the latent space of a neural network~\citep{mcgrath2022acquisition, kim2018interpretability, gurnee2023finding, conneau2018you, tenney2019bert, nanda_othello_2023}. The latent space refers to the space spanned by post-activation features of a neural network. 
Although our assumption of linearity is a strong assumption, it has a surprising amount of empirical support: linear probing and related techniques have successfully extracted a wide range of complex concepts from neural networks across multiple domains~\citep{mcgrath2022acquisition, kim2018interpretability, gurnee2023finding, conneau2018you, tenney2019bert, nanda_othello_2023}. 
Although we may miss concepts with nonlinear representations, we nevertheless show that we can find useful concepts for our goal using purely linear representations.  

What types of concepts do we aim to discover in the RL setting?
We aim to discover concepts that give rise to a plan, where a plan is a deliberate sequence of actions optimizing for one or more relevant concepts.
We take deliberate to mean that there is an underlying reason. More specifically, we assume a plan is motivated by one or more concepts. 
Although the \textit{terminal} goal of a plan the same across states -- maximizing the outcome (win or draw) -- plans in a specific state will have more context-specific \textit{instrumental} goals along the way, for instance, capturing a particular piece in an advantageous position, or maximising one's board control. We assume that plans in similar contexts will share similar instrumental goals, and thus give rise to similar concepts. 

\section{Discovering concepts} \label{sec:discovering_concepts}

Our method can be summarised into (1) excavating vectors that represent concepts in AZ using convex optimisation, (2) filtering the concepts based on teachability (whether it is \textit{transferable} to another AI agent) and novelty (whether it contains some information that is not present in human games). The resulting set of concept vectors is then used to generate chess puzzles (chess positions and solutions), which are presented to human experts (top chess grandmasters) for final validation. 

\subsection{Excavating concept vectors}\label{sec:convex_opt}
To find concepts, we develop a new method since (1) the model input is a mix of binary and real-valued
inputs (e.g., saliency maps typically take as input continuous values and are generally not suitable for binary values) and (2) we want to develop an interpretability tool to analyse both parts of AZ machinery -- the policy-value network and MCTS. Leveraging both the network and MCTS is crucial, since each component plays a different yet important role in deciding the move (see \S\ref{appendix: how AZ plays chess} for more detail).
We formulate concept discovery as a convex optimisation problem. 
Using a convex optimisation framework is not new; many existing methods for finding concept vectors, such as non-negative matrix formulation, can often approximated as a convex optimisation problem~\citep{ding2008convex}. 

For each concept vector we want to find, we formulate a separate convex optimisation problem. As mentioned in \S\ref{sec:assumptions}, we define a concept as \textit{a unit of knowledge}. Minimality is achieved by encouraging sparsity~\citep{tibshirani1996regression} through the $L_1$ norm 
\begin{align}
    & \min \| v_{c,l} \|_1 \nonumber  \\
    & \text{such that concept constraints hold},
\end{align}
where $v_{c,l} \in \mathbb{R}^{d_{l}}$ is a vector that lives in latent space of layer $l$ to represent concept $c$, and $d_l$ is the dimension of layer $l$. 

We outline the concept constraints used for two different types of concepts: static and dynamic concepts.
Static concepts are defined to be found in a single state, whereas dynamic concepts are found in a sequence of states. An example of a static concept in autonomous driving is that the car is \textit{located on the highway}. A dynamic concept is that the car is \textit{accelerating}. While our framework only aims to discover dynamic concepts, we use static concepts to validate our method. 

\subsubsection{Concept constraints for static concepts}\label{sec:cp_static_concept}
Static concepts are defined as concepts that only involve a single state. We use \textit{supervised data} (labels indicate whether a state contains a concept $c$) to learn static concept vectors. These concepts encode human knowledge, and therefore, we can use them to validate our approach.
One example of a static concept is the concept of `space', which we can infer from a single state. For now, assume we have binary concepts\footnote{We show how to handle non-binary concepts in \S\ref{appendix:cp_formulations}.} and denote the presence of concept $c$ (concept score) in chess position $x$ by $c(x)=1$, and $c(x)=0$ otherwise. For each concept $c$, we can split a general set of chess positions $\Xset$ into positive examples $\Xset^+$, where the concept is present, and $\Xset^-$, where it is absent
\begin{align*}
    \Xset^+ &= \{x\in \Xset: c(x) = 1\}\\
    \Xset^- &= \{x\in \Xset: c(x) = 0\}.
\end{align*}
These positive and negative examples allow us to generate corresponding positive and negative examples of latent representations (intermediate post-activation representations in the network). The function $f_l(x)$ generates an activation for layer $l$ given an input $x$:
\begin{align*}
    \Zset^+_l &= \{f_l(x): x\in \Xset^+\}\\
    \Zset^-_l &= \{f_l(x): x\in \Xset^-\},
\end{align*}
where $z_l = f_l(x)$ denotes the latent representation obtained at layer $l$ by passing input $x$ through the network. See \S\ref{appendix:AZ PV network} for further details on how $z_l$ is extracted. 

The convex optimisation goal is to learn a sparse vector $\vcl$ that represents a concept $c$. We hypothesise that the inner product $\vcld z_l$ is higher\footnote{A larger inner product corresponds to a higher cosine similarly.} for activations from $\Zset^+_l$ (the set where the concept is present) than for activations from $\Zset^-_l$ (the set where the concept is absent).
Thus, the formulation becomes
\begin{equation}\label{eq:convex_opt_basic}
    \min \| \vcl \|_1 \quad
    \textrm{such that}
    \quad \vcld z^+_l \geq \vcld z^-_l \quad \forall\,z_l^+ \in \Zset^+_l,\, z^-_l\in \Zset^-_l.
\end{equation}
We can evaluate how well a concept is represented by $\vcl$ in the supervised setting by splitting $\Xset$ into two sets:  $\Xset_{\mathrm{train}}$ and  $\Xset_{\mathrm{test}}$ and
then $\vcl$ only using  $\Xset_{\mathrm{train}}$.
We then measure the fraction of elements in $\Xset_{\mathrm{test}}$ on which the concept constraints hold.
If $\vcl$ represents the concept $c$ well, we expect the concept constraint to hold on previously unseen activations derived from $\Xset_{\mathrm{test}}$.

\subsubsection{Concept constraints for dynamic concepts} \label{sec:cp_dynamic_concept}
Dynamic concepts are defined as concepts found in a sequence of states. While $\vcl$ is found in the activation space of the policy-value network, we use the Monte Carlo Tree Search (MCTS) statistics to find candidates for meaningful sequences of states. MCTS generates a tree of possible moves and subsequent responses from the current chess position $x_0$
(for details on the implementation of MCTS see~\cite{schrittwieser2019mastering}). The exact details are not essential to understand for our procedure; what matters is that AZ chooses a rollout $\xp = (x^+_1, x^+_2, x^+_3, \ldots, x^+_T)$, where $T$ is the maximum depth of the rollout, which terminates in the most favourable state according to AZ. We contrast this optimal rollout $\xp$ with a sub-par rollout $\xm$, which is defined as a path in the MCTS search tree that is suboptimal, according to the value estimate or visit count in MCTS.  

The intuition behind our procedure is that $\xp$ is chosen over $\xm$ because of a concept $c$, and we assume the concept $c$ is detectable by a linear probe at some layer $l$. 
The concept presence may affect planning in different ways. 
Consider two rollouts in MCTS, one chosen by AZ ($\xp$), and one not chosen by AZ, ($\xm$). 
There are three different possible explanations for why AZ chooses $\xp$ over $\xm$:
\begin{enumerate}
    \item \textbf{Active planning} $\xp$ increases the presence of a concept $c$.
    For example, the rollout $\xp$ may increase the concept of piece activity.
    \item \textbf{Prophylactic planning} $\xp$ \textit{avoids} increasing the presence of a concept $c$.
    An example may be that the plan in $\xp$ avoids losing a piece.
    \item \textbf{Random} $\xp$ is arbitrarily chosen above the $\xm$, as all concepts are equally present in both rollouts and the value estimates of the final states are approximately equal.
\end{enumerate}
 
We are interested in scenarios 1 and 2 but not scenario 3. Scenario 3 can be filtered out 
by leveraging the fact that the two rollouts will have similar value estimates and visit counts in the MCTS statistics. 

Using a similar approach to static concepts, we derive our concept constraints on the vector $\vcl$ by contrasting the positive and negative examples, except that this time our contrasting examples are pairs from the chosen rollout $\xp$ and the subpar rollout $\xm$. We denote the activations at layer $l$ at depth $t$ by $z^+_{t, l}$ and $z^-_{t, l}$ for positive and negative examples, respectively. A single pair of positive and negative rollouts gives rise to the following convex optimisation problem
\begin{align} 
    \min & \quad \| \vcl \|_1 \label{eq:dynamic_concept} \\
     \textrm{such that} & \quad \vcl \cdot z^+_{t, l} \geq \vcl \cdot z^-_{t, l} \quad  \forall \, t\leq T, \label{eq:dynamic-conv-opt}
\end{align}
for scenario 1, with the inequality reversed for scenario 2.   

\begin{figure}[h] 
\centering
\caption{Contrasting the optimal rollout with subpar MCTS rollouts at different time steps. The green rollout shows the optimal rollout, and the red rollouts depict subpar trajectories. At each time step, MCTS finds subpar trajectories. We include each of these pairs in the concept constraints.}
\begin{minipage}{0.3\textwidth}
\includegraphics[width=\textwidth]{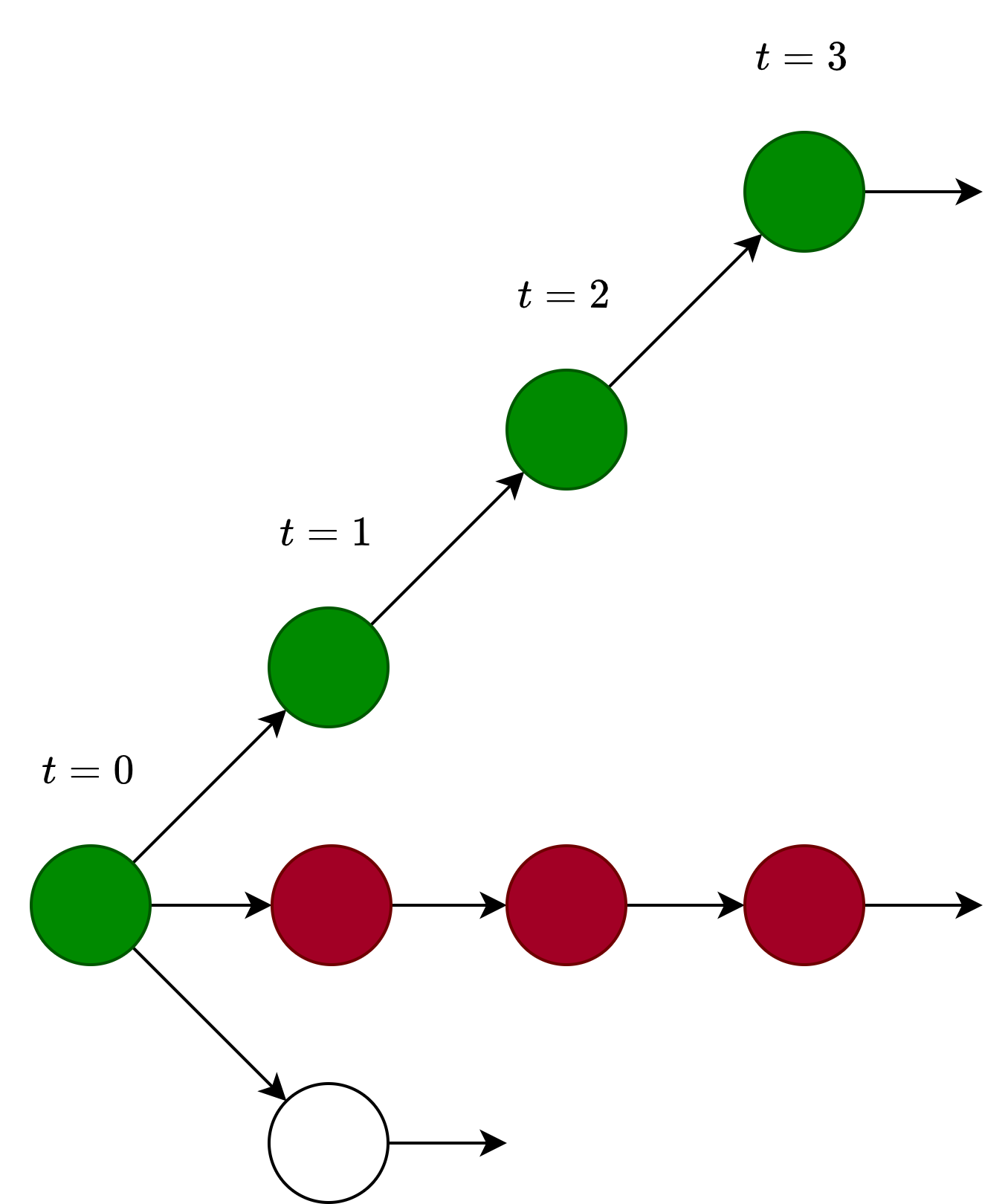}
\end{minipage}
\hspace{0.03\textwidth}
\begin{minipage}{0.3\textwidth}
\vspace{-0.223\textwidth}
\includegraphics[width=\textwidth]{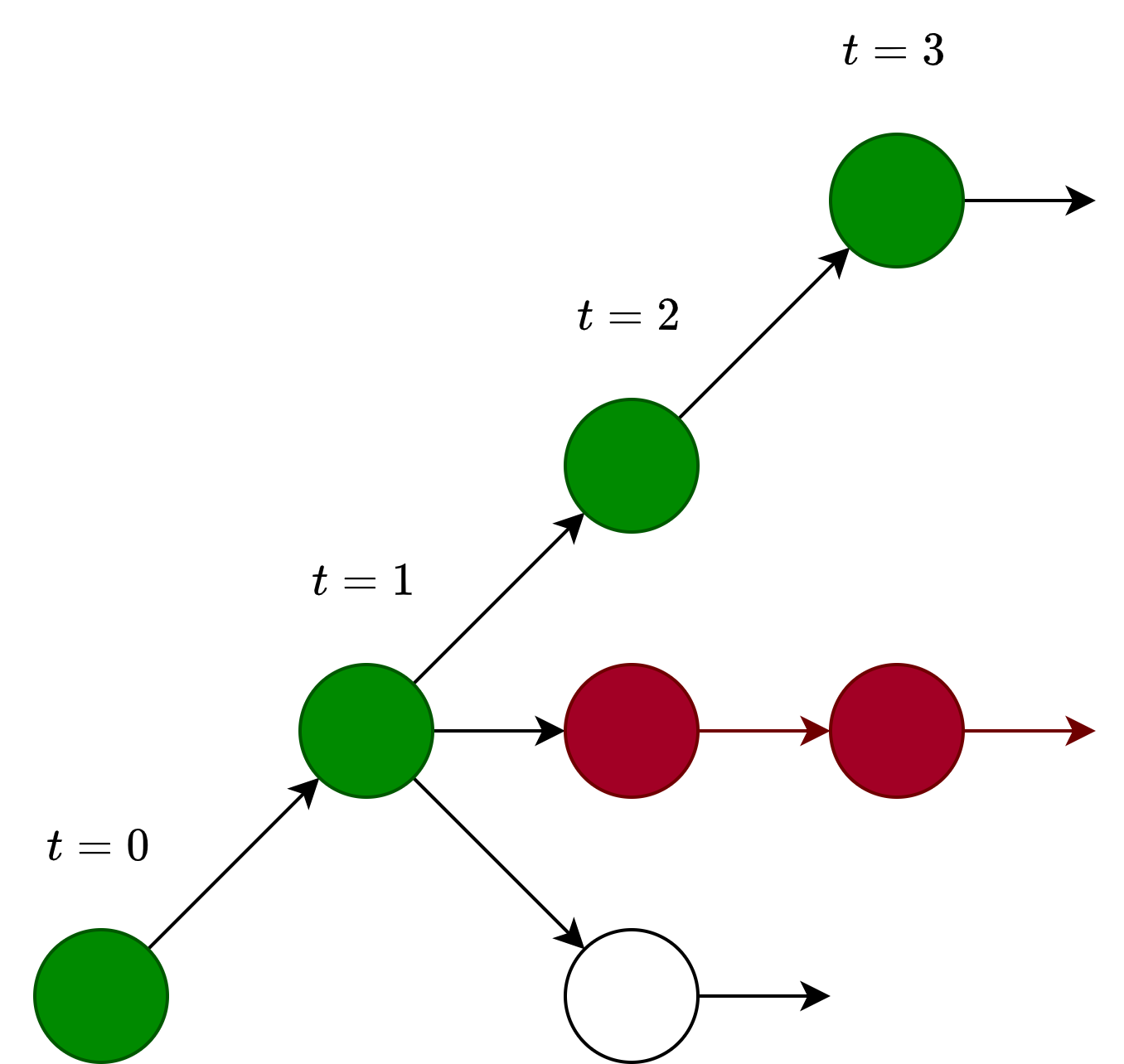}
\end{minipage}
\hspace{0.03\textwidth}
\begin{minipage}{0.3\textwidth}
\vspace{-0.223\textwidth}
\includegraphics[width=\textwidth]{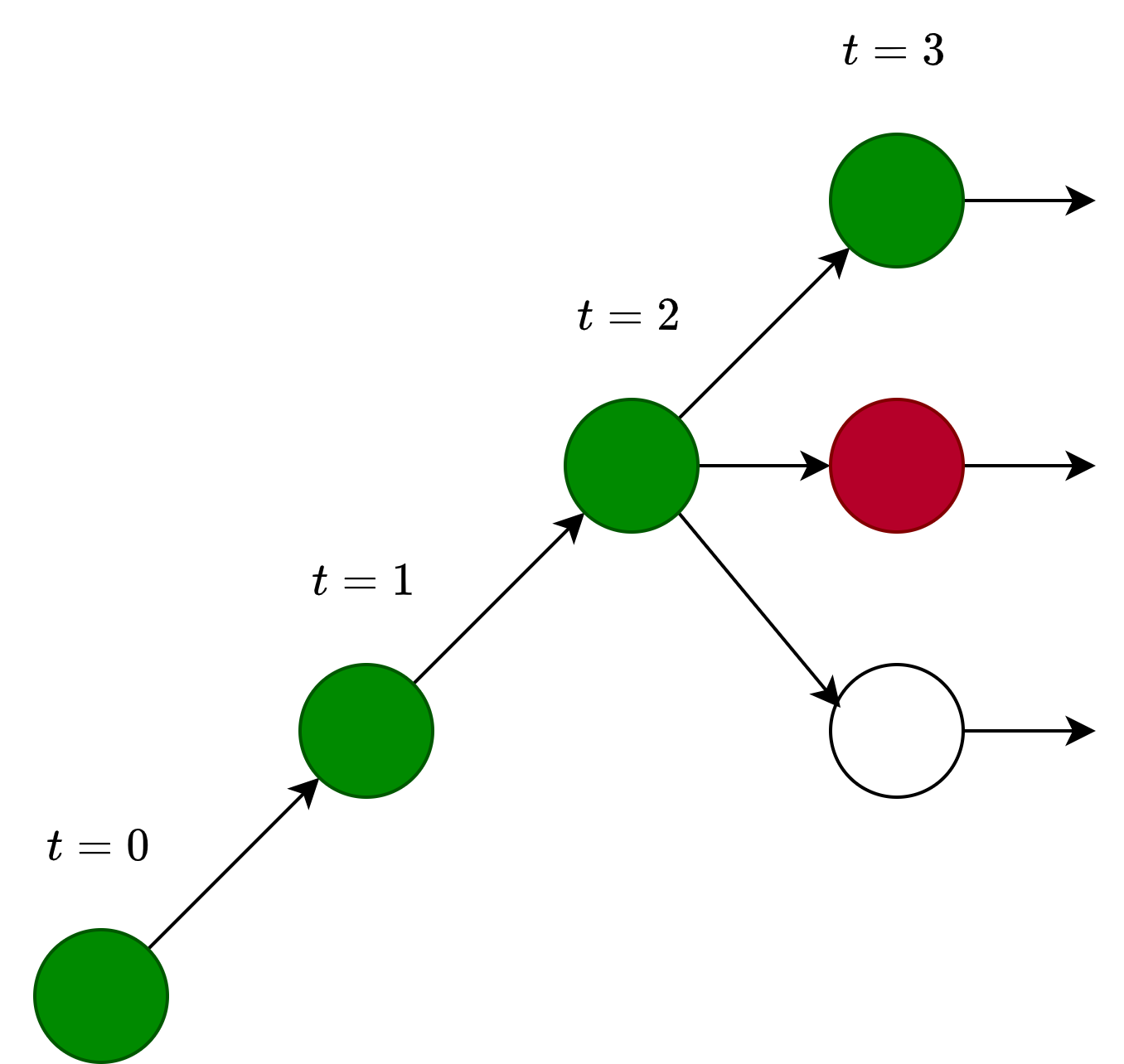}
\end{minipage}
\label{fig:mcts_tree}
\end{figure}
We extend this idea by contrasting the optimal trajectory with multiple subpar trajectories across different MCTS depths.
Figure~\ref{fig:mcts_tree} shows this idea. 
On the left side of Figure~\ref{fig:mcts_tree}, we find the optimal and subpar trajectory at the initial chess position, $t=1$. However, we can also use the MCTS statistics (value estimate and visit count) to find subpar trajectories at $t=2$ (shown in the middle) and $t=3$ (shown on the right). 
The idea behind using multiple subpar trajectories is to further reduce the solution space with the goal of reducing noise (thereby increasing the likelihood that the concept vector is meaningful) and decreasing the likelihood of learning a polysemantic vector. 

Let $\zp$ denote the latent representations in layer $l$ corresponding to the optimal rollout $\xp$, and $\zmj$ denote the latent representations in layer $l$ corresponding to the subpar rollout $\xmj$ selected at time step $j$.
We find the dynamic concept as follows:
\begin{align} \label{eq:tree_formula}
 \min & \quad ||\vcl||_1 \\
 \textrm{such that} & \quad \vcld z^+_{t,l} \geq \vcld z^{-}_{t,l,j} \quad \forall t \leq T, j \leq \tilde{T}  \nonumber,
 \nonumber
\end{align}
where $\tilde{T}$ denotes the maximum depth at which we find suboptimal rollouts. $\tilde{T} =3$ in Figure~\ref{fig:mcts_tree}. In general, we set $\tilde{T} = T-5$ to ensure the rollout is sufficiently deep. 
Details on how $T$ is set can be found in \S\ref{sec:hypers}. 

\subsection{Filtering concepts\label{sec:filtering}}
Our approach (described in \S\ref{sec:convex_opt}) provides many concept vectors, some or many of which represent known concepts or non-generalisable concepts (i.e., only applicable to a single chess position). In this section, we describe how we further filter concepts to ensure that they are useful (transferable) and novel. Our first filtering for usefulness is to see if we can teach a concept to a student network such that it leads to an improved performance on concept test positions. We describe this process of selecting only teachable concepts in \S\ref{sec:teachability}.
We further filter concepts based on novelty (\S\ref{sec:novelty}) by finding representations in AZ's self-play games that do not occur in a top-level human play dataset.

\subsubsection{Teachability} \label{sec:teachability}
Recall that we defined a concept as \textit{a unit of knowledge} -- a key aspect is that it is teachable to another AI agent or person, who can apply the concept to solve an unseen task (\S\ref{sec:assumptions_defs}). 
To ensure our concepts are teachable, we use \textit{teachability} as a selection criterion for the final concepts. 
The idea is simple: 
\begin{enumerate}
    \item \textbf{Phase 1: Baseline} find an agent that does not \textit{know} the concept. We can estimate the agent's knowledge by evaluating its performance on a concept-related task. 
    \item \textbf{Phase 2: Teach} the agent the concept using concept prototypes. 
    \item \textbf{Phase 3: Evaluate} the agent's performance on a concept-related task
\end{enumerate}
If a concept is teachable, we expect the agent's performance to improve between the first and third steps. We use a similar process to evaluate when we evaluate our approach with humans (\S\ref{sec:human}). 
\FloatBarrier

\paragraph{Selecting prototypes.}
In Phase 2, we use AZ as a teacher to supervise a student network on a set of chess positions called `prototypes' -- chess positions that exemplify the use of a concept.
For each candidate concept $c$, we have a concept vector $\vcl$. We want chess positions $x$ from a dataset $\Xset$ that epitomises the concept $c$. 
For static concepts, we do this by computing a concept score $c(x) = \vcld f_l(x)$ for every $x\in \Xset$ and then selecting the top $2.5\%$ of $\Xset$ according to the concept score $c(x)$.
We use $2.5\%$, as we found the concept score $c(x)$ to be comparable to $\vcld z_l^+$, where $z_l^+ \in \Zset_l^+$ is the training set used to find $\vcl$.
This procedure gives us a prototype set $\Xset^{\mathrm{proto}} = \left\{x\in \Xset: c(x)\text{ in 2.5th percentile of } c(x) \right\}$. 
For dynamic concepts, we find chess positions using the MCTS statistics. For each chess position $x \in \Xset$, we ran MCTS. Next, we found the chosen rollout $\Xset^{+}$ (and the corresponding latent representations $\mathbb{Z}^{+}$) and subpar rollout $\Xset^{-}$ (and the corresponding latent representations $\mathbb{Z}^{-}$) as in \S\ref{sec:cp_dynamic_concept}. For a prototype $x_i \in \Xset^{\mathrm{proto}}$, we require that $\vcl \cdot z^{+}_{i,t} \geq \vcl \cdot z^{-}_{i,t} \, \forall \, t \leq T$. 

\paragraph{Teaching and measuring learning.}
Intuitively, we can interpret the set of prototypes as a curriculum. We can split $\Xset_{proto}$ into a train set $\Xset_{train}$ and a test set $\Xset_{test}$.
The teacher (AZ) teaches the student by 
minimizing the KL divergence between the policies of the teacher ($\pi_t$) and the student ($\pi_s$) on the training prototypes: $\sum_{x_i \in \Xset_{train}} \text{KL}[\pi_t(x_i), \pi_s(x_i)]$. 
Then, to determine whether the student has acquired the new knowledge, we evaluate the student's performance on the test set prototypes by estimating how often the student and teacher select the same top-1 move
\begin{equation} \label{eq:teachability_train}
   T =  \sum_{x_i \in \Xset_{test}} \mathbbm{1} [ \text{argmax}(\pi_s(x_i)), \text{argmax}(\pi_t(x_i)) ].
\end{equation}
Teaching using any curriculum may improve students' performance on the task. To distill general learning from concept-specific learning, we compare the student's performance when taught using concept prototypes versus random chess positions sampled from AZ's games (but with meaningful plans). We sample the chess positions from AZ's games instead of human games for two reasons: (1) AZ's games tend to be of a higher quality than human games (as AZ has a higher Elo rating), and (2) the data is closer to AZ's natural training data (avoiding any confounding effects due out of distribution data).  

\begin{figure}[h]
\caption{Teachability: AZ Concepts. The y-axis shows how often the student and teacher select the same move (normalised version of  Equation~\ref{eq:teachability_train}), and the x-axis shows the training time step. 
The dark dotted lines show the level of a training checkpoint at which AZ obtains the same level on the concept set as our student. Each plot is a different concept found in layer $19$ (top) and in layer $23$ (bottom).}
\label{fig:teachability}
\centering
\includegraphics[width=\textwidth]{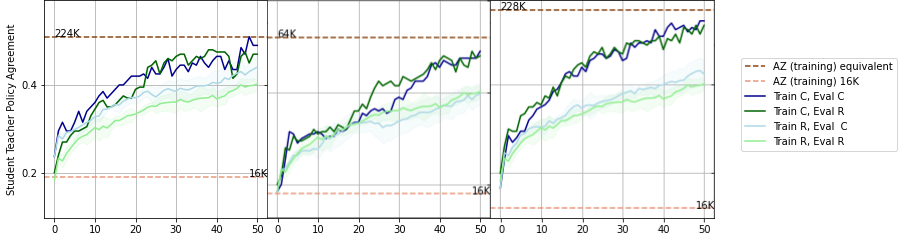}
\includegraphics[width=\textwidth]{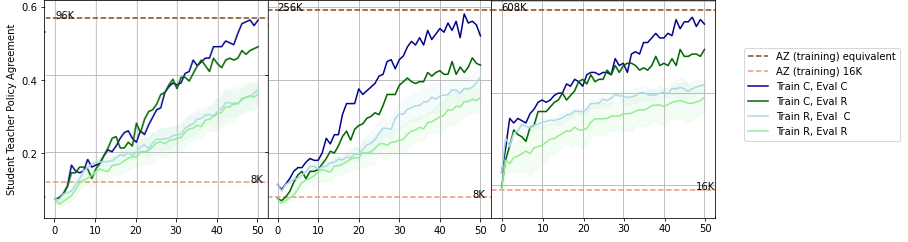}
\end{figure}

Figure~\ref{fig:teachability} shows the student's performance in four settings: 
(1) student trained on concept $c$ and evaluated on concept $c$ (dark blue line);
(2) student trained on concept $c$ and evaluated on random data from AZ's games (dark green line);
(3) student trained on random data and evaluated on concept $c$ (light blue line);
and (4) student trained on random data and evaluated on random data (light green line).
When teaching a student with concept-specific prototypes, the student also improves its performance on a random set of prototypes (dark green line) but less than on concept-specific prototypes (dark blue line). When a student was taught with randomly sampled chess positions, labelled with optimal play, it improves their performance significantly less (light lines) than when it was taught with concept-specific prototypes (dark lines). Naturally, the student learns quicker when taught with concept-specific prototypes (dark blue line) than random prototypes (light blue line).
We also observe that concepts can be taught efficiently. The student's performance after training for 50 epochs on a small set of prototypes would have taken $10K-100K$ epochs using self-play.
Recall that the student is evaluated on a holdout test set, ruling out the possibility that the student network memorised the chess positions. 

We select the student to be the latest checkpoint for which the top-1 policy overlap is less than $0.2$, resulting in $97.6 \%$ of the concepts being filtered out.

\subsubsection{Novelty \label{sec:novelty}}
There are many different ways to ensure the novelty of concepts. We take two simple approaches: (1) ensure concepts are learnt during the later stages of AZ's training and (2) filter concepts based on a novelty metric. 

\paragraph{Concepts are learnt during a late stage in AZ's training.}
To find chess positions that are complex, we leverage AZ's training history. We use two versions of AZ that differ by 75 Elo points.\footnote{For reference, the expected score of a player rated 75 Elo points higher is $0.68$, where $1$ point is given for a win, $0.5$ for a draw, and $0$ for a loss. 75 Elo points is a large Elo difference, particularly at AZ's playing strength.}
To find interesting chess positions, we run through each game and select chess positions where the two versions of AZ disagree on the best move according to their policies. We only use these chess positions to discover concepts. 

\paragraph{Setup to measure novelty.}
While the previous section ensures that concepts that emerge in the final stages of training in AZ are complex by construction, it does not ensure that the concepts are novel to humans. One way to validate novelty is to determine whether the concepts arise in AZ's games but not in human games. 
Leveraging the fact that concepts are represented in the latent space as a vector, 
we can compare the vector space of AZ games to that of human games.
Specifically, let $Z_l^a$ denote a matrix where we stack the latent representations in layer $l$ of $17,184$ chess positions sampled from AZ's games. 
Each row represents a chess position, and each column represents a dimension in latent space in layer $l$.
Similarly, we define $Z_l^h$ as a matrix of $17,184$ chess positions sampled from human games. 
Using the latent representations, (1) we first get evidence that AZ's game is likely to contain new concepts using a rank experiment, and then (2) measure the novelty scores by regressing concepts onto AZ's games vector space and human games vector space on which we filter concepts based. 

\paragraph{AZ games likely to contain new concepts compared to human games.\label{sec:rank}} 
First, we aim to establish whether AZ games contain \textit{something new} compared to human games. We approach the question by contrasting the number of concepts encoded in both. The number of the basis vectors (or ranks of $Z_l^h$ and $Z_l^a$) estimates the size of the span of the latent representations of the games - informally, we can think of it as a proxy for the number of concepts. 

As shown in Table~\ref{table:rank}, the ranks for $Z_l^h$ and $Z_l^a$ vary across different layers. We focus on the final layers in the architecture as these more directly impact the output. 
Note that while the rank of inputs for both humans and AZ games are similar, AZ games' rank is higher than the humans games' rank at layers $19$ (final layer in main bottleneck) and $23$ (policy layer), hinting that there might be concepts present in AZ games that are not in human games.
Therefore, we focus on finding new concepts in these layers. 

\begin{longtable}{llrrrrr} 
\caption{Rank of latent representation of Human Games and AZ's Games}
\label{table:rank} \\ 
\toprule
&   &  Input & Layer 19 & Layer 20 &  Layer 21 &  Layer 23 \\
\midrule
\endhead
\midrule
\multicolumn{5}{r}{{Continued on next page}} \\
\midrule
\endfoot
\bottomrule
\endlastfoot
\multirow{2}{*}{Rank}     &      Human data &   730 &    7857 &  64 &  86 & 6544 \\
    &     AZ data &   728  & 8269 &    64  &  88 & 6771 \\ \midrule 
\multirow{2}{*}{}     & Theoretical maximum rank & 7616 & 16384 & 64 & 256 & 16384 \\ 
    &     Dimensions & [8,8,119] & [8,8,256] & [64] & [256] & [8,8,256] \\ 
\end{longtable}

\paragraph{Novelty Scores for filtering} \label{sec:res_novelty}
We define the novelty scores based on how well a concept vector can be reconstructed using a set of basis vectors that arise in AZ's games. Naturally, a lower reconstruction loss means that the concept is better represented using a set of given basis vectors. In other words, we look for concepts that are better explained using AZ's language (basis vectors) than humans'.
We define novelty score as the difference between concept's reconstruction loss (see Equation ~\ref{eq:novelty}) when using basis vectors from humans' game and AZ games. A higher score means a closer alignment with the basis vectors arising from AZ's games. 

Specifically, for $Z_l^h$ and $Z_l^a$, we find the singular value decomposition to find the basis of the space spanned by AZ's and human games
\begin{align*}
Z_l^h &= U_l^h \Sigma_l^h V_l^{h \top}, \\
Z_l^a &= U_l^a \Sigma_l^a V_l^{a \top},
\end{align*}
where the columns of $U_l^h$ and $U_l^a$ form an orthonormal basis for the rows of $Z_l^h$ and $Z_l^a$, respectively; $\Sigma_l^h$ and $\Sigma_l^a$ are the singular value matrices; and the columns of $V_l^{h \top}$ and $V_l^{a \top}$ form the orthonormal basis for the columns of $Z_l^h$ and $Z_l^a$ respectively.

The novelty score of concept vector $\vcl$ is defined as
\begin{equation} \label{eq:novelty}
\min_{\beta_{i,l}} \left(\vcl - \sum_{i=1}^k \beta_{i,l} u^h_{i,l}\right)^2 - \min_{\gamma_{i,l}} \left(\vcl - \sum_{i=1}^k \gamma_{i,l} u^a_{i,l} \right)^2,
\end{equation}
where $\beta_{i,l}$, $\gamma_{i,l}$ are coefficients estimated using linear regression, $u_{i,l}^{a}$ and $u_{i,l}^{h}$ are columns of $U_l^{a}$ and $U_l^{h}$, respectively, and $k$ is the number of basis vectors used. 
We do not set $k$ as the rank of the matrix because they differ for $Z_l^a$ and $Z_l^h$, and doing so would favour the matrix with the largest rank. Instead, we estimate Equation~\ref{eq:novelty} for various values for $k$.

\begin{figure}[!ht]
\caption{Filtering concepts based on novelty scores. Concepts for which the reconstruction error using AZ's basis vectors is less than the reconstruction error using human game's basis vectors for every $k$ are accepted (not filtered). The darker green and blue lines show the average over the accepted and rejected concepts.}
\centering
\includegraphics[width=0.5\textwidth]{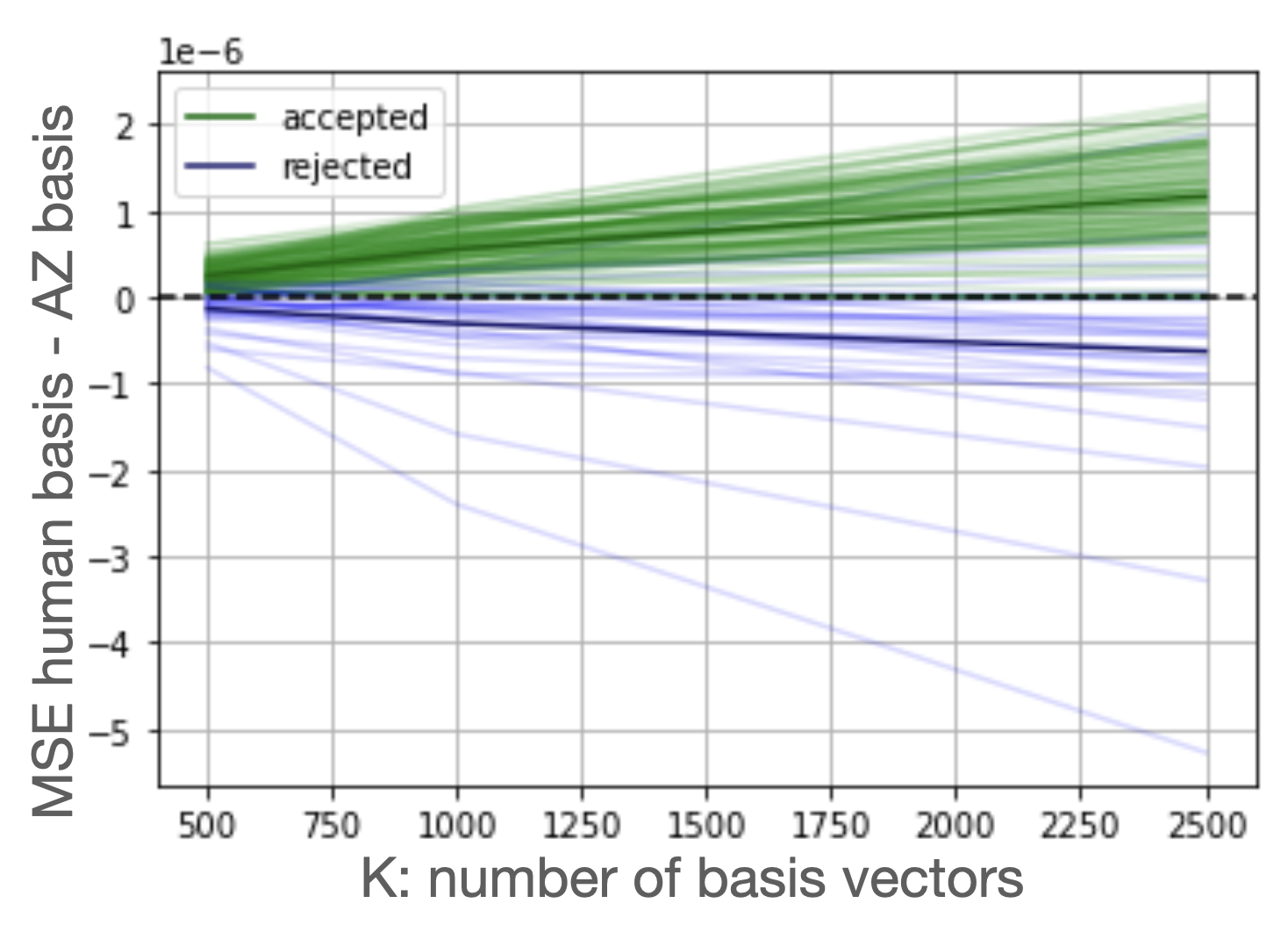}
\label{fig:novelty_score}
\end{figure}

Figure~\ref{fig:novelty_score} shows the novelty scores for concepts in layer $19$ for $120$ concepts. We accept concepts for which the reconstruction error using AZ's basis vectors is less than that of the human game's basis vectors for every $k$.
The light green lines denote the novelty scores for the concepts we accept, and the light blue lines denote the novelty scores for the concepts we reject. 

\FloatBarrier

Of the remaining concepts after teachability-based filtering, we remove a further $27.1 \%$ using the novelty metric. 

\section{Method evaluation} \label{sec:method_eval}
This section includes algorithmic evaluations of our proposed concept discovery method. Table~\ref{table:summary_dataset_main} summarises the datasets used; further details on each dataset can be found in \S\ref{appendix:dataset}, and implementation details for each dataset can be found in \S\ref{appendix:cp_formulations}. 

\begin{longtable}{p{4cm} c c c c} 
\caption{Datasets Summaries: from concepts more known to humans (top rows) to AZ (bottom rows). S denotes strategic and T denotes tactical.}
\label{table:summary_dataset_main} \\ 
\toprule
\textbf{Name} &  \multicolumn{2}{c}{\textbf{Concept Type}} &  \textbf{Type of}  & \textbf{Complexity} \\  
& \textbf{S vs. T} & \textbf{Game Phase} &\textbf{Knowledge} &  \\ 
\midrule
\endhead
\midrule
\multicolumn{4}{r}{{Continued on next page}} \\
\midrule
\endfoot
\bottomrule
\endlastfoot
Piece  & N/A & All & Human & Low \\
Stockfish  & Both & All & Human & Varies \\ 
Strategic Test Suite (STS) &  S & Middle/End & Human & Medium \\ 
Opening  & S & Begin & Human/AZ & Varies \\
AlphaZero (games)  & Both & All & AZ & High \\ 
\end{longtable}

\subsection{Evaluation of the proposed convex optimisation framework for finding concept vectors} \label{sec:eval_supervised}
Using the datasets mentioned above, we find concept vectors using the approaches described in \S\ref{sec:cp_static_concept} and \S\ref{sec:cp_dynamic_concept}. While we use layer $19$ for all other sections due to its potential novelty according to spectral analysis in \S\ref{sec:novelty}, we conduct our evaluation on a few additional layers here:
the first latent representation in the policy head (layer $23$); and the latent representations in the value head (layer $20$ and $21$) (See \S\ref{appendix:AZ PV network}). These layers are selected due to their proximity to the network outputs -- the policy and value estimate.

We first validate our approach by showing the convex optimisation formulation can be used to find the vector representations of a concept using labelled data (\S\ref{res:test_constraints}) and show that this can be done efficiently with a small number of labels (\S\ref{res:sample_eff}). 
Next, we further validate our approach by showing that amplifying the concept vector in the latent representation 
makes AZ's moves to be more similar to the concept (\S\ref{res:concept_amplification}). 
This section focuses on validating the convex optimisation framework. 

\subsubsection{Do the concept constraints hold for a test dataset?
} \label{res:test_constraints}
To evaluate our framework and thus the quality of the vector representations of the supervised concept, we measure the percentage of times the concept constraints hold on the test set ($80/20$ train/test split), as shown in Table~\ref{table:constraint_percentage}. 
We find that most datasets led to a high accuracy, which is an indication of quality concept vectors. A concept-level breakdown of the accuracy for each dataset can be found in \S\ref{appx:res:cp}.

\begin{longtable}{lrrrr}
\caption{Evaluation: $\%$ the concept constraints hold on test data. The standard error is shown in parentheses.}
\label{table:constraint_percentage} \\ 
\toprule
 Concept  &  Layer 19 &  Layer 20 &  Layer 21 &  Layer 23 \\
\midrule
\endhead
\midrule
\multicolumn{5}{r}{{Continued on next page}} \\
\midrule
\endfoot

\bottomrule
\endlastfoot
    Pieces &     0.99 (0.00) &     0.98 (0.00) &     0.95 (0.00) &     0.99 (0.00) \\
 Stockfish &     0.76 (0.03) &     0.75 (0.03)&     0.73 (0.02) &     0.77 (0.02)\\
       STS &     0.92 (0.09) &     0.92 (0.09)&     0.92 (0.06)&     0.90 (0.06) \\
   Opening (general) &     1.00 (0.00)&     1.00 (0.00) &     1.00 (0.00) &     1.00 (0.00) \\
  Openings (per line) &     0.99 (0.13) &     0.99 (0.14) &     0.99 (0.14) &     0.99 (0.13) \\
\end{longtable}

\subsubsection{How many data points do we need to learn a concept?} \label{res:sample_eff}
We can use labelled examples to evaluate the concept vector quality (as in the previous section), but they can be hard to come by in practice. This section shows that our formulation can find concept vectors efficiently using a few examples (for the concept constraints). 
We measure the test set accuracy (as in \S\ref{res:test_constraints}) while varying sizes of the training set for two datasets: pieces and the strategic test suite (STS). Here, we discuss the result for the piece dataset. The results for the STS dataset are similar and can be found in \S\ref{appx:sample_efficiency}.

\begin{figure}[ht] 
\centering
\caption{Sample efficiency of convex optimisation framework, averaged across $10$ seeds, for the bottleneck layer ($19$), value head ($20$) and policy head ($23$).}
\includegraphics[width=0.32\textwidth]{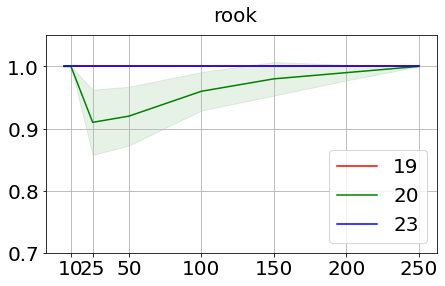}
\includegraphics[width=0.32\textwidth]{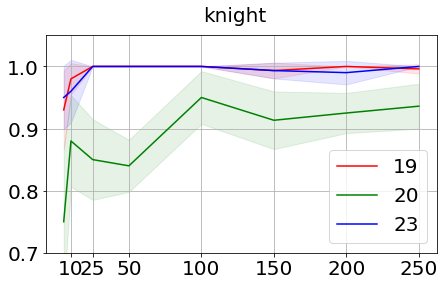}
\includegraphics[width=0.32\textwidth]{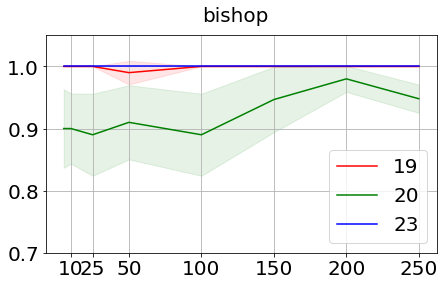}
\includegraphics[width=0.32\textwidth]{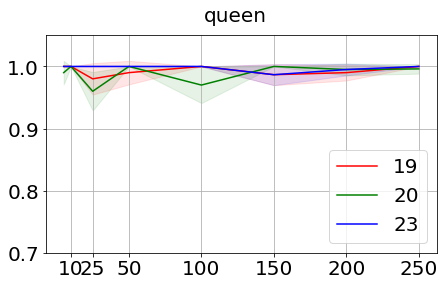}
\includegraphics[width=0.32\textwidth]{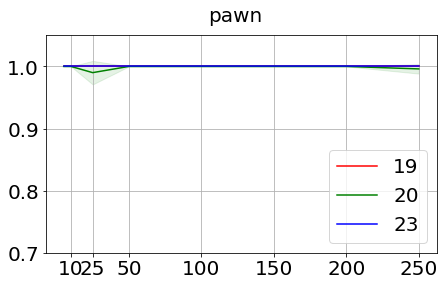}
\label{fig:sample_efficiency}
\end{figure}

As shown in Figure~\ref{fig:sample_efficiency},
we find that the method reaches close to full-set accuracy with only a few samples -- often as little as $10$ data points on the pieces dataset. 
Interestingly, we observe relatively lower performance in the value head (layer $20$) than the policy head (layer $23$). 
One potential explanation is that the concept of a specific piece no longer has to exist when estimating the value, which is a scalar -- a highly compressed representation of the state of the game.
We speculate that it is possible that simple concepts are combined with other concepts.
For example, the network may encode the presence of light pieces, i.e., bishops and knights, rather than the presence of bishops. This may explain the relatively low, but not significantly lower performance in layer $20$. 

\subsubsection{Does amplifying concept vectors increase concept-related behavior?} \label{res:concept_amplification}
\begin{figure}[ht] 
\centering
\caption{Performance improvement in solving puzzles using STS dataset across different $\alpha$ values (x-axis). Layers $18$ (left) and $19$ (centre) are before value/policy head split, and the policy head (layer $23$) (right). Each line indicates a set of concepts with different quality (measured by test accuracy as done in \S\ref{res:test_constraints}). Higher quality (high threshold, orange line) achieves the highest improvement.}
\includegraphics[width=0.3\textwidth]{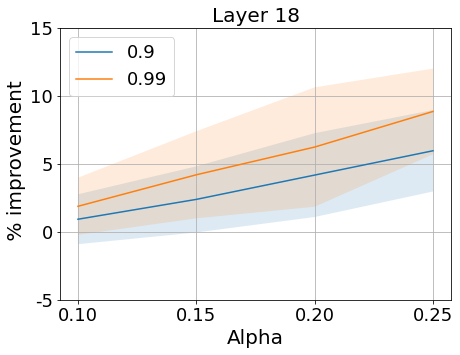}
\hspace{0.02\textwidth}
\includegraphics[width=0.3\textwidth]{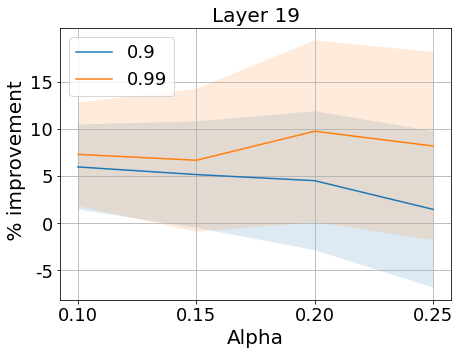}
\hspace{0.02\textwidth}
\includegraphics[width=0.3\textwidth]{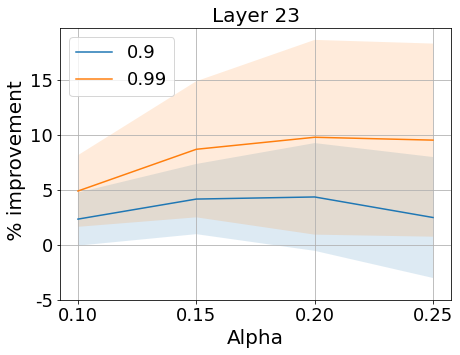}
\label{fig:amplification}
\end{figure}

We want to (1) determine whether the concept vector captures the intended concept and (2) understand whether the concept influences AZ's output (policy). (2) is important as a concept may exist in a latent representation but not be used by the network for its predictions. To investigate these properties, we use concept amplification.
Let $z_{l}$ denote the latent representation in layer $l$ of a chess position $x$. 
To amplify the presence of a concept, we nudge the latent representation in the direction of the concept vector $\vcl$
\begin{equation} \label{eq: concept_amplify}
    \Tilde{z}_{l} = (1-\alpha)z_{l} + \alpha \beta \frac{\| z_{l} \|}{
    \| \vcl \|} \vcl, 
\end{equation}
where $\alpha \in [0,1]$ and $\beta$ are hyper-parameters for 
the size of the perturbation; and $||\cdot||$ is the $\ell_2$ norm. We use cross-validation to determine $\beta$, and find that $\beta = 0.01$ is the best overall (see \S\ref{appx:beta}).
We report our results for various $\alpha$ in Figure~\ref{fig:amplification} using the 
STS dataset. The STS dataset includes different types of puzzles grouped according to a strategic theme (such as 'square vacancy'). For each concept, there are $100$ chess positions ($\Xset$) and a solution set $\mathbb{S}_i$ for each chess position $x_i \in \Xset$. 
The solution moves require applying the concept given a chess position. 
As a baseline, we first evaluate AZ's performance by recording the percentage of times the move selected by AZ under the policy is in the solution set:
\begin{equation}
    A = \sum_i \mathbbm{1} [\text{argmax} \ \pi_l(z_{i,l}) \in \mathbb{S}_i],
\end{equation}
where $\pi_l(z_{i,l})$ is AZ's policy on the latent representation $z_{i,l}$, and $\mathbbm{1}[\cdot]$ is an indicator function that is equal to $1$ if the move selected by AZ is in the solution set $S_i$. 
We compare the performance difference between with and without concept amplification and report the normalised values $(\tilde{A}-A)/A$ in Figure~\ref{fig:amplification}.

As this experiment focuses on the impact of the concept on the predicted move (only the policy output, not the value),
we analyse the performance of concepts found in layers $18$ and $19$ (layers before policy and value head split, see \S\ref{appendix:AZ PV network}), and layer $23$ (policy head). 

Each line in Figure~\ref{fig:amplification} represents the results for a different concept quality, where concept quality is measured by test accuracy as in \S\ref{res:test_constraints}.
We observe that amplifying the concept can improve AZ's performance on the puzzles of the concept. Naturally, the quality of concepts influences this; 
concepts with higher test accuracy lead to a larger performance improvement, suggesting 
a higher test score is a good proxy for how well the concept vector captures the semantic meaning of the concept. 

\section{Human evaluation} \label{sec:human}
We investigate whether top chess grandmasters can successfully learn and subsequently apply the concepts we discovered (in \S\ref{sec:discovering_concepts}). In particular, we investigate if it is possible to learn these concepts via exposure to a small set of prototypes of each concept.
These prototypes are found using a concept vector as in \S\ref{sec:teachability}, and then filtered based on a set of criteria aimed at identifying the most relevant high-quality prototypes (see \S\ref{appx:human} for more information).
Learning from prototypes is similar to the established approaches to teaching chess. Chess students are often presented with puzzles sampled according to a theme (opening, chess position type, piece sacrifices, etc.), and practicing puzzle solving (i.e., finding the correct next moves) is one way of improving their overall strength and ability~\citep{chuchelov_interview}. 

Henceforth, we use puzzles to denote prototypes and their `solutions' (AZ's selected move). 
Human evaluation with grandmasters follows three phases, similar to how teachability is measured \S\ref{sec:teachability}:
\begin{itemize}
    \item \textbf{Phase 1: Measuring baseline performance.} Each grandmaster provides solutions for a set of provided puzzles corresponding to a set of concepts. 
    This phase determines the baseline performance: the number of puzzles in which the chess grandmaster gets the continuation correct \textit{before} the learning phase.
    \item \textbf{Phase 2: Learning from AZ's calculations.} The same puzzles as in Phase 1 are shown to chess grandmasters alongside the associated AZ's suggested top line based on MCTS calculations for each puzzle. This serves as the simplest way of teaching. 
    \item \textbf{Phase 3: Measuring final performance.} Grandmasters are tasked with providing solutions for a test set of unseen puzzles sampled from the same concepts they have seen in Phase 1.
    We compute the grandmasters' accuracy on the puzzle test set and compare it to their performance on the puzzle training set in Phase 1 to measure whether their performance changes.
\end{itemize}
We work with four players, all of whom hold the grandmaster title; one of our participants is rated 2600-2700, and three are rated 2700-2800. 
At each stage, we also asked the grandmasters to provide a summary of their thought process in free form.
As each puzzle is complex, the study participants would likely spend considerable time on each puzzle to analyse the chess positions in great depth.
Therefore, to avoid over-burdening the study participants, they were presented with four puzzles (per concept) for 3-4 discovered concepts at each stage of the study. In total, each grandmaster saw 36 to 48 chess puzzles. While there was some overlap between study participants in terms of puzzles shown, different participants were shown different concepts. Given that the participants had a limited amount of time and the high time investment per puzzle, this randomisation allows us to explore a more extensive concept set (compared to showing every grandmaster the same concepts and puzzles). 
While some chess concepts may be more teachable than others, it is difficult to establish this beforehand without inserting human bias.

\subsection{Grandmaster performance}
\begin{longtable}{c r r r r} 
\caption{Improvements in grandmasters' performance. 
The percentage scores are the \% of puzzles that the grandmaster solved correctly (according to AZ's solution). \# Puzzles is the number of puzzles shown to the grandmaster in total.}
\label{table:perf} \\ 
\toprule
Grandmaster & \multicolumn{3}{c}{Percentage Score} & 
\multirow{2}{*}{\# Puzzles}\\
 &  Phase 1 & Phase 3 & Improvement &  \\ 
\midrule
\endhead
\midrule
\multicolumn{4}{r}{{Continued on next page}} \\
\midrule
\endfoot
\bottomrule
\endlastfoot
1  & 0 & 42 & +42 & 36 \\ 
2  & 33 & 58 & +25  & 36 \\ 
3 & 25 & 42 & +16  & 36 \\ 
4  & 38 & 44 & +6   & 48 \\ 
\end{longtable}

Overall, we find that all study participants improve notably between phases 1 and 3, as shown in Table~\ref{table:perf}, suggesting that the chess grandmasters were able to learn and apply their understanding of the represented AZ chess concepts. 
The magnitude of improvement does not correlate with the chess player's strength (i.e., Elo rating).
Below, we discuss factors that may have influenced performance:
\begin{itemize}
    \item \textbf{Variability in difficulty and quality.} We filter prototypes (as described in \S\ref{sec:filtering}) to ensure quality and complexity. However, the difficulty and quality of puzzles players received may vary across puzzles. 
    \item \textbf{Variability in teachability.}
    While we filter based on teachability (as described in \S\ref{sec:teachability}), the teachability metric is based on teaching the concept to another AI agent, which may be inherently different from teaching humans. 
    \item \textbf{Overthinking.} We observed that grandmasters often mention AZ's move in Phase 3 in their free-form comments but ultimately did not choose the move (which was not counted as `correct'). We speculate that this may be because players are more familiar with existing strategies in their decision process, despite having learnt the concept.
\end{itemize}

\subsection{Qualitative analysis of the concepts and illustrative examples}
In this section, we provide a qualitative analysis of some concepts with the associated puzzles (plotted using the \texttt{chess} python package \citep{chesspackage}), along with the grandmasters' analysis of the puzzles and the provided AZ suggestions. Any text or notation referring to the chess board or moves is written in \ct{this font}. When giving direct quotes, we refrain from providing player's names to preserve anonymity. These examples will cover some cases of successful and some cases of unsuccessful learning. We discuss potential sources of differences between humans and AZ in \S\ref{sec:diff_human_az}.

Overall, the grandmasters appreciated the concepts, describing them as `clever' (Figure~\ref{fig:concept6_phase1_left}), `very interesting' (Figure~\ref{fig:teaching_succ_1}), and `very nice' (Figure~\ref{fig:teaching_succ_1_phase3}). 
Further, they found that the ideas often contained novel elements,
commenting that the moves were `something new' and even `not natural' (Figures \ref{fig:teaching_succ_1}) and \ref{fig:concept6_phase1_right}). 
Often, the grandmasters found the positions were very complex -- making remarks such as that it was ``very complicated - not easy to understand what to do''. Even when seeing AZ's solutions, they remarked that it was a `very nice idea which is hard to spot' (Figure~\ref{fig:risk}).

\subsubsection{Concept example: positive knowledge transfer}

Here, we explore a concept that possesses both strategic and prophylactic characteristics, involving plans that improve the player's piece placement while restricting the opponent's piece activity. This concept contains an additional element of exploiting tactical motifs and weaknesses, combining strategic and tactical play. We speculate that the concept is learnable to humans, as the grandmaster who trained on the concept puzzles improved their performance between Phases 1 and 3. In Phase 1, they were not able to identify AZ's plan in any concept puzzle (0/4), whereas they successfully identified the correct AZ plan in 2/4 of the concept puzzles in Phase 3. 

\begin{figure}[!ht]
\caption{Puzzle of shown in Phase 1. White is to play.}
\centering
\includegraphics[width=0.45\textwidth]{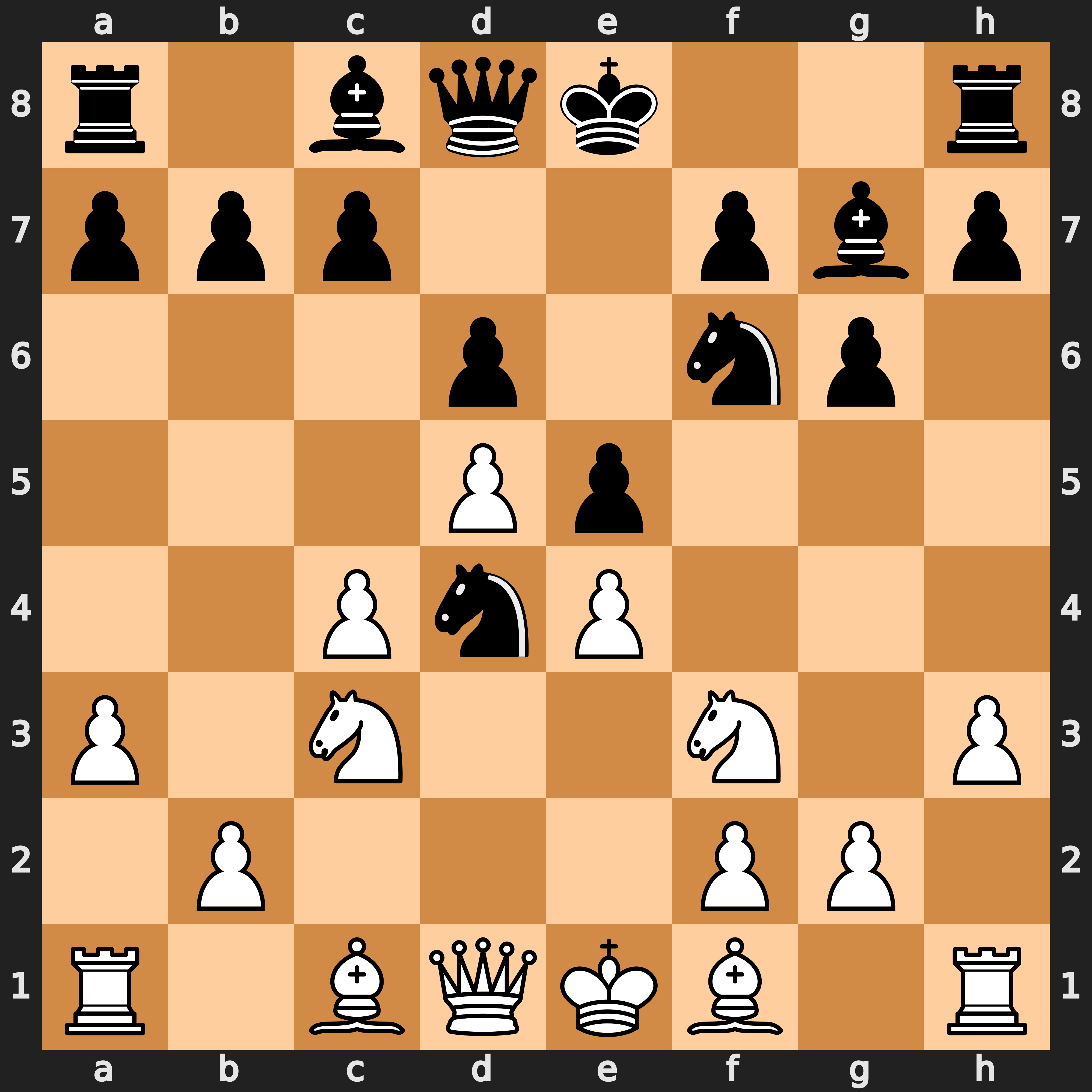} \\
\vspace{0.2cm}
\fbox{\begin{minipage}[t]{\textwidth}
\begin{flushleft}
\textbf{AZ's calculations:}
\ct{9.Bg5 (9. Be3 Nxf3+ (9...O-O 10.Nxd4 exd4 11.Bxd4 Nxe4 12.Nxe4 Qh4 13.Bxg7 Qxe4+ 14.Qe2 Qxe2+ 15.Bxe2 Kxg7 16.Kd2 Bd7) 10.Qxf3 Nh5 11.g3 O-O 12.Be2 f5 13.exf5 Bxf5 14.Qg2 Bd7 15.O-O-O a6 16.c5) 9...Nxf3+ 10.Qxf3 h6 11.Be3 b6 12.Bd3 Qe7 13.Qe2 a6 14.Qd2 Nd7 15.Bc2 h5 16.O-O-O h4 17.g4 White is slightly better}
\end{flushleft}
\end{minipage}}
\label{fig:concept6_phase1_left}
\end{figure}

\begin{figure}[!ht]
\caption{Second puzzle shown in Phase 1. White is to move.}
\centering
\includegraphics[width=0.45\textwidth]{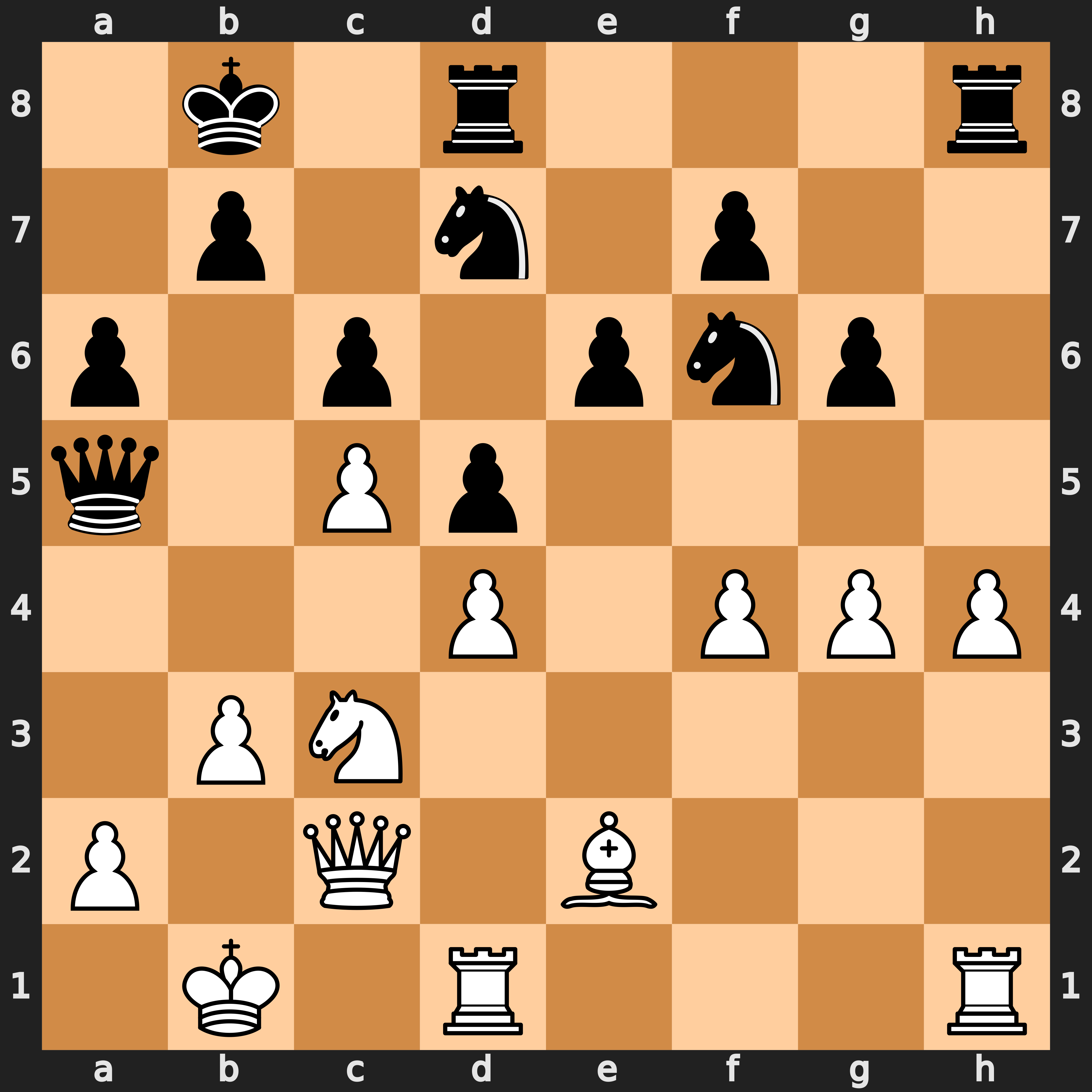} \\ 
\vspace{0.2cm}
\fbox{\begin{minipage}[t]{\textwidth}
\begin{flushleft}
\textbf{AZ's calculations:} \ct{21.Qd2 Qc7 (21...Rh7 22.h5 Rdh8 23.Rh3 Qc7 24.b4 Ne4 25.Nxe4 dxe4 26.g5 gxh5 27.a4) 22.Bf3 Rh7 23.Rh3 Rdh8 24.Rdh1 Ne8 25.Bd1 (25.b4 f6 26.Bd1 e5) White is slightly better}
\end{flushleft}
\end{minipage}}
\label{fig:concept6_phase1_right}
\end{figure}

Figures~\ref{fig:concept6_phase1_left} and~\ref{fig:concept6_phase1_right} show two of the puzzles provided to a grandmaster in Phase 1. 
In Figure~\ref{fig:concept6_phase1_left}, AZ plays the move \ct{9.Bg5}; the idea is to provoke \ct{9...h6} before retreating to the square \ct{e3}, thereby inducing a structural weakness.
Instead, the grandmaster chose \ct{9.Be3}, a natural move to develop the bishop. After seeing AZ's calculations, the grandmaster acknowledged the strength of provoking this weakness:
\begin{displayquote}
``\ct{9.Be3} allows Black the clever option of playing \ct{[9...Nxf3 10.Qxf3] Nh5} [as provided by AZ, followed by] f5 ... but \ct{9.Bg5} is clever as it provokes h6 after which f5 is not great and also [the pawn on] h6 serves as a hook for the pawn advance \ct{g4}-\ct{g5}. Blacks' plan of stopping \ct{c5} with \ct{b6} and playing \ct{h4}-\ct{h5} is interesting too but with \ct{g4} anyways White manages to open lines so I would prefer White there."
\end{displayquote}

The idea of playing provocative bishop moves to induce pawn weaknesses is not new and arises in human play. However, the interesting and potentially novel element here lies in the planned strategic queen sacrifice that emerges in one of the critical lines in the MCTS calculations.
Consider one such critical continuation: 
\begin{displayquote}
\ct{9.Bg5 h6 10.Be3 O-O 11.Nxd4 exd4 12.Qxd4 Ng4 13.hxg4!}
\end{displayquote}

\begin{figure}[!ht]
\caption{Further analysis of the puzzle in Figure~\ref{fig:concept6_phase1_left}. In both puzzles, Black is to move.}
\centering
\includegraphics[width=0.45\textwidth]{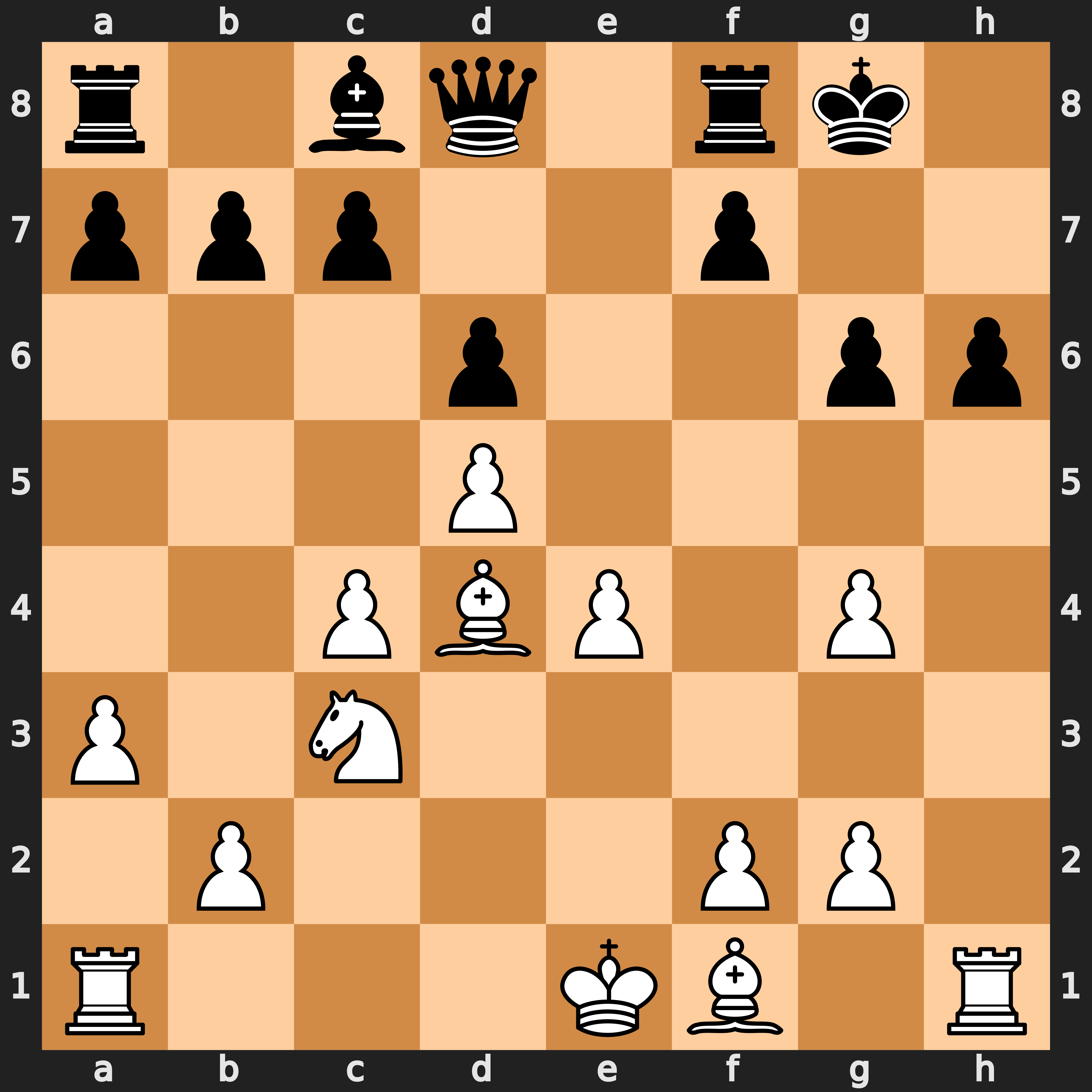}
\hspace{0.02\textwidth}
\includegraphics[width=0.45\textwidth]{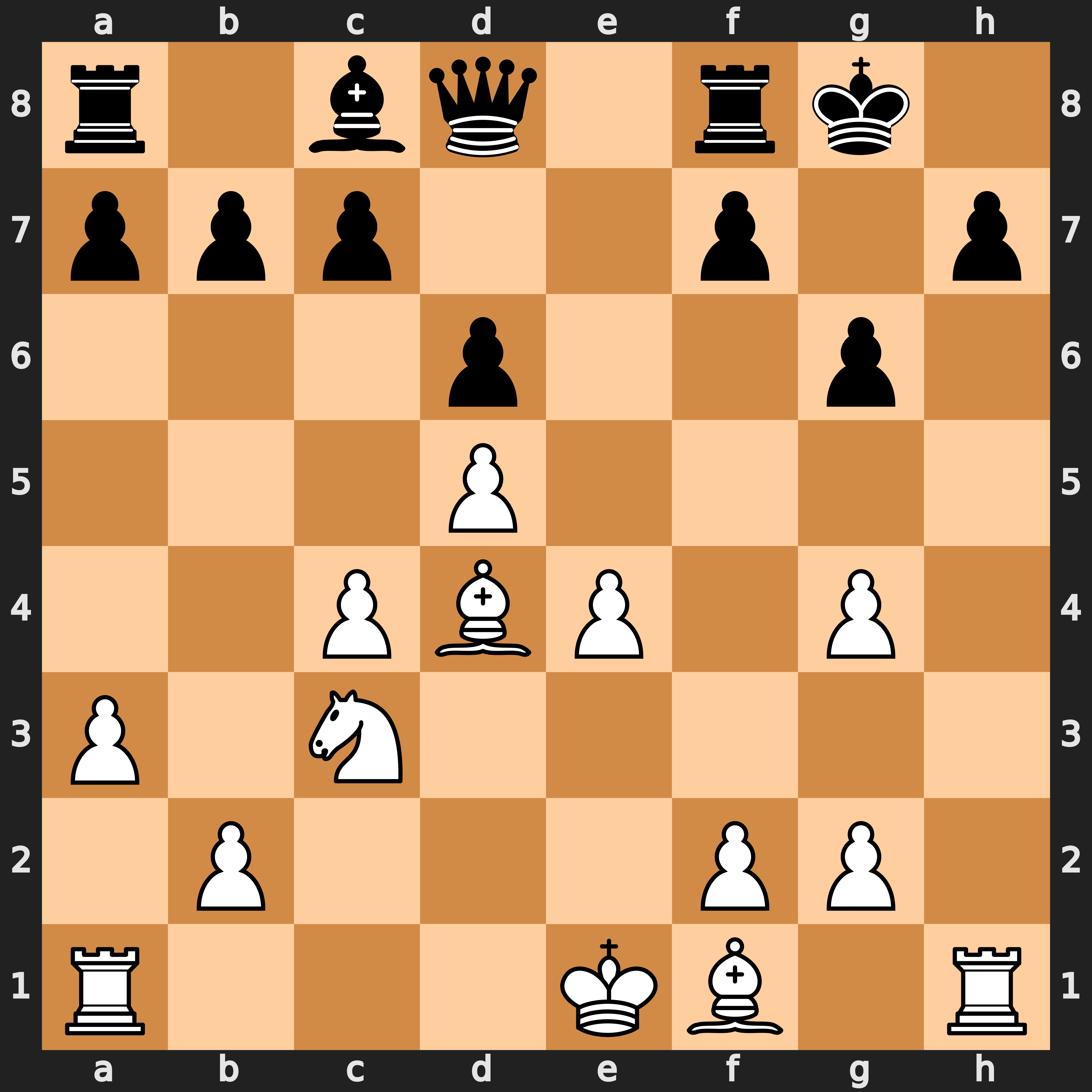}
\hspace{0.02\textwidth}
\label{fig:digging_deeper_ex2}
\end{figure}

Queen sacrifices are among the most beautiful (and rare) tactical motifs in chess, as they go against the established chess principles --  do not trade more valuable pieces (i.e., the queen) for less valuable pieces (a knight and bishop).
However, here AZ's queen sacrifice is strategic -- after sacrificing the queen, White continues developing their pieces.
The line continues \ct{13...Bxd4 14.Bxd4}, as shown in the left of Figure~\ref{fig:digging_deeper_ex2}. Here, due to the pawn on \ct{h6}, Black's king is vulnerable, and White is better. 
Therefore, it was critical to play \ct{9.Bg5}, rather than \ct{9.Be3}, to make the queen sacrifice feasible by creating this weakness. The puzzle on the right of Figure~\ref{fig:digging_deeper_ex2} arises in the same line if one opts for immediate \ct{9.Be3} instead, failing to induce \ct{h6} first. Without the pawn on \ct{h6}, White is lost, highlighting this is the critical positional element. 

In general, some moves in the AZ's calculations are more important for the concept than others.
In our convex optimisation formulation, we do not require the concept to be equally important for every chess position (see Equation~\ref{eq:tree_formula}). This is illustrated in the previous example, where $\ct{Bg5}$ is more important for the concept.

The next puzzle from the same concept, also given in Phase 1, is shown in Figure~\ref{fig:concept6_phase1_right}. This puzzle is of particular interest due to the unconventional plan of AZ -- it increases space on both sides of the board, expanding with the pawn move \ct{b4} while the king is still on \ct{b1}.
To this end, AZ initiates this plan with the prophylactic move \ct{21.Qd2}, preventing the immediate \ct{21...Ne4} by Black. 

Unlike AZ, the grandmaster's suggestion in this puzzle was \ct{21.g5}, with the intention of following up with \ct{21..Nh5 22.Bxh5 Rxh5 23.Rh3 Rdh8 24.Rdh1}. Upon seeing the suggested AZ line starting with \ct{21.Qd2}, however, the grandmaster study participant remarked
\begin{displayquote}
``\ct{21.Qd2} is a useful move as it stops Ne4 and protects f4 and can be better placed in case of b4 in the future. One curious line [given by AZ] is \ct{21...Rh7 [22.h5 Rdh8] 23.Rh3 gxh5 24.g5 Ng4} White can just play \ct{25.Rf1} and then focus on getting the b4 [pawn] break, which is not natural." 
\end{displayquote}

\begin{figure}[!ht]
\caption{Expanding on the critical line in the puzzle shown in Figure~\ref{fig:concept6_phase1_right}. White has just played \ct{b4}.}
\centering
\includegraphics[width=0.45\textwidth]{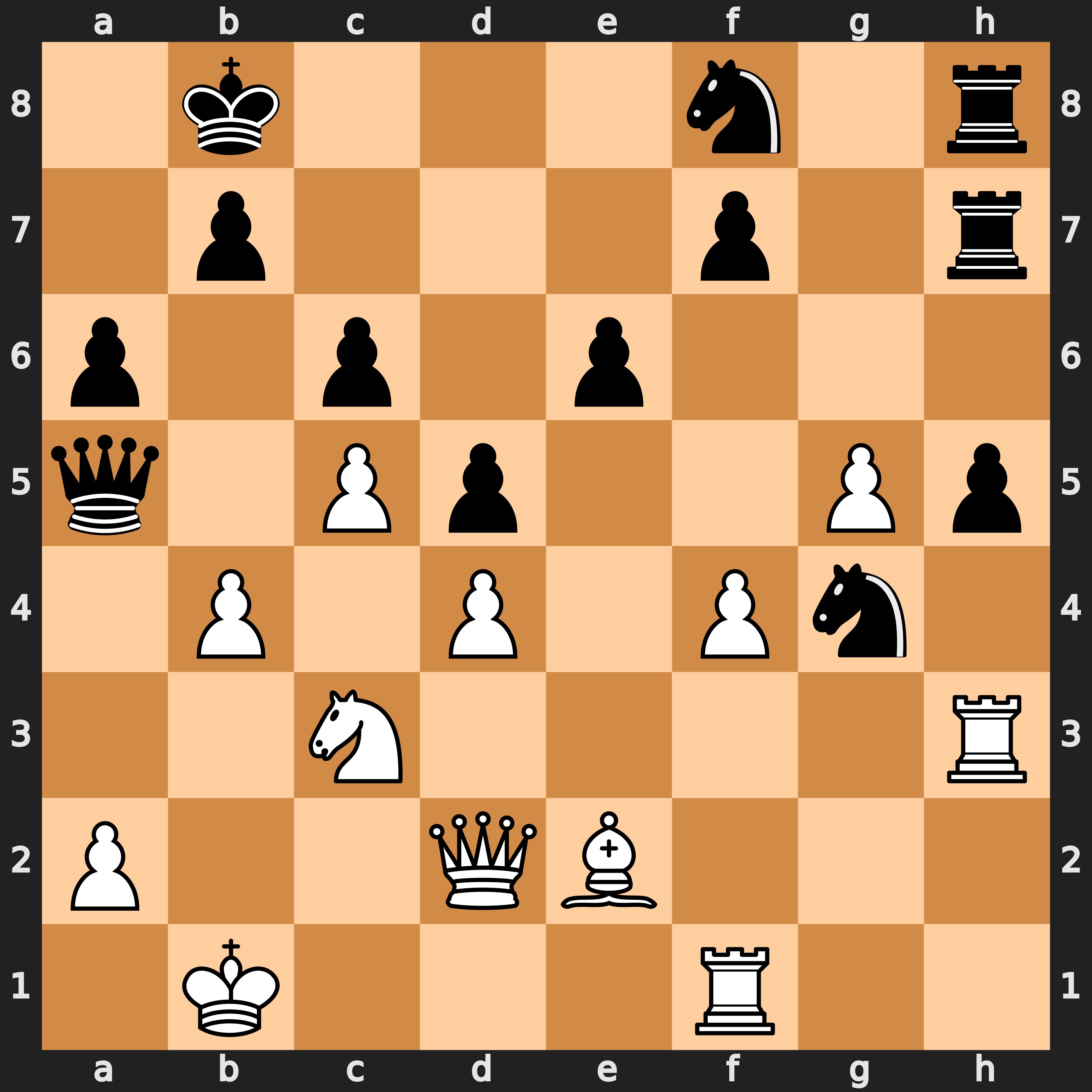} \\ 
\label{fig:qd2_b4}
\end{figure}

The unconventional plan of pushing the pawn to \ct{b4} with the king potentially exposed on \ct{b1} is particularly strong in this chess position (shown in Figure~\ref{fig:qd2_b4}) as it allows White to gain space and open up the chess position under unfavourable circumstances for Black, and claim an advantage. Therefore, the more general rules here are discarded based on concrete analysis. For example 
\begin{displayquote}
\ct{21.Qd2 Rh7 22.h5 Rdh8 23.Rh3 gxh5 24.g5 Ng4 25.Rf1!?~Nf8 26.b4!!~Qxb4+ 27.Ka1 Qa5 28.f5!~exf5 29.Qb2 Ne6 30.Nxd5!!~cxd5 31.Ra3 Qc7 32.Bxa6 White is better.}
\end{displayquote}
In this line, we see the dynamic play of AZ: the rooks from \ct{f1} and \ct{h3}, switch over to the b-file to attack Black's king. 

So far, the ideas in both positions were missed by the grandmaster. AZ's ideas require unconventional continuations that go against common human chess principles. Both of these observations hint at the existence of super-human knowledge $(M-H)$.  

\begin{figure}[!ht]
\caption{In the left puzzle, White is to move. In the right puzzle, Black is to move.}
\centering
\includegraphics[width=0.45\textwidth]{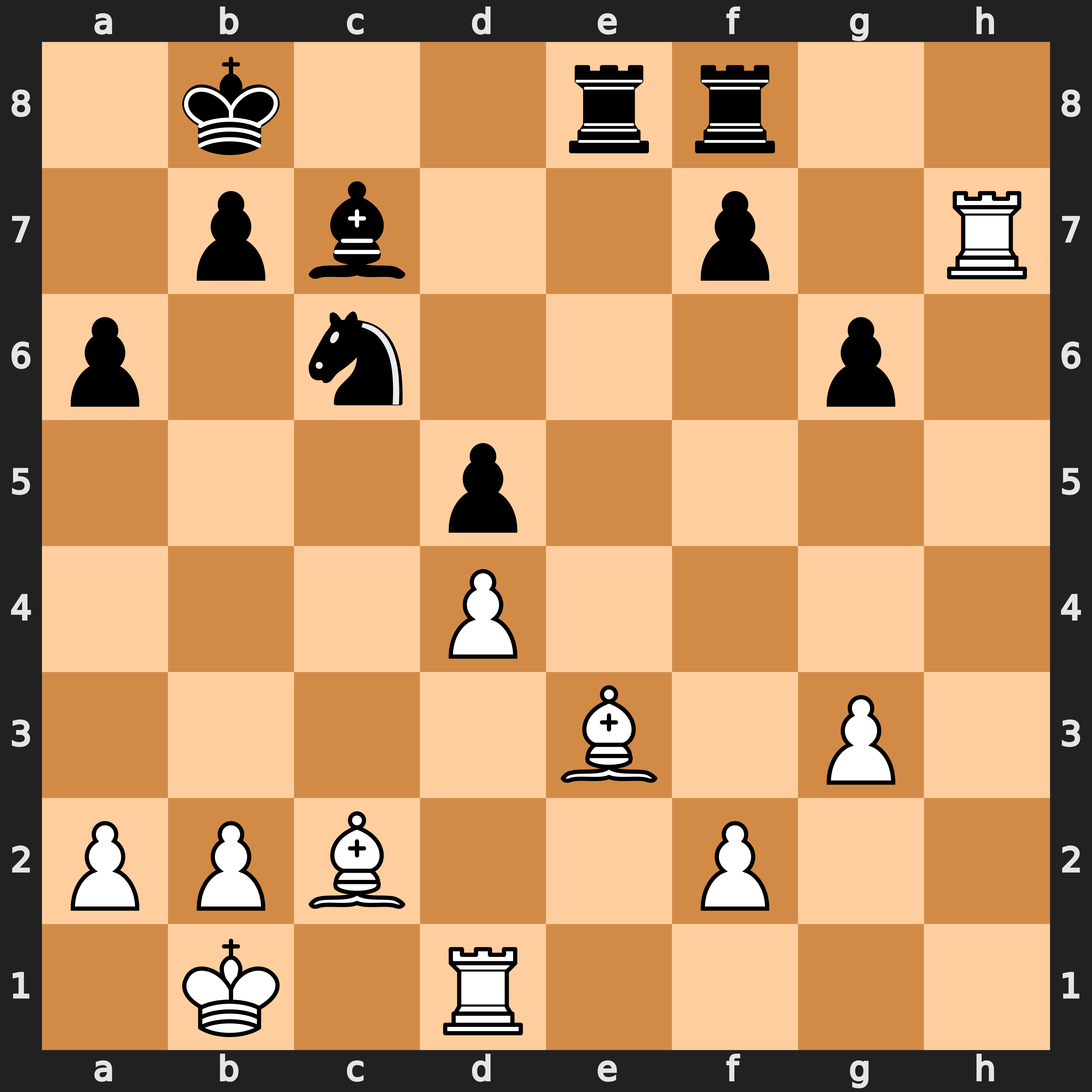}
\hspace{0.02\textwidth}
\includegraphics[width=0.45\textwidth]{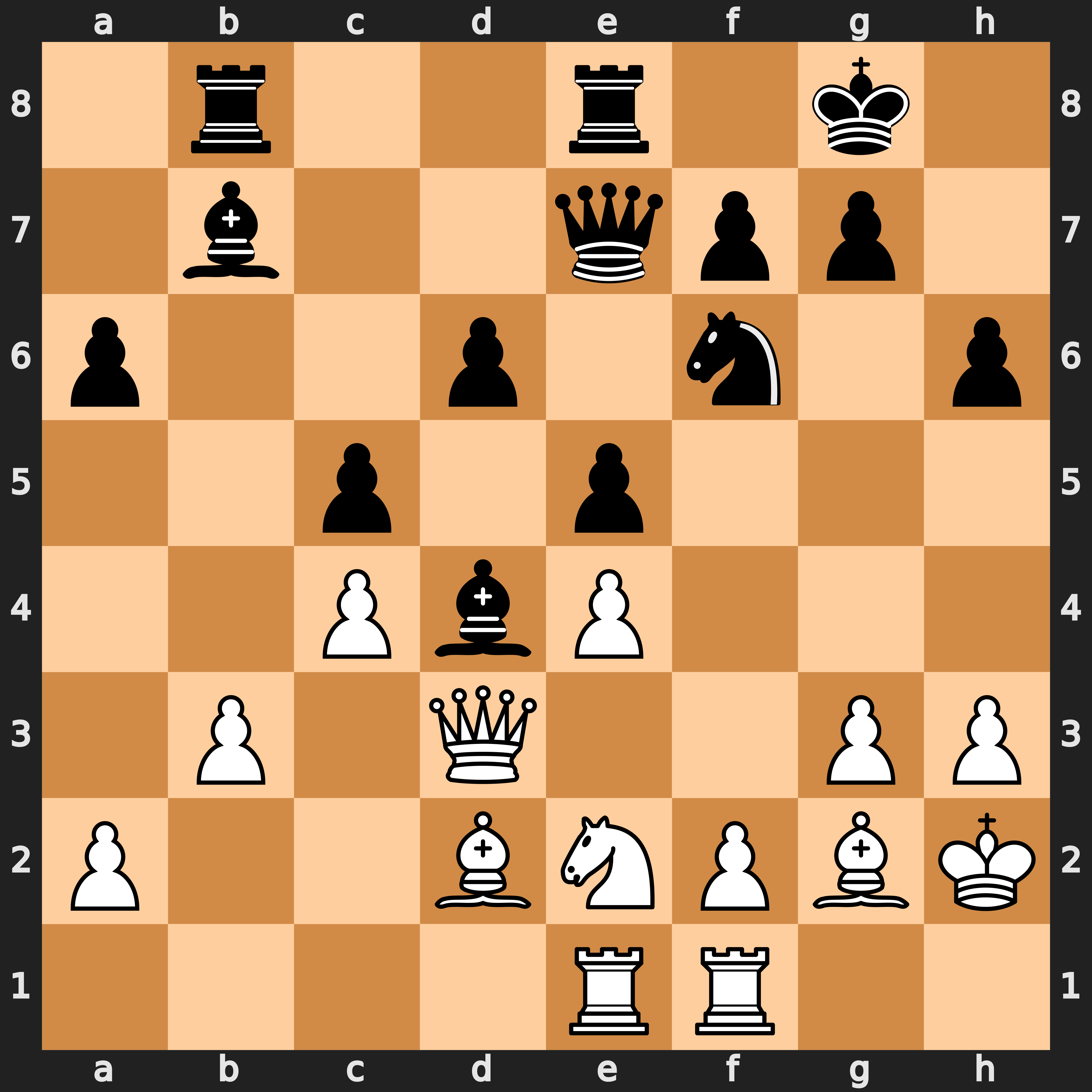}
\label{fig:concept_apply}
\end{figure}

Figure~\ref{fig:concept_apply} shows the puzzle from the same concept provided to the same grandmaster in Phase 3; Phase 3 tested whether the grandmasters had learnt the concept.
As before, the puzzles underscore the concept's multifaceted attributes, encompassing its prophylactic characteristics and its integration of tactical and strategic elements.

In the puzzle (from Phase 3) shown on the left of Figure~\ref{fig:concept_apply}, the grandmaster correctly found the move suggested by AZ: \ct{24.Bb3}, with the idea of forcing the Black rook on \ct{e8} into a more passive position (\ct{d8} to defend the pawn) prior to commencing activity on the other side of the board. 
The idea can be seen in a possible continuation: 
\begin{displayquote}
\ct{24...Rd8 25.Ba4 Na5 26.Rdh1 Nc4?~27.Bh6 and White picks up the pawn on f7.}
\end{displayquote}
As in the previous puzzles, AZ uses both sides of the board to optimise piece activity. 

In the puzzle, shown on the right in Figure~\ref{fig:concept_apply}, the grandmaster again found the correct idea: \ct{22...Bc6}, with the idea of preventing the critical pawn advance \ct{23.f4} because of:
\begin{displayquote}
\ct{23.f4 Qb7 24.Nc3 exf4 25.Rxf4, resulting in a weakened pawn structure,}
\end{displayquote}
where White cannot recapture with \ct{25.gxf4} as \ct{e4} is hanging. 
The move \ct{22...Bc6} is both prophylactic and tactical; it prevents White from executing their plan to advance the kingside pawns while improving Black's position by activating the bishop and rook. 

The overall improvement of the grandmaster on this concept suggests that they may have learnt AZ's concept, thereby expanding $H$ with $(M-H)$. Other examples of concept learning can be found in \S\ref{appx:proto_human}. 

\begin{figure}[!ht]
\centering
\caption{Graph of AZ's concept in Figures~\ref{fig:concept6_phase1_left}, \ref{fig:concept6_phase1_right} \ref{fig:digging_deeper_ex2} and \ref{fig:concept_apply} between AZ's (white), strategic (green) and Stockfish concepts (purple).
The information in the parentheses means the layer in which the concept is found, and w = white, b=black, eg= endgame, mg = middlegame, ph=phased. The edge color denotes the edge weight.}
\includegraphics[width=0.6\textwidth]{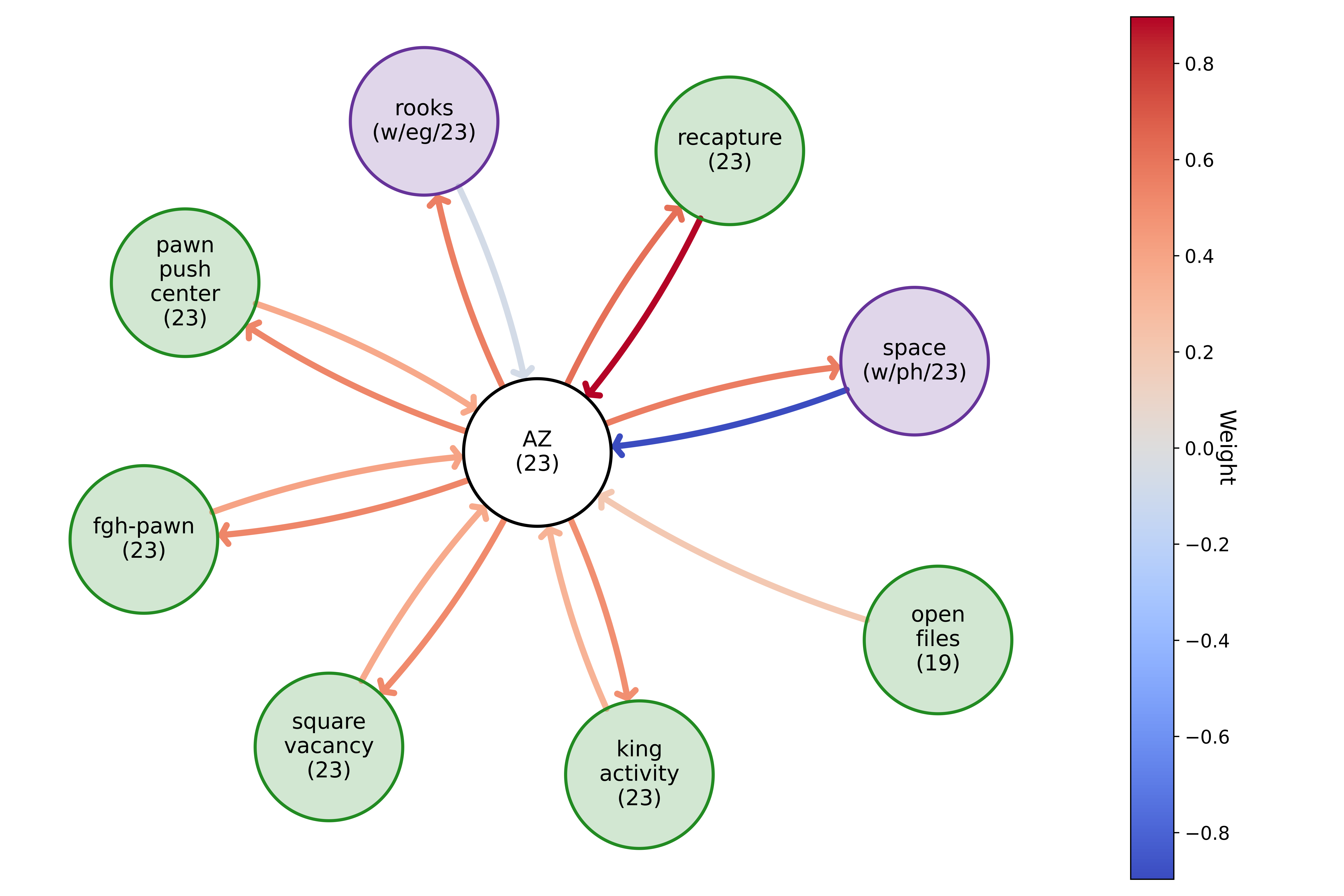}
\label{fig:graph}
\end{figure}

\paragraph{Understanding AZ's concept using graph analysis and human-labelled concepts.}
Using a simple way to learn a graph (see \S\ref{appx:graph_analysis}) between concept vectors, we discover strong relationships between existing and discovered concepts to gain further insight into the concept meaning (shown in Figure~\ref{fig:graph}).
Edge weight is influenced by (1) the strength of the relation between two concepts and (2) the frequencies at which concepts co-occur. 
Below, we further discuss the two concepts with the largest edge weights.

\textbf{Space.} AZ's concept has a strong positive weight on the outgoing edge with the (White-side) space concept. In puzzles where White is to move (Figures~ \ref{fig:concept6_phase1_left}, \ref{fig:concept6_phase1_right} and \ref{fig:concept_apply}), an important component of the plan is to increase space. 
In a similar vein, given that the idea is to \textit{increase} space, which is `easier'/more likely if the initial value is lower, AZ has a negatively weighted incoming edge with the concept space. 

\textbf{Recapture.} We observe positive incoming and outgoing edge weights with the recapture concept. Recall that we have dynamic concepts, which refer to a sequence of states. As such, we postulate that this connection is because the plan may be to recapture/gain material in the subsequent chess positions, as in the puzzles in Figures~\ref{fig:concept6_phase1_left} and \ref{fig:concept_apply} (left side). 

\subsubsection{Concept example: unsuccessful learning} 
\begin{figure}[!ht]
\caption{In both puzzles, White is to move.}
\centering
\includegraphics[width=0.45\textwidth]{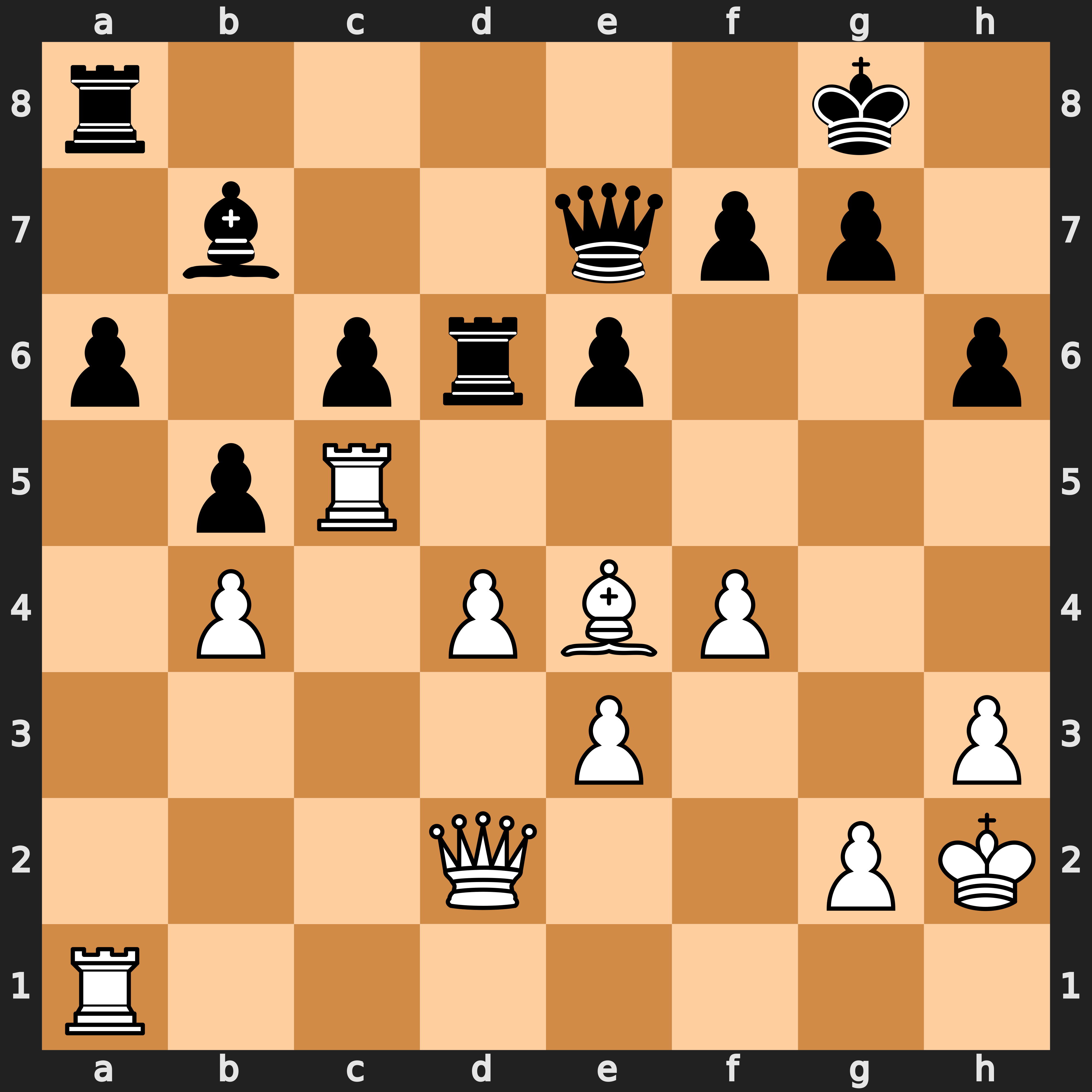}
\hspace{0.02\textwidth}
\includegraphics[width=0.45\textwidth]{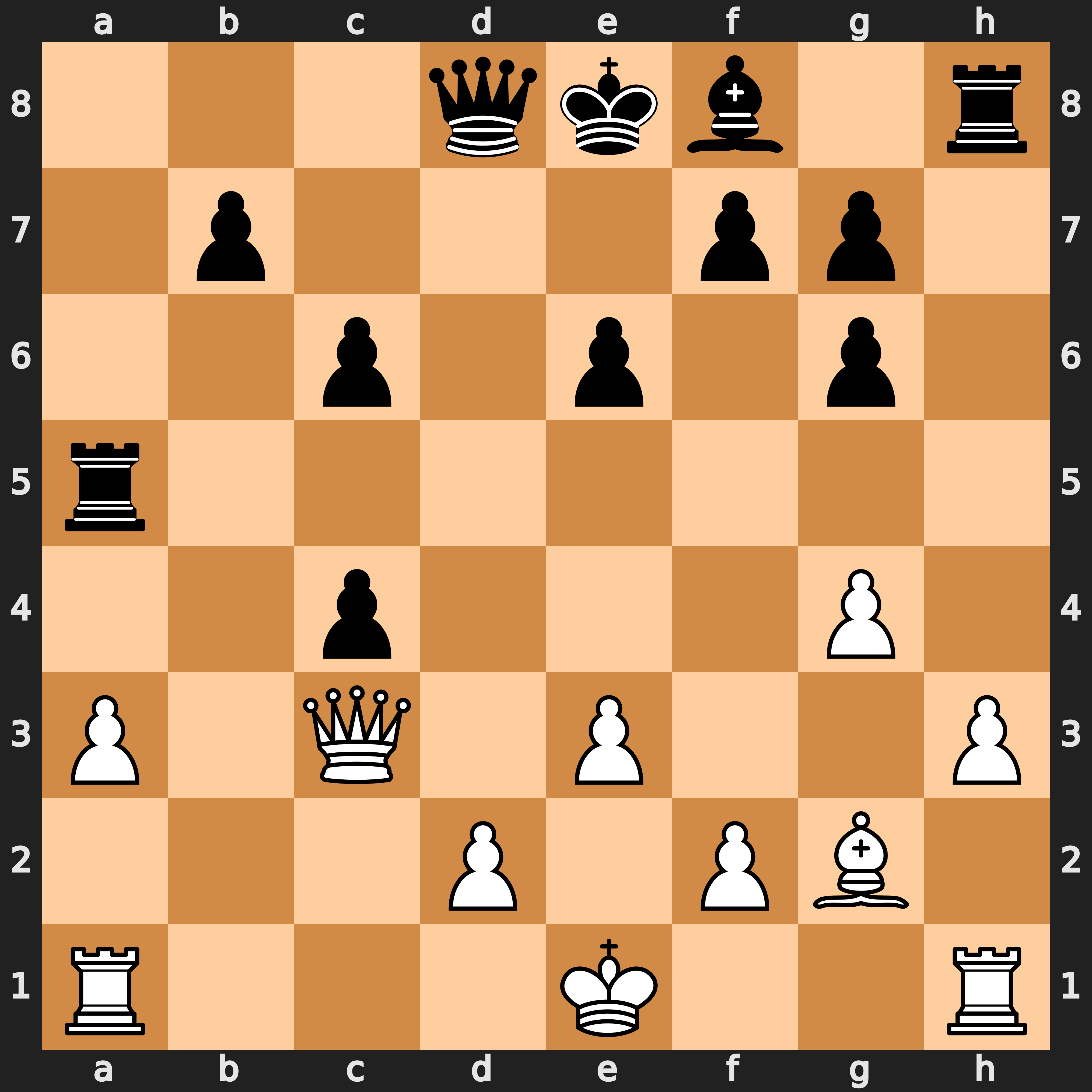} \\ 
\vspace{0.2cm}
\fbox{\begin{minipage}{\textwidth}
\begin{flushleft}
\textbf{AZ's calculations:} \ct{33.g4 g6 (33...f6 34.Qg2 Rf8 35.Rf1; 33...Rad8 34.Qf2 f6 35.Rg1) 34.Qf2 Kh8 35.h4 White is better}
\end{flushleft}
\end{minipage}}
\label{fig:space_lessmaterial}
\end{figure}

This concept is related to gaining and playing with a space advantage with positional advantage despite less material. 
In this section, we provide an example of when a grandmaster found the correct move in Phase 1, but provided an incorrect (i.e., not AZ's choice) move in Phase 3. 
The puzzle on the left side in Figure~\ref{fig:space_lessmaterial} was provided to the grandmaster in Phase 1, and they correctly chose the same move as AZ: \ct{g4}.
However, while the grandmaster found the correct move, there were further finesses in AZ's calculations that the grandmaster missed during their time-constrained analysis:
\begin{displayquote}
White is a pawn down, [and AZ plays the move I suggested] \ct{g4} ... The computer plays with \ct{h4}-\ct{h5} [and] \ct{g4} and the queen on \ct{f2} or \ct{g2}. White is playing on the kingside.\footnote{For readers - the queenside refers to the left side marked a-d and the kingside refers to the right side of the board marked e-h.} Black has no active moves, is playing \ct{Rd8} or \ct{[R]f8}, \ct{Kh8}. Seems convincing to me. On \ct{g6} [AZ] goes \ct{[Q]g2} which is nice, it didn’t occur to me. I was mainly focused on making \ct{h4} work for White. White is not in a hurry, will at some point play \ct{g5}. Compensation, zero counterplay, and AZ is acting on these premises.
\end{displayquote}

As remarked by the grandmaster, AZ is a pawn down, however, Black's pieces, particularly the Bishop on \ct{b7}, are placed passively. Instead of prioritising regaining material, AZ focuses on improving the kingside position. While the grandmaster understood the general plan, they missed the intricate idea \ct{Qg2}, and slowly advancing the g and h pawns. 

The puzzle on the right in Figure~\ref{fig:space_lessmaterial} was provided to the grandmaster in Phase 3. 
Similar to the previous example, AZ focuses on space rather than recapturing material. It continues with the move \ct{18.a4}, which was rejected by the grandmaster on account of \ct{18.Rc5}, where Black tries to maintain the material advantage. 
However, AZ finds the rook on c5 misplaced and continues
\begin{displayquote}
\ct{18.a4 Rc5 19.a5 Be7 20.a6 where after Bf6?~White has 21.d4}
\end{displayquote}
The idea behind the pawn advance is to weaken Black's pawn structure.  

Overall, the evidence suggests that the grandmaster did not learn this concept. Here, 
AZ selects stronger moves due to its prioritisation of concepts (e.g., focusing on space and activity).
Humans tend to prioritise these concepts differently (e.g., prioritising bringing the king to a safe location as soon as possible). This concept may be inherently difficult and require further examples to learn. 

\begin{figure}[!ht]
\centering
\caption{Graph of AZ Concept in Figure~\ref{fig:space_lessmaterial}  between AZ's (white), strategic (green) and Stockfish concepts (purple).
The information in the parentheses is the layer number in which the concept is found. w = white, b=black, eg= endgame, mg = middlegame, ph=phased. The edge color denotes the edge weight.}
\includegraphics[width=0.6\textwidth]{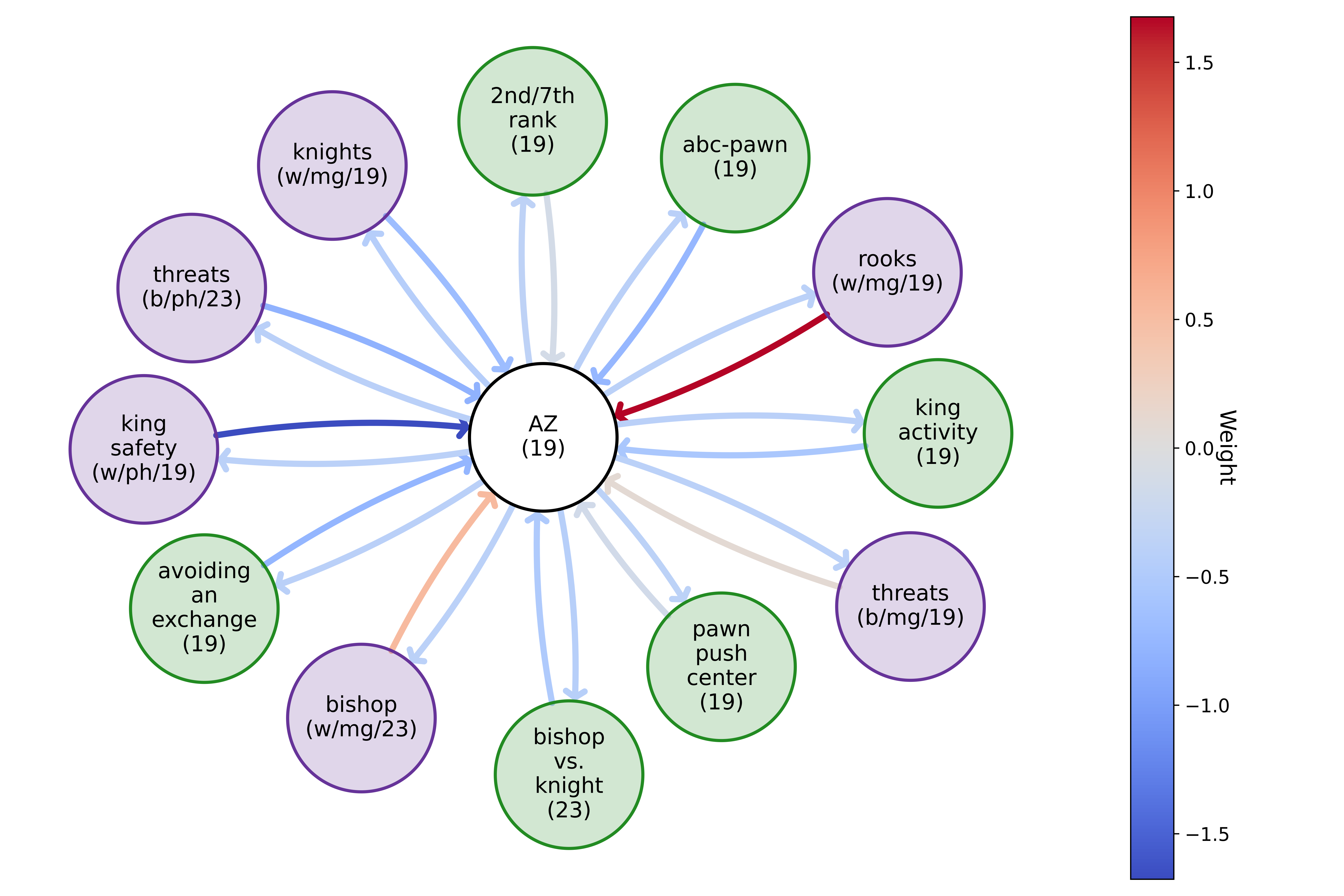}
\label{fig:graph_3}
\end{figure}

\paragraph{Understanding AZ's concept using graph analysis and human-labelled concepts.} 
Figure~\ref{fig:graph_3} shows the relationship between AZ's concept and high-quality human-labelled concepts. The graph is dense, and we elaborate on the two concepts with the largest edge weight. 

\textbf{Rook.} AZ's concept has an incoming positive edge with the rook (activity) concept. In the puzzles in Figure~\ref{fig:space_lessmaterial}, we observe that white White has active rooks (the rooks on \ct{a1} and \ct{c5}, in the left chess position) or plans to activate the rook (the rook on \ct{a1}, in the chess position on the right).

\textbf{King.} AZ's concept has an outgoing negative edge with king safety. In the chess positions in Figure~\ref{fig:space_lessmaterial}, the king is less safe than usual. In the left chess position in Figure~\ref{fig:space_lessmaterial}, AZ pushes forward the pawn to \ct{g4} (and later the pawn to \ct{h4}) around the king -- thereby removing some of the king's defenders. In the right position, White does not castle to improve the king's safety but instead leaves the king in the centre. 

Further examples of concept puzzles can be found in Section \ref{appx:proto_human}.
\subsubsection{Differences between humans and AZ} \label{sec:diff_human_az}

In this section, we share a few observed differences between the grandmasters and AZ and speculate where they come from. While we do not have definitive answers, the discussion may lead to further research. 

The qualitative examples suggest that AZ has different priors over the relevance of concepts in a chess position than humans. 
Human chess players formulate and adopt heuristic chess principles to inform their analysis, predisposing them to biases that influence which concepts they deem relevant for specific chess positions. An example is the three `golden rules' of the opening: control the centre, develop your pieces, and bring your king to safety~\citep{openings_carsten, step_method, 10goldenrules}.
Consequentially, in opening, humans may focus on moves that align with these guidelines. 
Instead, AZ is self-taught and does not seem to have the same priors over chess concepts as humans. 
We believe this lack of prior allows AZ to be more flexible -- it can apply concepts to various different chess positions and change plans quickly.
In essence, AZ formulates its own priors over the relevance of chess concepts for a given chess position.
Examples of this behavior are that AZ plays over the entire board, as opposed to focusing on a specific side (see, e.g., Figures~\ref{fig:teaching_succ_1}, \ref{fig:followup_left}, \ref{fig:concept_apply}, and \ref{fig:qc1_app}); places less importance on the material value of pieces, and prioritises space and piece activity (see, e.g., Figures~\ref{fig:digging_deeper_ex2} or \ref{fig:space_lessmaterial}).
This may result in the super-human application of concepts, and new concepts.  

\begin{figure}[!ht]
\caption{AZ simplifies the chess position for the draw whereas humans would continue to try to win. Black is to move.}
\centering
\includegraphics[width=0.5\textwidth]{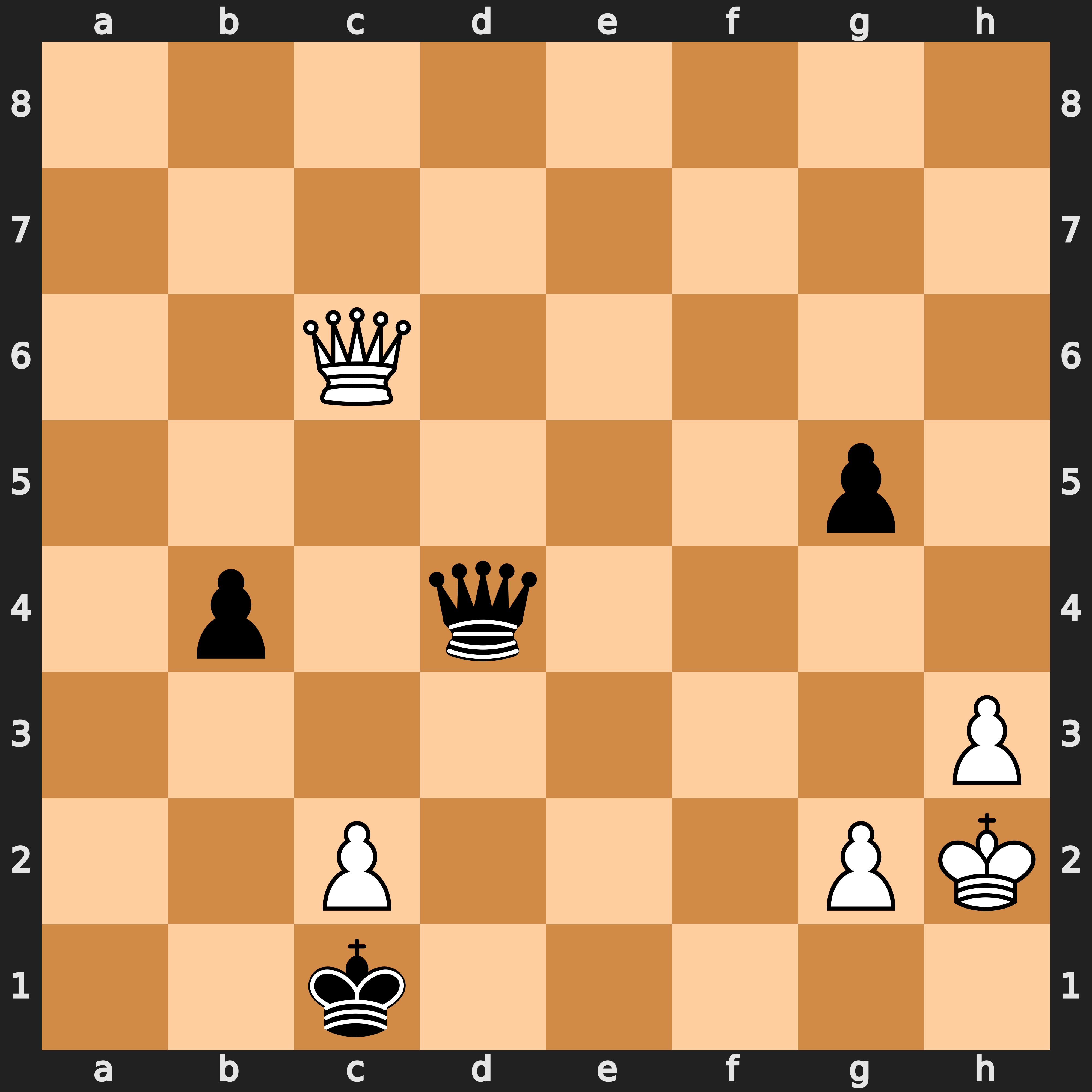} \\
\vspace{0.2cm}
\fbox{\begin{minipage}{\textwidth}
\begin{flushleft}
\textbf{AZ's calculations:} \ct{58...g4 59.hxg4 (59.Qc7 gxh3 60.Kxh3 Qc3+) 59...Qxg4 60.Qc5 Qh4+ 61.Kg1 Qe1+ 62.Kh2 Qc3 (62...Qh4+) Draw}
\end{flushleft}
\end{minipage}}
\label{fig:g4}
\end{figure}

One may ask where do the differences between AZ's and humans' play come from? We conjecture that they may arise from differences in objectives and capabilities.
AZ learnt to play chess against itself. As such, AZ assumes optimal play and information symmetry.\footnote{With information symmetry, we mean that Black and White have the same general knowledge and perform the same calculations in a given chess position.}
On the other hand, humans play chess against other humans and, therefore, may assume information asymmetry and imperfect play. 
This leads to a difference in behaviour: while AZ focuses on finding the best move, human chess players often make \textit{practical} choices. 
Humans' choices do not always increase the expected outcome against an optimal opponent (their choices may even slightly decrease the expected outcome) but may increase their odds against another human.
For example, in drawn chess positions, humans may try to complicate the chess position or opt for continuations where the best moves are less clear-cut in hopes that their opponent makes a mistake (see Figure~\ref{fig:g4}). 
However, AZ will try to find the optimal plan, disregarding aspects such as complexity. Therefore, AZ's play may be fundamentally different and better reflect conceptually relevant plans in a chess position.  

\begin{figure}[!ht]
\caption{White is to move.}
\centering
\includegraphics[width=0.5\textwidth]{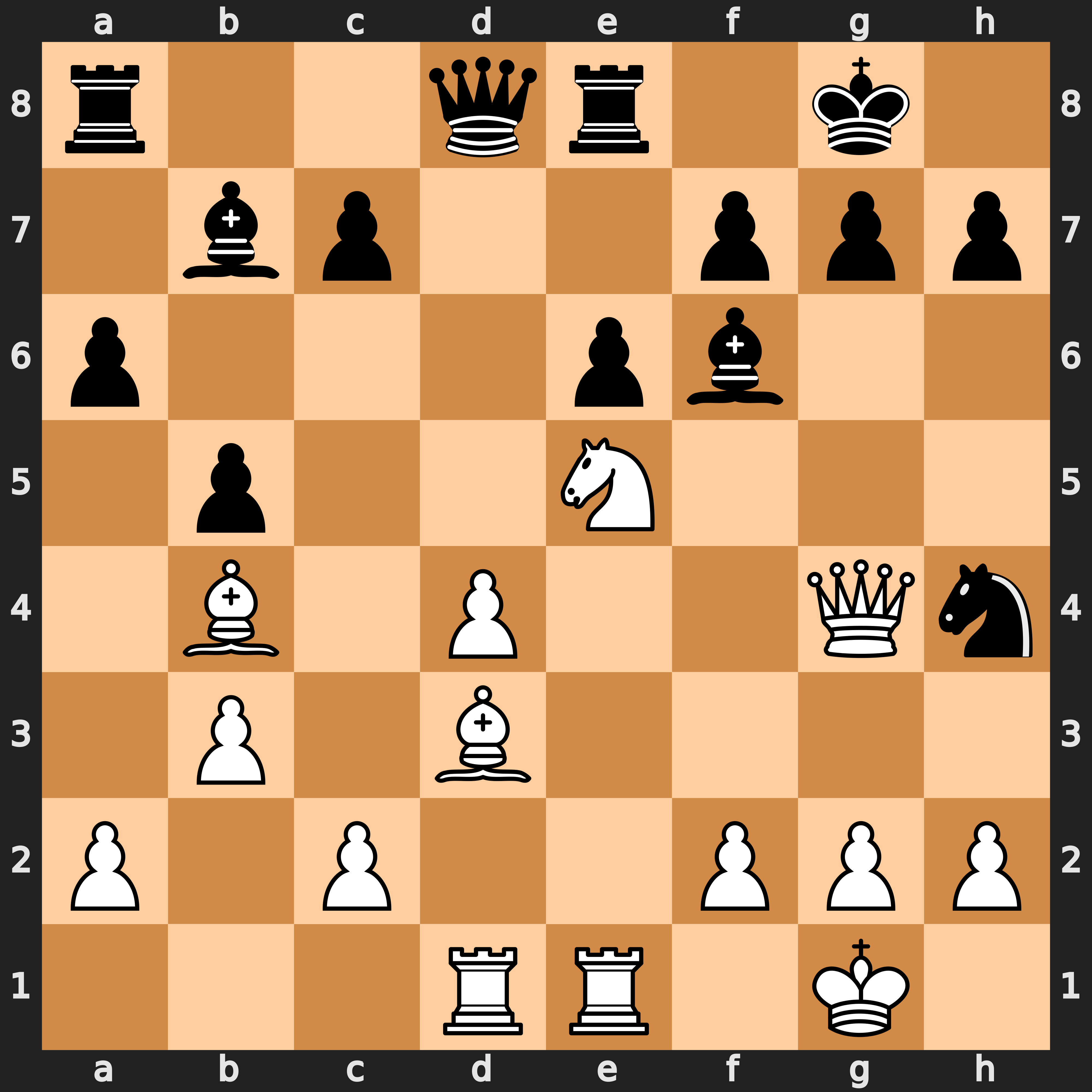} \\
\vspace{0.2cm}
\fbox{\begin{minipage}{\textwidth}
\begin{flushleft}
\textbf{AZ's calculations:} \ct{18.f3 (18.Bf1 Ng6 (18...Nf5 19.c4 g6 20.cxb5 axb5 21.Bxb5 h5 22.Qh3 Rxa2 23.Bxe8 Qxe8) 19.Bd3 Nh4 (19...Bh4 20.Bxg6 hxg6 21.Rd3) 20.f3) 18...Ng6 19.Bc5 Bh4 20.Bxg6 hxg6 21.g3 White is better}
\end{flushleft}
\end{minipage}}
\label{fig:f3}
\end{figure}

Another difference between humans and AZ's play is the role of time. 
Humans have limited energy\footnote{Classical time control ~\citep[see match time controls in][]{fide_handbook} chess games take hours.} and time allocation for a game.
In chess positions where humans are better, they often simplify the chess position to try to secure the win as quickly as possible and minimise risk (see, e.g., the grandmasters' chosen moves in Figures~\ref{fig:risk} and \ref{fig:f3}). 
However, AZ does not care about how quickly the game finishes. The training loss function does not have a penalty term to encourage winning as quickly as possible. As a result, it has a different treatment of time. This results in sometimes choosing slow strategic wins (as can be seen in the chess positions in Figure~\ref{fig:f3}). 
While the lack of time constraint may lead to super-human concepts, it also may result in 
complex concepts that are difficult for humans to learn.  

Naturally, AZ and humans have different computational capacities. As a result, AZ can opt for more computationally expensive moves and, therefore, defend complicated chess positions where humans might be more hesitant. 
In terms of playing style, AZ will often opt for what it believes is the \textit{optimal} move, while humans have a more limited computing budget and may opt for a safer move. In chess, \textit{safer} is used to describe continuation where there is less probability of making an incorrect move. Sometimes, humans may even play a slightly suboptimal move to minimise the risk. We see this phenomenon in Figures~\ref{fig:risk} and \ref{fig:g4}. 
While computational capacity cannot be transferred, it may still lead to super-human concepts, as AZ can find new ideas that can still be taught to humans.

\section{Conclusion} \label{sec:conclusion}
Our research represents a first step toward understanding the potential of human learning from Artificial Intelligence (AI). In this work, we focused on AlphaZero (AZ) -- an AI model that learned to play chess at a super-human level through self-play without prior knowledge or human bias. 
Through spectral analysis, we show that AZ's games encode features that are not present in human games, providing evidence for the existence of super-human knowledge.
To extract knowledge from AZ's representational space, we developed a framework to uncover new chess concepts in an unsupervised fashion.
We discover unsupervised concepts by leveraging AZ's training history to curate a set of complex chess positions.
We ensured each concept was informative, by verifying that the concept can be taught to another AI agent, and novel, through a spectral analysis of human and AZ games.
Communicating novel concepts requires a
common language between humans and AI. We bypass the need for this language by creating puzzles for each concept.

We collaborated with four world-top grandmasters to (1) validate the human capacity to comprehend and apply these concepts by studying AZ's concept prototypes and (2) improve our understanding of the differences between AZ's and humans' chess representation space. All four grandmasters improved their performance after learning concepts compared to baseline performance.
We speculate that the differences between AZ and humans may stem from (1) prior biases over concepts, including their perceived applicability, importance, and how they can be combined with other concepts. 
For example, AZ shows a reduced emphasis on factors such as material value and is more agile in switching between playing on different sides of the board.
(2) a difference in the motivation and objectives when playing chess; AZ is trained to accurately evaluate the current chess position, assuming optimal play.

There are several aspects of the work that could be further explored.
In our work, we found a subset of all possible concepts. For example, we limited our investigation to linear sparse concept vectors. However, other concepts may be discovered in the form of non-linear vectors.
Additionally, the current work focuses on finding a single concept to explain a plan. However, a plan may contain multiple concepts. As such, an interesting aspect to further explore is how these concepts relate to each other and influence the plan. 

Further work could also explore the optimal conditions for humans to learn novel concepts. 
We allotted a fixed time budget for grandmasters to assimilate the concepts. However, it is plausible that an unlimited time budget could yield more profound and more intricate insights.
In our research, we provided grandmasters with part of AZ's Monte Carlo Tree Search (MCTS), in which the rollouts are motivated by the concept, as an explanation for the concept. We used this approach to keep the explanations as familiar and simple as possible.
Nonetheless, it would be interesting to augment this phase with an interactive component: e.g., for each puzzle, humans can actively engage with AZ by playing moves and \textit{asking} AZ what its response is. This interactive element would allow humans to investigate counterfactual scenarios, allowing for a deeper understanding why AZ did not select their solutions or approaches.

\section*{Acknowledgements}
We would like to thank the four grandmasters who participated in our study -- Vladimir Kramnik, Dommaraju Gukesh, Hou Yifan, and Maxime Vachier-Lagrave.
Without them, this work would not have been possible. 
We would also like to thank Tom Zahavy, Adam Pearce, Kevin Waugh, Julian Schrittwieser, and Blair Bilodeau for their help, discussions and feedback on this work.

\newpage 
\bibliographystyle{apalike}
\bibliography{references.bib}

\newpage

\section{Appendix}

\subsection{Human experiments: more concept puzzle examples} \label{appx:proto_human}
In this section, we provide more examples of concept puzzles.

\subsubsection*{Additional concept example: positive knowledge transfer}
On a high level, this concept appears to be intrinsically related to centre control and improving piece activity. However, a more detailed analysis unveils a nuanced dimension to this concept, as AZ leverages unconventional manoeuvres to achieve these goals.
The grandmaster improved performance by +2/4 between Phases 1 and 3, suggesting that this concept was not part of their existing knowledge and is human-learnable.

\begin{figure}[!ht]
\caption{Concept Puzzle 1: Black is to move.}
\centering
\includegraphics[width=0.45\textwidth]{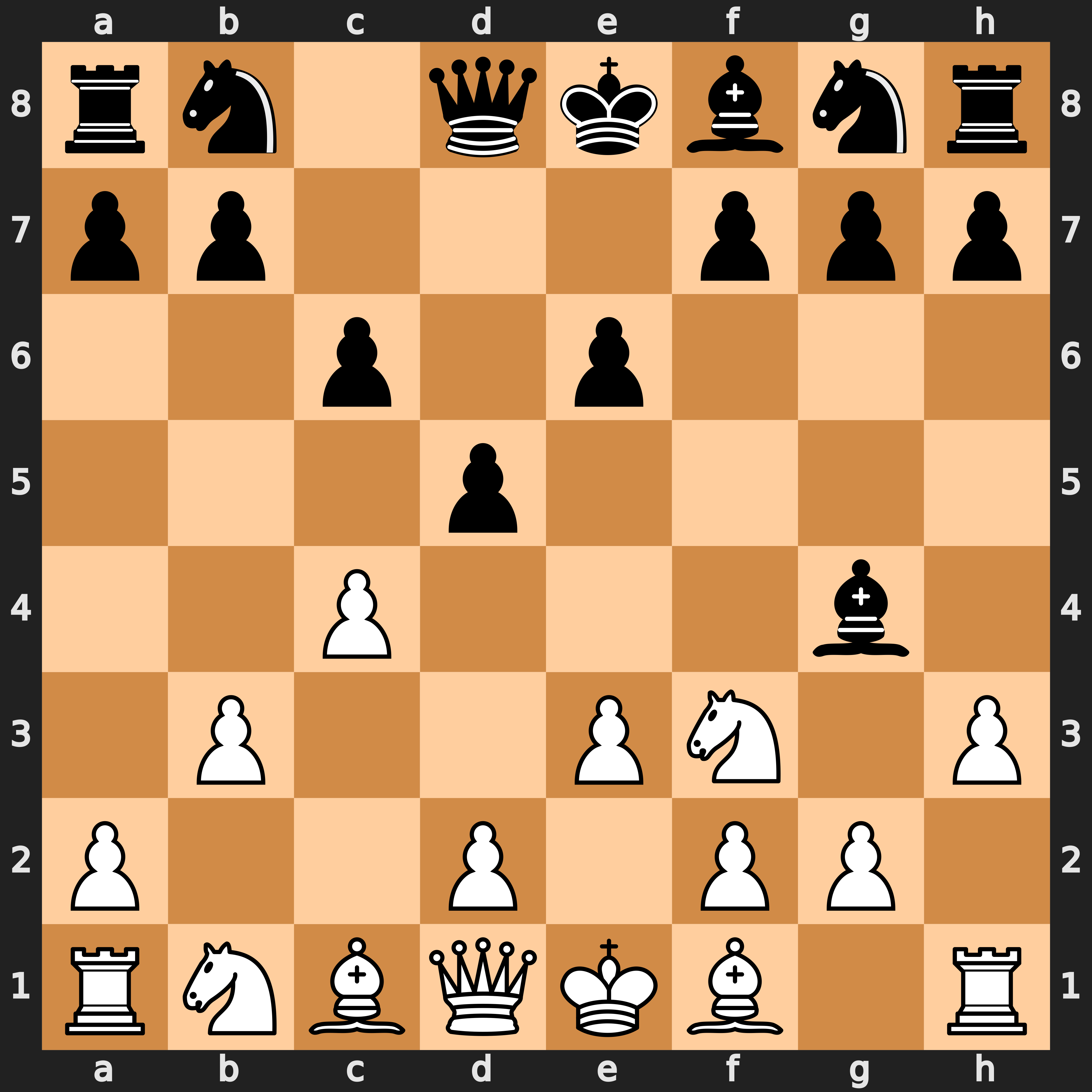} \\
\vspace{0.2cm}
\fbox{\begin{minipage}{\textwidth}
\begin{flushleft}
\textbf{AZ's calculations:} \ct{5...Bf5 (5...Bh5 6.Bb2 Nd7 7.cxd5 cxd5 8.Nc3 a6 9.Rc1 Ngf6 10.g4 Bg6 11.Nh4 Be4 12.Nxe4 Nxe4 13.Nf3) 6.Nc3 h6 7.Bb2 Nf6 8.d4 Bb4 9.a3 Bxc3+ 10.Bxc3 Ne4 11.Rc1 Nxc3 12.Rxc3 Qe7 13.Qc1 O-O 14.Be2 Nd7 Approximately equal}
\end{flushleft}
\end{minipage}}
\label{fig:teaching_succ_1}
\end{figure}

Figure~\ref{fig:teaching_succ_1} shows a concept puzzle that was shown in Phase 1. Here, AZ plays the move \ct{5...Bf5} to control the square e4. 
The grandmaster chose \ct{5...Bh5} while also considering the moves \ct{5...Bxf3} and \ct{5...Qf6}. 
After seeing the solution, the grandmaster commented 
\begin{displayquote}
``\ct{5...Bh5} line looks quite natural ... [however, AZ's move] \ct{5... Bf5} with the concept in mind \textbf{is very interesting} as after \ct{8.d4} [ the continuation for Black of] \ct{Nbd7}-\ct{Bd6} is more natural but \textbf{\ct{Bb4} is something new}. I was curious about the idea after \ct{11.Bb2 Nd7 12. Bd3} where \ct{h5!?} was probably the point.''
\end{displayquote}
We explore the question posed by the 
grandmaster -- what happens after \ct{11.Bb2}? The ideas is
\begin{displayquote}
\ct{5...Bf5 6.Nc3 h6 7.Bb2 Nf6 8.d4 Bb4 9.a3 Bxc3+ 10.Bxc3 Ne4 11.Bb2 Nbd7 12.Bd3 h5 13.0-0 g5!} 
\end{displayquote}

\begin{figure}[!ht]
\caption{Digging Deeper into the Concept in Figure~\ref{fig:teaching_succ_1}. In both positions, White is to play.}
\centering
\includegraphics[width=0.45\textwidth]{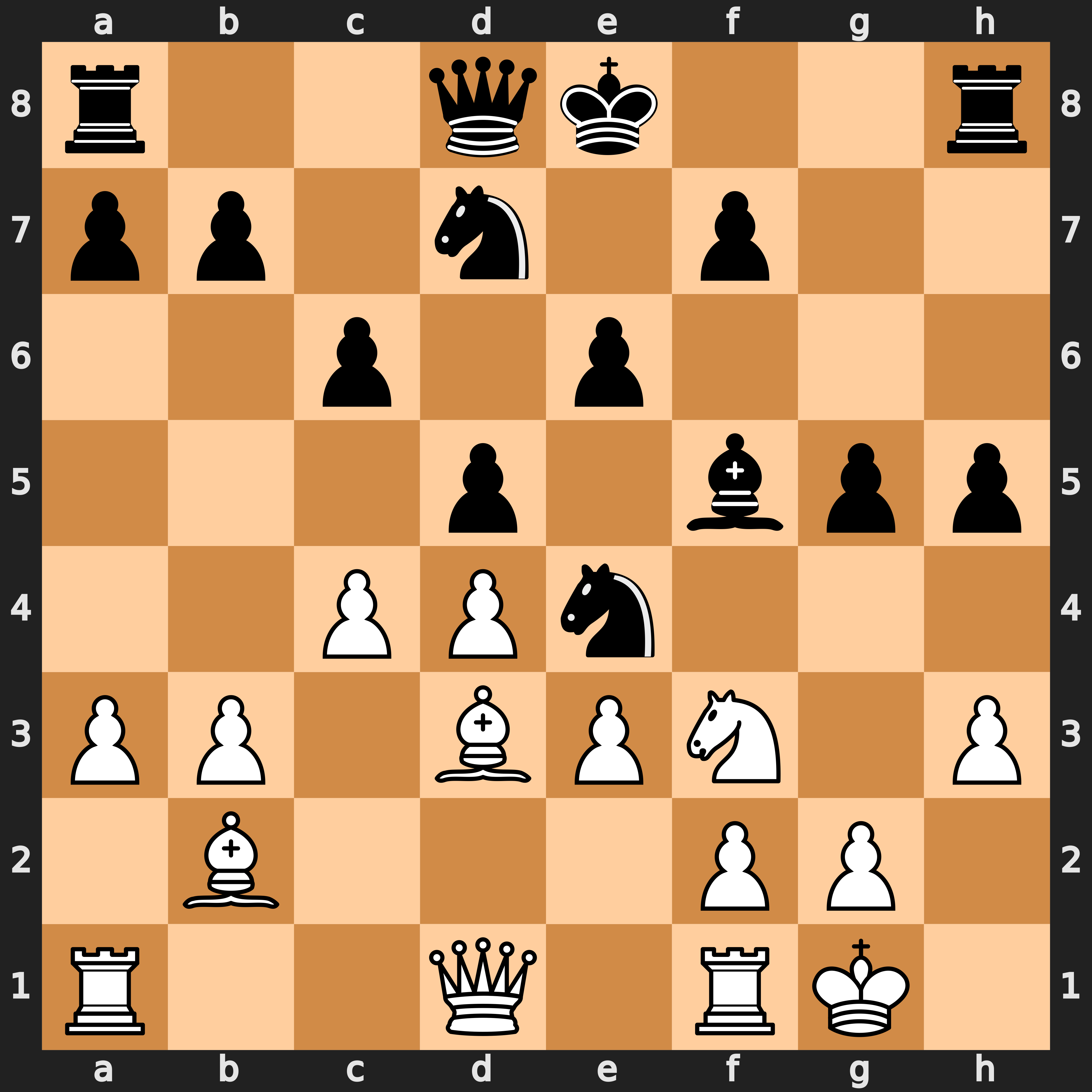}
\hspace{0.02\textwidth}
\includegraphics[width=0.45\textwidth]{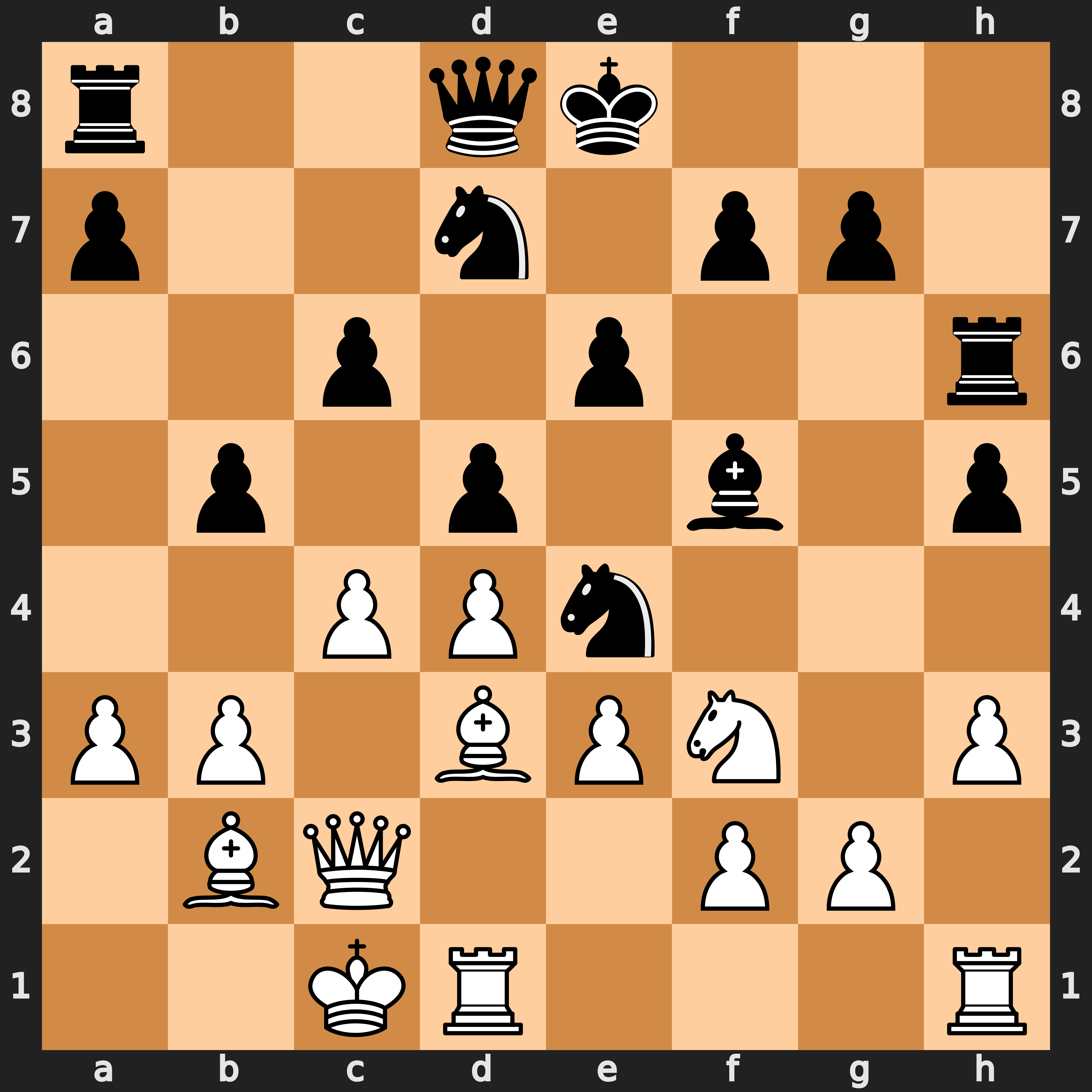}
\label{fig:followup_left}
\end{figure}

This resulting chess position is shown on the left of Figure~\ref{fig:followup_left}. If White tries to castle kingside with \ct{13.Qd2} instead of \ct{13.0-0}, Black can pursue a king-side pawn advance:
\begin{displayquote}
\ct{13.Qc2 Rh6 14.0-0-0 b5!?}.
\end{displayquote}
The resulting chess position is shown in the right chess position in Figure~\ref{fig:followup_left}.
Both continuations are unorthodox; conventional human-designed chess principles emphasise completing piece development, securing the king's safety and maintaining the bishop pair over trading it for a knight, as outlined in \cite{10goldenrules}. However, AZ deviates from these principles favouring a continuation that prioritises a strong control of the centre, space, and piece activity. 

\begin{figure}[!ht]
\caption{Concept Puzzle 2: White is to play.}
\centering
\includegraphics[width=0.45\textwidth]{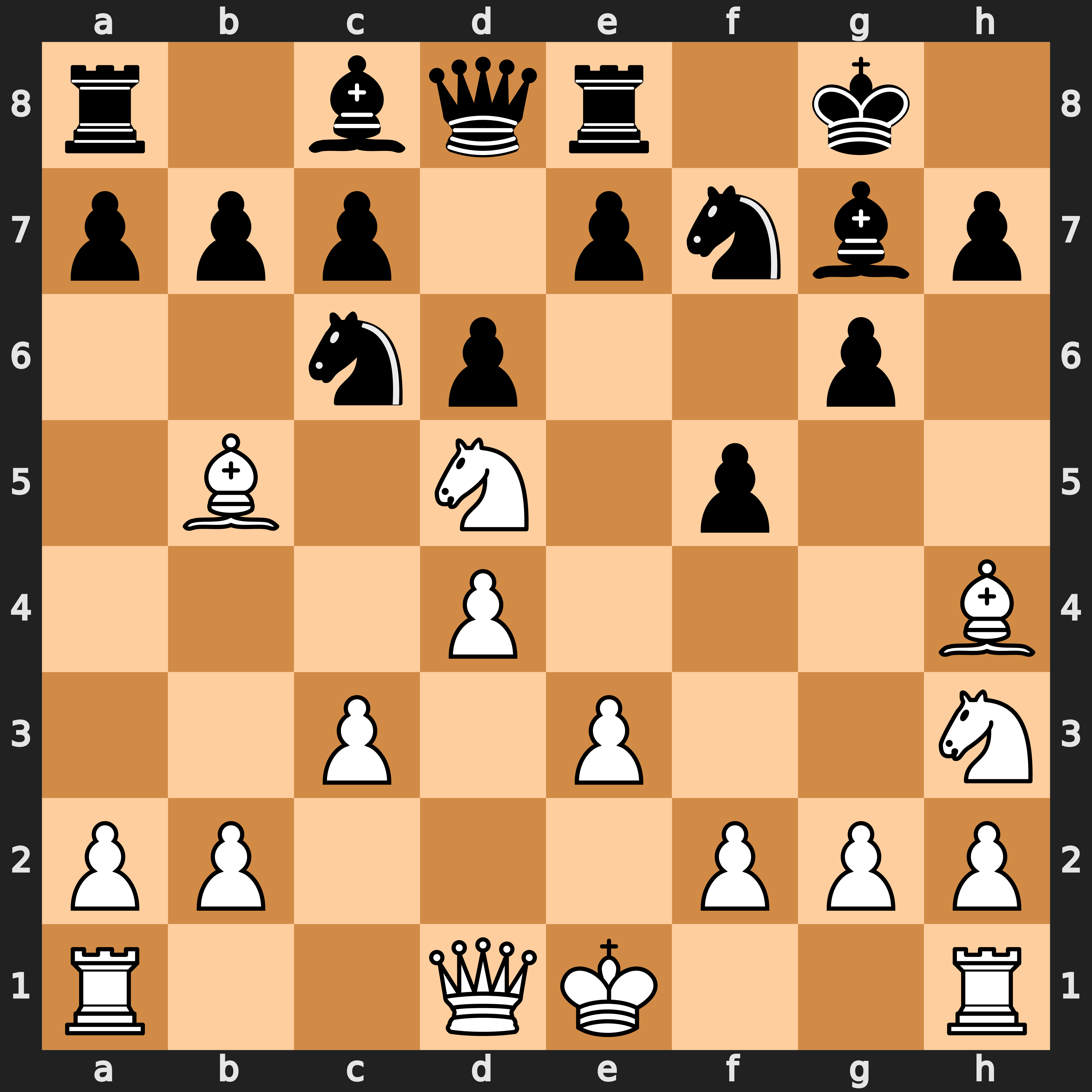}
\label{fig:teaching_succ_1_phase3}
\end{figure}

Another puzzle from the same concept is shown in Figure~\ref{fig:teaching_succ_1_phase3}, given in Phase 3.
Here, the grandmaster found the best continuation according to AZ: \ct{10.Ndf4} threatening \ct{d5}. The ideas are
\begin{displayquote}
\ct{10...a6 11.Qa4 \\ 
10...Bd7 11.Bc4 gaining control over the square e6 \\
10...d5?~11.Qb3 and the pawn on d5 is lost \\ 
10...Bf6 Black's best option 11.Bxf6 exf6 (12.d5?~a6 13.Qa4 then Black has the intermediate move 13...Re4!)~12.Ng1 a6 13.Ba4 Bd7 14.Nge2}
\end{displayquote}
The knight manoeuvres \ct{10.Ndf4} and \ct{12.Ng1} are against the common rules which advocate for finishing piece development and bringing the king to safety, above further improving a developed piece~\citep{10goldenrules, step_method, openings_carsten}. As in the previous puzzle, AZ prioritises controlling the centre and piece activity.

The grandmaster missed the idea {12.Ng1}, although did appreciate it remarking that ``\ct{Ng1} [is] quite nice actually, [knight] on \ct{h3} is gone, and then we probably go for \ct{h4} at some point.''

\subsubsection*{Informative puzzles}
In some puzzles, we observe informative manoeuvres from AZ. We provide a few examples here.

\begin{figure}[!ht]
\caption{Queen Manoeuvre. White is to move.}
\centering
\includegraphics[width=0.45\textwidth]{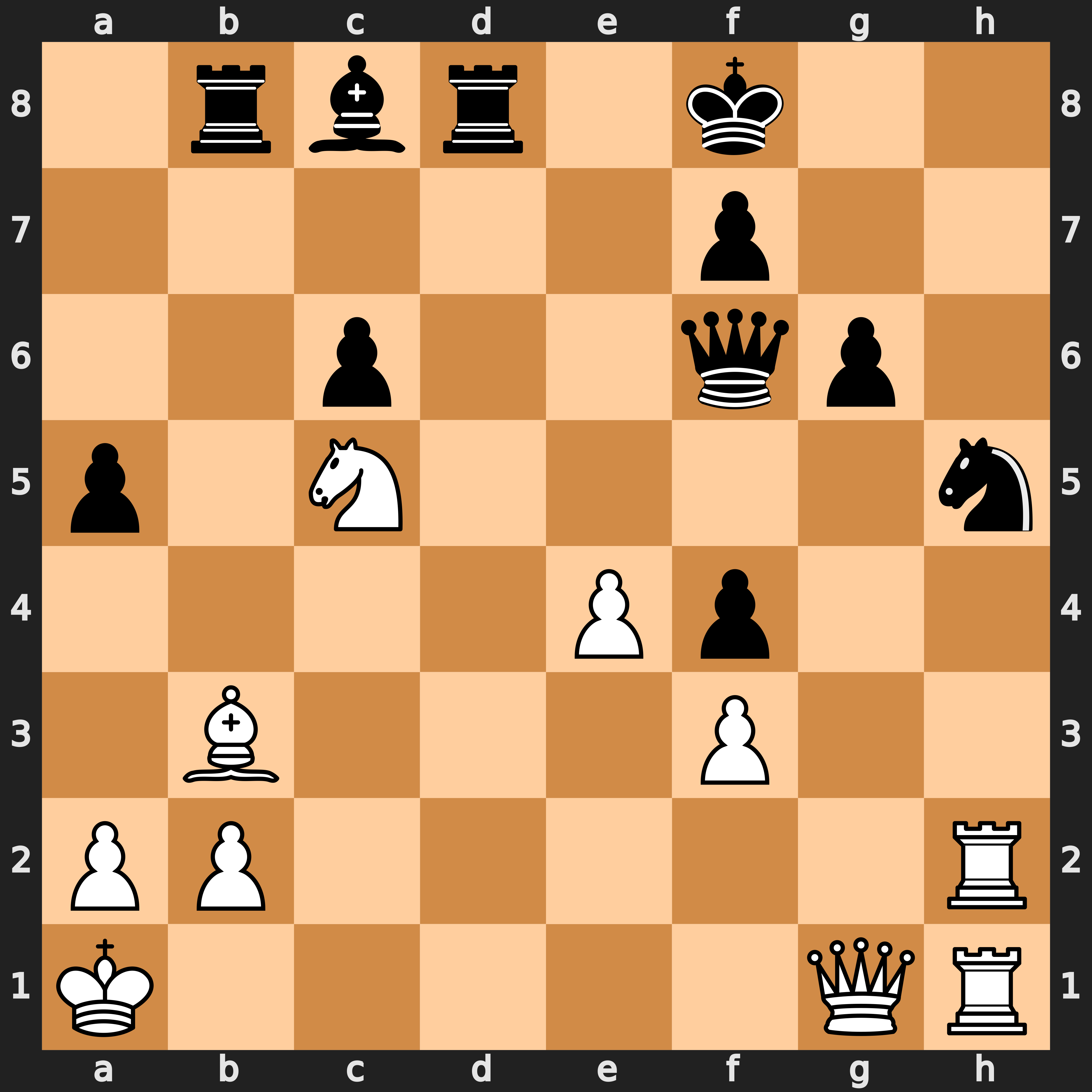} \\
\vspace{0.2cm}
\fbox{\begin{minipage}{\textwidth}
\begin{flushleft}
\textbf{AZ's calculations:}
\ct{37.Qc1 Kg7 (37...Rb5 38.a4 Rb4 39.Ka2; 37...Qe5 38.Qc4 Be6 39.Nxe6+ fxe6 40.Qxc6) 38.Re1 Qe5 39.Rc2 Rb4 40.Ba4 Qd6 41.a3 Rd4 42.e5 Qd5 43.Bxc6 Qxc6 44.Nb3 White is better}
\end{flushleft}
\end{minipage}}
\label{fig:qc1_app}
\end{figure}

In the chess position in Figure~\ref{fig:qc1_app}, AZ plays \ct{Qc1} with the idea of manoeuvring it to \ct{c4}. 
Most human chess players would find this idea
unconventional, as White's pieces seem to be active on the kingside. However, there is no way to break through Black's position. AZ's idea is the only way to maintain an advantage. The plan is to re-position the pieces to the queenside, with ideas like \ct{Re1, Ba4 and e5}. 

When analysing this position (and only spending a fixed amount of time), the grandmaster misted the idea and opted for \ct{Rxh5}, which was the only way for White to equalise according to the grandmaster. 
We speculate that the difference between AZ and humans is because AZ is more flexible in changing its plan. 
In this position, humans are likely primed to continue playing on the kingside. 

\begin{figure}[!ht]
\caption{Positional Tactics. White is to move in both positions.}
\centering
\includegraphics[width=0.45\textwidth]{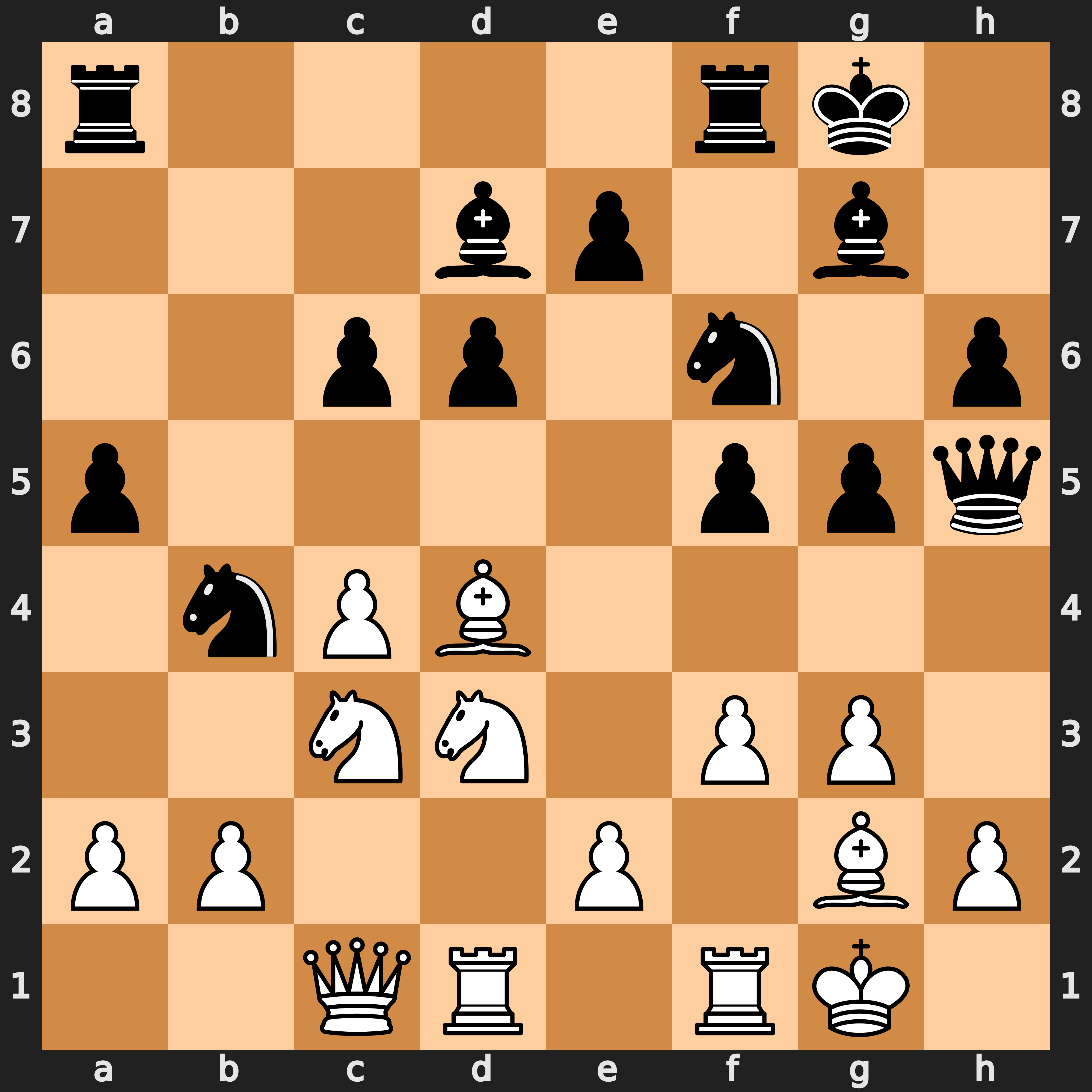}
\includegraphics[width=0.45\textwidth]{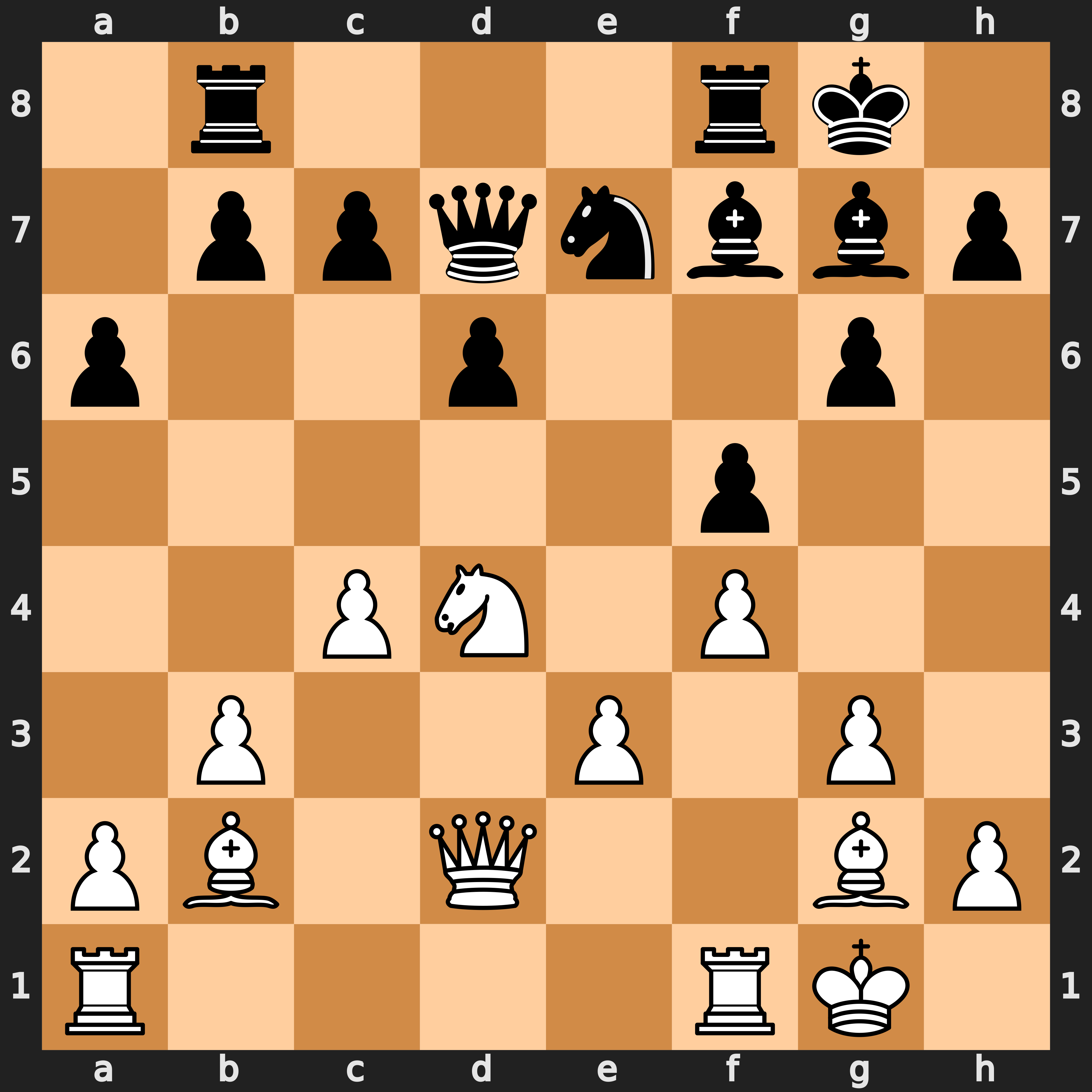} \\
\vspace{0.2cm}
\fbox{\begin{minipage}{\textwidth}
\begin{flushleft}
\textbf{AZ's calculations (left):} \ct{21.Nc5 dxc5 (21...Bc8 22.N5a4 Be6 (22...c5 23.Bxf6 Bxf6 24.a3 Ra6 25.f4)  (22...f4 23.Nb6 Rb8 24.Nxc8 Rbxc8 25.a3 Na6 26.e3) 23.a3 Na6 24.c5 dxc5 25.Nxc5 Nxc5 26.Bxc5) 22.Bxf6 Bxf6 23.Rxd7 Bd4+ 24.Kh1 f4 (24...Qe8 25.Rb7 Rf7 26.f4 Qc8 27.Rb6 Qd8 28.Na4) 25.Ne4 White is slightly better}
\end{flushleft}
\end{minipage}}
\vspace{0.2cm}
\fbox{\begin{minipage}{\textwidth}
\begin{flushleft}
\textbf{AZ's calculations (right):} \ct{18.Rad1 Rbe8 (18...b5 19.c5 Rbe8 20.Ba3 Nd5 21.Rfe1 b4 22.Bxb4 Nxb4 23.Qxb4) 19.Nf3 Bxb2 20.Qxb2 Nc6 21.Rfe1 h6 22.Qc3 Re7 23.b4 Rfe8 24.a4 White is better}
\end{flushleft}
\end{minipage}}
\label{fig:pos_tactics}
\end{figure}

The puzzles in Figure~\ref{fig:pos_tactics} correspond to the same concept. In both positions, AZ uses tactics to obtain a positional advantage by maintaining a space advantage.
In the left puzzle, the best move is \ct{21.Nc5}, which stops Black from advancing their $c5$ pawn to control the center. The tactic behind the idea is \ct{21.Nc5 dxc5 22.Bxf6 Bxf6 23.Rxd7}.

In the puzzle on the right of Figure~\ref{fig:pos_tactics}, Black plays \ct{18.Rad1} which is prophylactic against \ct{18...b5} as White has tactic \ct{19.c5 dxc5 20.Nc6}.
Both of these continuations were found by the grandmaster, who appreciated the importance of \ct{Nc5}.
The grandmaster explained that they took a long time to analyse this position as they found it complicated. While recognising Black's threat of c5 (followed by \ct{Bc6} or \ct{Nc6}), they first explored several other options, including moves such as \ct{a3}, \ct{c5}, \ct{Bf2} or \ct{Na4}. However, after exploring other moves, they found \ct{Nc5}. The grandmaster commented that the idea was very strong and 'by far the best move'. 

In this puzzle, the grandmaster opted for the tactical move \ct{Bxh7} but also considered quieter moves such as \ct{Re3}, \ct{Be4} or \ct{Be1}. When analysing this position, they commented 
\begin{displayquote}
``This is tricky, if White decides to protect the pawn, it's clearly better due to the weakness in Black’s structure, but somehow getting addicted to
more forced attacking lines. \ct{Bxh7} is hard to figure out, maybe \ct{Kxh7} [followed by] \ct{Qh5}-\ct{Qf7} then \ct{Rd3}, \ct{Bxg2} ... [I] didn’t calculate until clear much better position, but thought even with some play, h-file the long problem, maybe certain chances [to win].''
\end{displayquote}
However, this sequence ends in a draw, and AZ instead opts for \ct{f3}, maintaining the advantage for White. When reading AZ's analysis, the grandmaster commented:
\begin{displayquote}
``Wow, it’s a completely positional play. Well, my decision to [sacrifice] is too emotional. To be honest, this choice \ct{f3} ... [makes] sense as Black does not have any breakthrough idea, so if White successful [in] controlling both \ct{c5} and \ct{e5} square then its clearly much better. [My conclusion is] technically strong but again within [the] human perspective.''
\end{displayquote}
Here, we see a difference in style between humans and AZ. AZ opts for a slower, longer-dominance play in chess positions where grandmasters tend to consider more forcing sequences. 

\subsubsection*{Human vs AI Play: AZ opts for less forced lines than humans}

\begin{figure}[!ht]
\caption{AZ takes on more risk than humans due to computational capacity. White is to move.}
\centering
\includegraphics[width=0.5\textwidth]{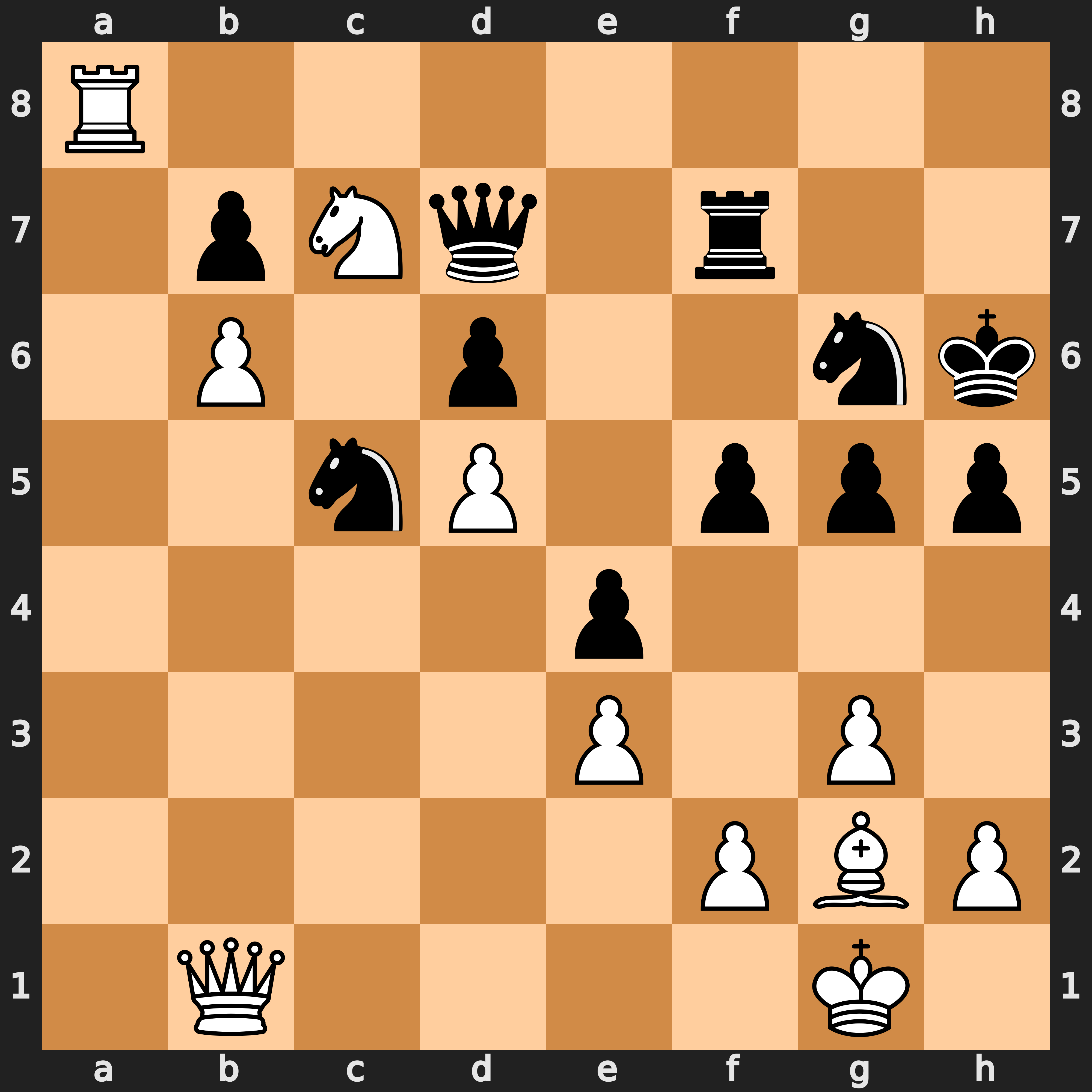} \\
\vspace{0.2cm}
\fbox{\begin{minipage}{\textwidth}
\begin{flushleft}
\textbf{AZ's calculations:} \ct{31.Qa1 f4 (31...Qe7 32.Re8 Qf6 33.Qxf6 Rxf6 34.Rd8 Ne5 35.Ne8)  (31...h4 32.Re8 f4 33.exf4 gxf4 34.Bxe4 Nxe4 35.Rxe4 Qf5 36.Qe1 Ne5 37.Qe2 hxg3 38.hxg3 fxg3 39.fxg3) 32.exf4 gxf4 33.Ne6 Nxe6 34.Bh3 Qb5 35.dxe6 Rg7 36.Ra5 White is slightly better}
\end{flushleft}
\end{minipage}}
\label{fig:risk}
\end{figure}

\subsubsection*{Difficult or non-instructive puzzles}
There may exist concepts that are intrinsically hard to understand and learn for human chess players due to differences in ways of abstract thinking, overall capabilities, and their computational budgets.
The example in Figure~\ref{fig:risk} highlights that humans and AI have different computational capacity, allowing AZ to make moves that appear risky to humans.

In the puzzle in Figure~\ref{fig:risk}, AZ plays \ct{Qa1} to activate the queen. 
This move requires calculating carefully to ensure that Black has no counterplay due to an attack on the kingside. 
As such, humans may perceive this move as risky, and it was not chosen by the grandmaster. When seeing AZ's calculations, they remarked
\begin{displayquote}
``I would be really worrie[d] to keep the queens on the board because of the threat with \ct{f4} but AZ has a tactical solution. \ct{33.Ne6 Nxe6 [34.]Bh3} is a very nice idea which is quite hard to spot. Black should probably stay still and try to hold with \ct{31...h4 [32].Re8 Re7}.''
\end{displayquote}
Here, we see that humans are more risk-averse than AZ. This is logical, given that AZ has a much larger computational capacity and can calculate more/deeper than humans can to more accurately assess the chess position (and thereby take on less risk).

\FloatBarrier
\subsubsection*{Differences in motivation in human and AI play}

The next example shows that AZ and humans play chess with different motivations.
AZ forces the draw with \ct{g4}. Upon seeing AZ's calculations, the grandmaster commented:
\begin{displayquote}
``... this is a very clear and important theme to understand. So, g4, the move it proposes, in a practical sense it’s a very big move, because you see, in such situations, the engine already knows the final result. For engine it doesn’t matter which move it plays because it calculated it’s a draw, but g4 is basically forcing it. After \ct{g4} Black has no winning chances, but otherwise I have a feeling that after Black plays let’s say \ct{Qd2}, it’s not ... easy practically for White to make the draw. For an engine it’s ok, but practically no one would play it because \ct{g4} is basically offering a draw - and with other moves Black is running zero risk, yet has practical chances to win the game if White makes a mistake. An engine doesn’t understand the concept of practical play - while this is a draw, it’s not an easy draw for White. \ct{g4} is one of many moves leading to a draw, but in a practical sense the worst one as it gives Black zero chances to win. So that is my understanding, that it’s not the objective best move. Practically definitely a wrong move.''
\end{displayquote}

This underscores a fundamental difference between AZ's playing style and human's playing style. AZ was trained to obtain the \textit{expected outcome} without an explicit term in the loss function, encouraging it to win. The incentive to find the best move comes from the exploration and move selection criteria in MCTS. Further, AZ assumes an equally strong opponent. For AZ, there is no difference between different equalising moves, even if one move requires a much more precise sequence of moves to equalise. 
In contrast, humans assume that their opponent may make suboptimal moves. As human chess players play competitively (i.e., their goal is to maximise the outcome), they will try to leverage these chances. 
This example highlights how the difference in objectives and assumptions may lead to different behaviour of AZ and humans when playing chess.

\subsubsection*{Shortcomings of method for generating prototypes}
In this puzzle, AZ chooses \ct{Rd1} whereas the grandmaster wanted to play \ct{Re1} or \ct{Ke1}. When seeing AZ's calculations, the grandmaster commented:
\begin{displayquote}
``Drawing position? At first trying to find winning moves for White, but really didn’t see any plan to make improvements. In the meantime, considering the possibility for Black to push h pawn to h3, maybe tiny
chances, its better to plan \ct{Ke1}-\ct{Qf1}-\ct{Qf3} at the beginning, or moving the rook to e1 with the idea \ct{Re7}, forcing ... \ct{Ra1} check then White rook retreat to \ct{e1}, ... [draw by] repetition.''
\end{displayquote}

\begin{figure}[!ht]
\caption{White is to move }
\centering
\includegraphics[width=0.5\textwidth]{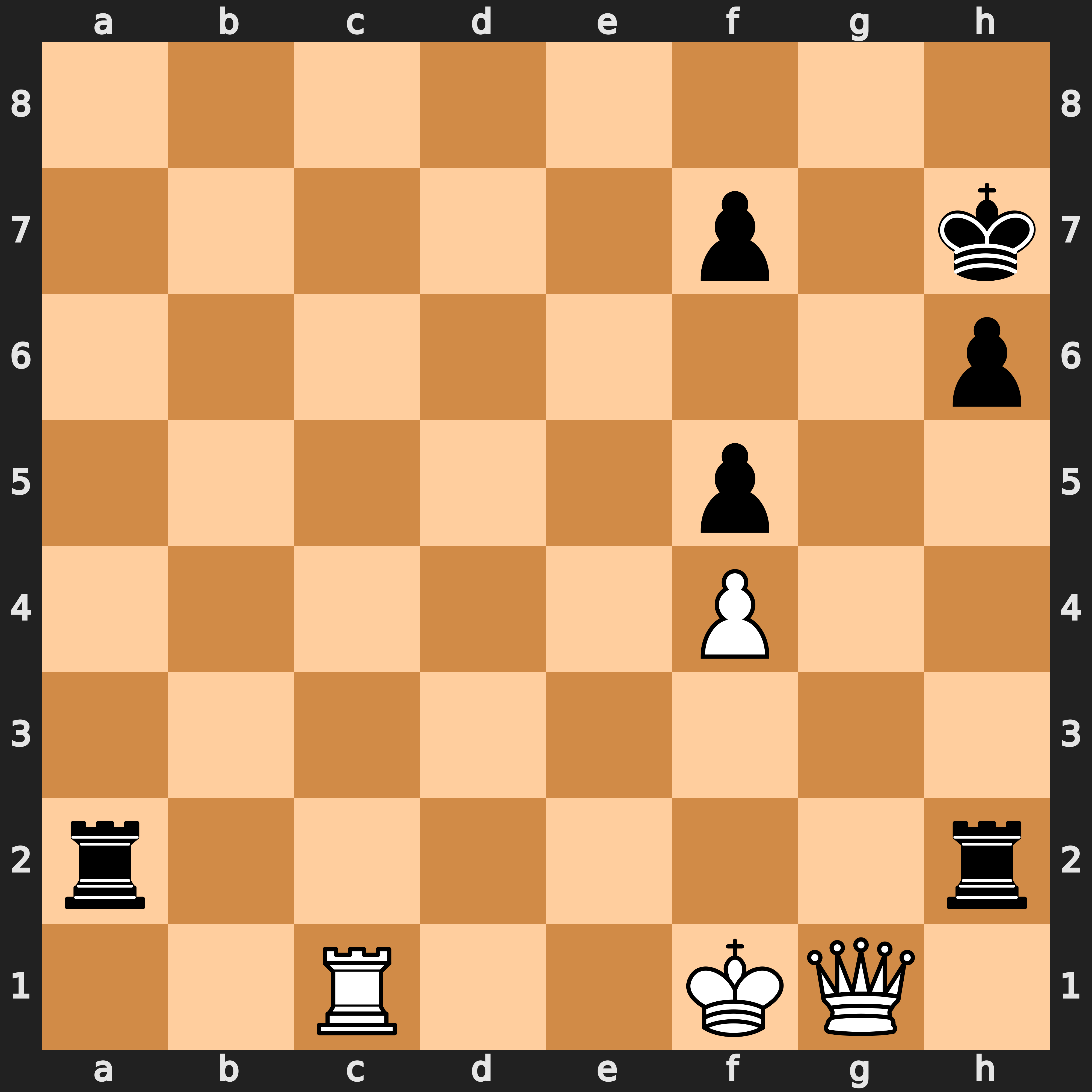} \\
\vspace{0.2cm}
\fbox{\begin{minipage}{\textwidth}
\textbf{AZ's calculations:}
\begin{flushleft}
\ct{65.Rd1 Rh4 (65...Rac2 66.Rb1 Rh4 67.Ra1 Rxf4+ 68.Ke1 Rg4 69.Qa7 Kg7) 66.Ke1 (66.Rb1 Rxf4+ 67.Ke1 Re4+ 68.Kd1 Rf4 69.Qh1 Rff2) 66...Rhh2 67.Rc1 Rag2 68.Qf1 h5 69.Rc7 Kg7 Draw}
\end{flushleft}
\end{minipage}}
\label{fig:rook_queen}
\end{figure}

This puzzle can be seen as a shortcoming of our method for finding prototypes. We only filter positions based on the criteria described in \S\ref{appx:human}; however, this position does not fall under one of our categories. 
This puzzle is not informative for humans as there are many other viable options. As such, it is more difficult to understand the concept from the sequence of moves.  

\vfill

\subsection{Background: AZ policy value network} \label{appendix:AZ PV network}

For a complete description of AZ, see \cite{schrittwieser2019mastering} or  \cite{mcgrath2022acquisition}. Here, we provide a brief description. 
AZ has two main components: a policy value network and MCTS. Below, we describe the policy-value network. 

For a given input (consisting of a chess position and metadata), the network outputs a policy and value estimate. As described in \cite{mcgrath2022acquisition}, the main `body' of the network has a ResNet backbone. The body consists of
$19$ residual blocks. Each block contains two convolutional layers, followed by a skip connection. Let $z_l$ denote the post-activation latent representation corresponding to block $l$:
\begin{equation}
z_l = \text{ReLU}(z_{l-1} + g_l(z_{l-1})),
\end{equation}
where $g_l(\cdot)$ is the composition of two convolution layers. 

In the main body of the text, we refer to layers using integers. For the main body of the network, these integers denote the ResNet block. In the value head, we have three layers, which we will refer to as layers $20$, $21$ and $22$ (from the body to value output). 
In the policy head, we have two layers, which we refer to as layers $23$ and $24$ (from body to policy output).

\begin{figure}[h]
\caption{Policy Value Network in AZ, adapted from \cite{mcgrath2021acquisition}. }
\centering
\vspace{0.2cm}
\scalebox{0.64}{
\input{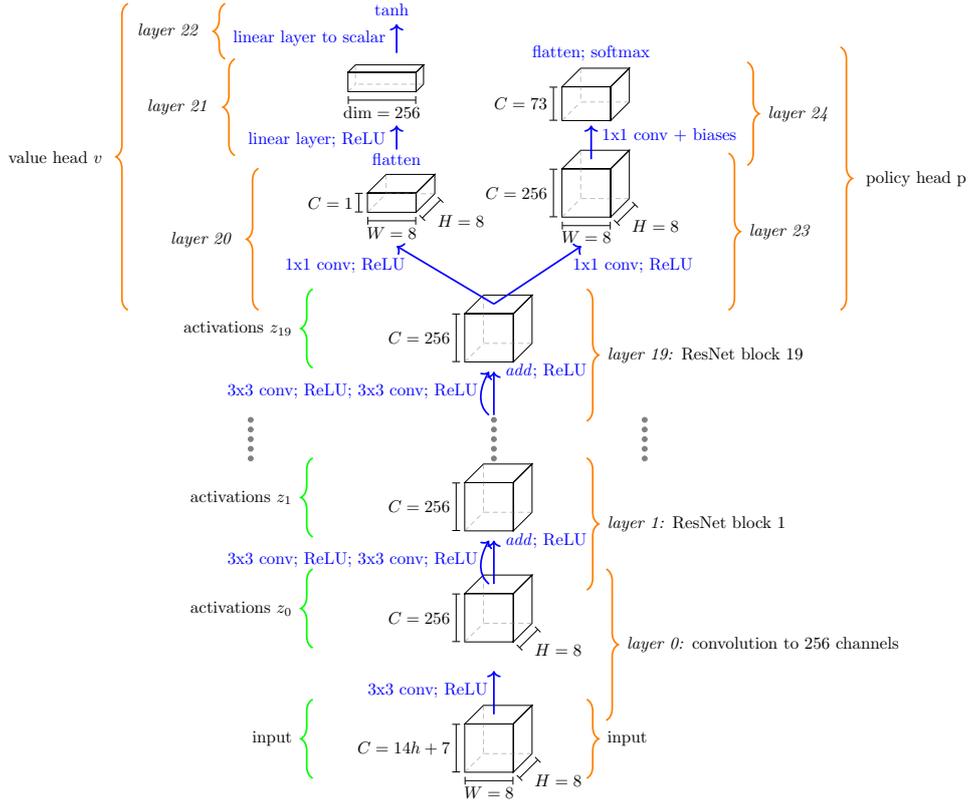}
}
\label{fig:az_network}
\end{figure}
\FloatBarrier

\subsection{Background: how humans and AZ play chess} \label{appendix: how AZ plays chess}
In this section, we provide further intuition on how humans play chess, and how this relates to AZ's system. This section provides context as to why concepts should explain the policy value network and MCTS to provide a holistic view of chess. 

\begin{figure}[ht] 
\centering
\caption{Simplified Summary of How Humans Play Chess}
\includegraphics[width=0.85\textwidth]{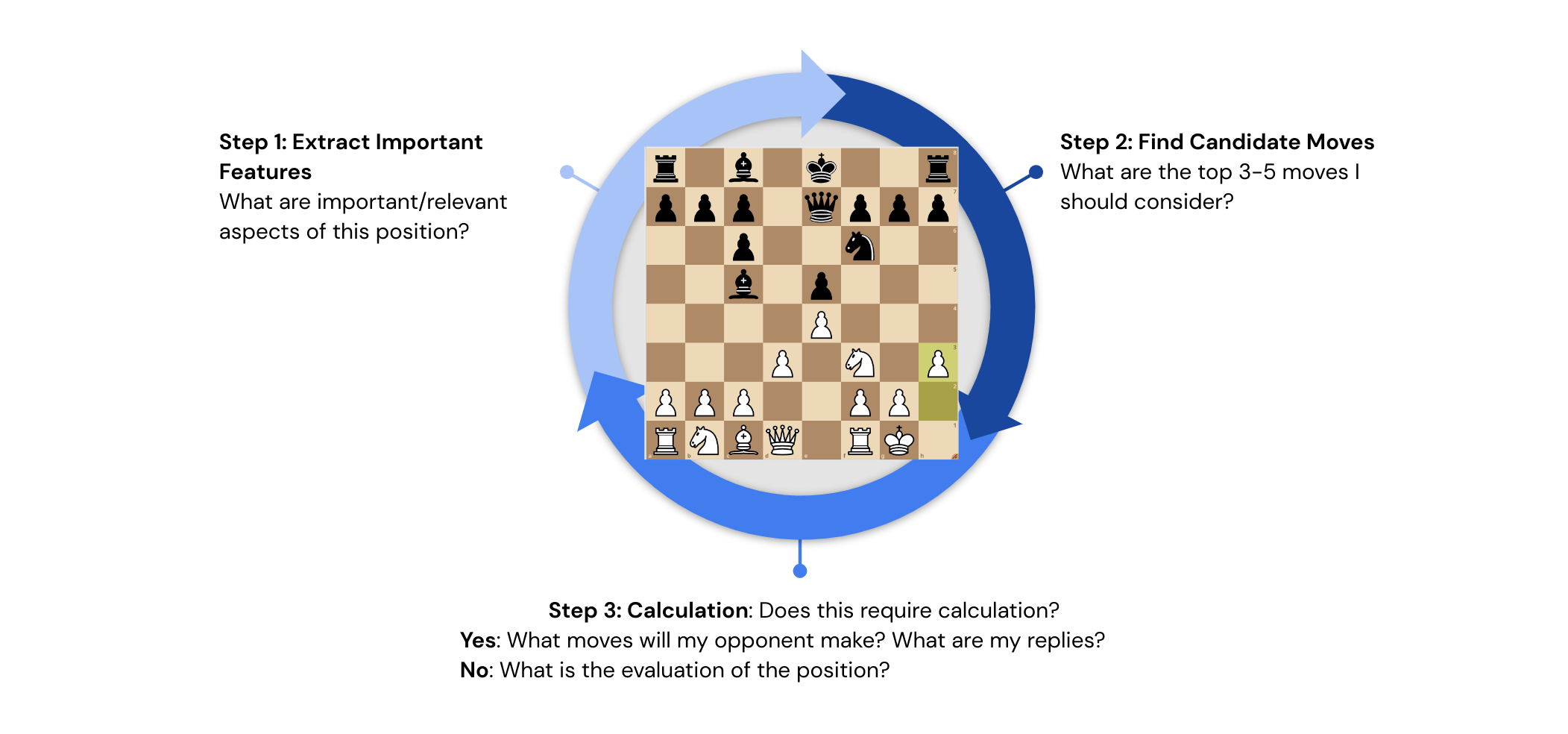}
\label{fig:human_calc}
\end{figure}

Figure~\ref{fig:human_calc} shows a simplified summary of how human chess players analyse a chess position. They generally ask the following questions:
\begin{enumerate}
    \item What are the critical aspects of this chess position? E.g., on which side of the chess board do I have more space? How do I want to develop my pieces? What are my opponent's weaknesses? 
    \item Based on step 1, a chess player will find a \textit{couple candidate moves} -- actions they could play in the chess position. 
    \item For each move, they may calculate a likely continuation -- i.e., what is the likely sequence of moves to follow? 
\end{enumerate}
This process loops until the player has considered all candidate moves, calculated the relevant move sequences, and determined the optimal trajectory.\footnote{This is a simplified model - in practice, other factors such as time are important.} To play well, chess players must understand the important features of a chess position and calculate move sequences to understand the correct evaluation of the chess position. 
However, there are several different types of chess positions (e.g., endgames or attacking chess positions) where principles alone are insufficient to determine the optimal continuation, and calculation is necessary.  

\begin{figure}[ht] 
\centering
\caption{Simplified Summary of How AZ Plays Chess}
\includegraphics[width=0.85\textwidth]{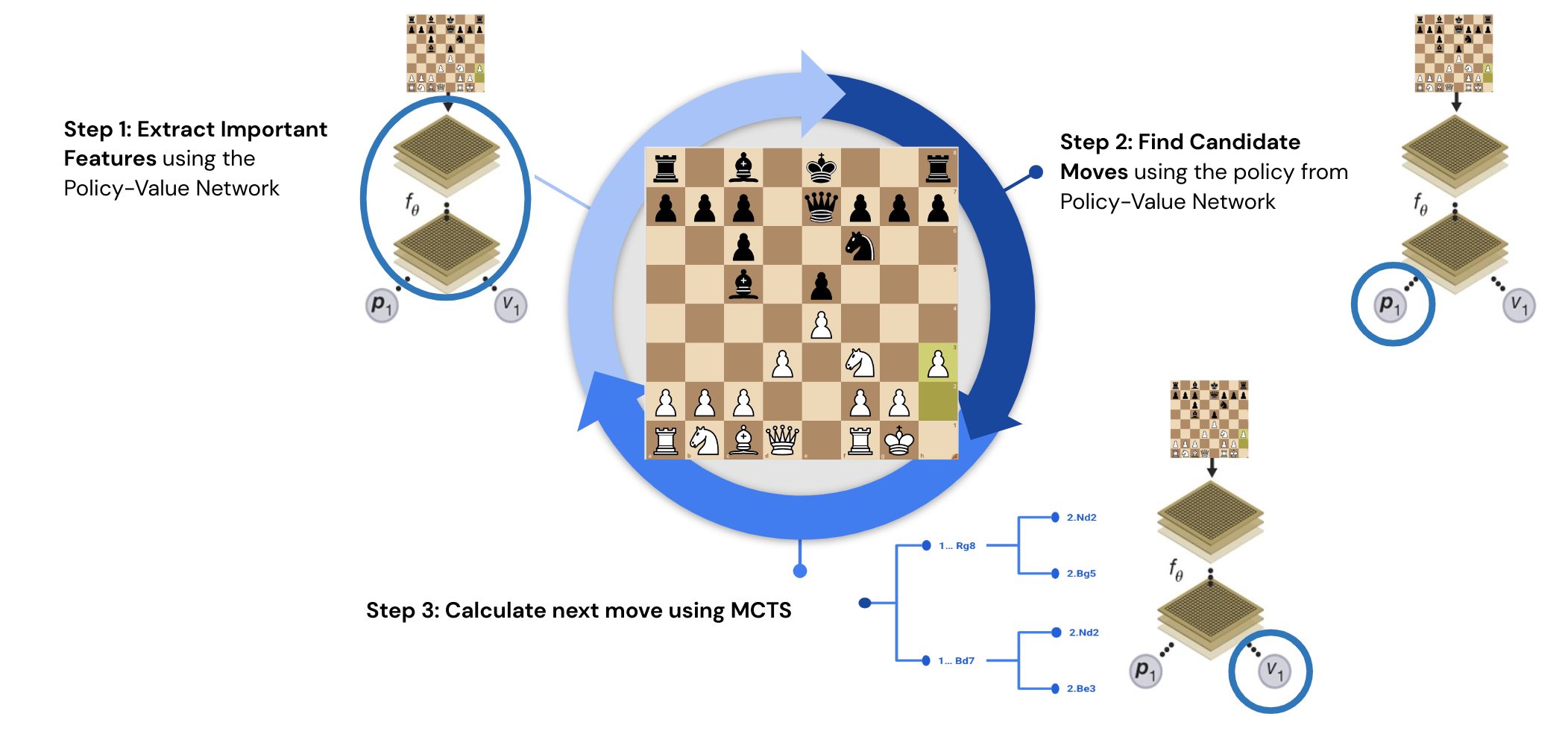}
\label{fig:az_calc}
\end{figure}

AZ uses a similar approach. In a given chess position, AZ extracts features using the layers in the policy-value network and outputs a policy and value estimate. 
The policy weighs the different possible actions, and the moves with the most probability mass can be interpreted as the \textit{candidate moves}. 
Next, the policy is passed on as an input to MCTS, a search algorithm that calculates the optimal move. 
By drawing parallels between AZ's system and how humans play chess, we highlight the importance of each component of AZ. This motivates our method's design, which incorporates all components of AZ.

\subsection{Further details: convex optimisation formulation for concepts}

This section provides further details on our convex optimisation framework to find concepts.
Subsection~\ref{sec:hypers} describes how we set the dynamic concept hyperparameters. 
\S\ref{appendix:cp_formulations} explains how the convex optimisation formulations were implemented for the different datasets. 

\subsubsection{Dynamic concept hyperparameters} \label{sec:hypers}
As AZ learnt to play chess through self-play, the latent representations alternates between the player's and the opponent's perspective within a rollout. 
For concepts, we may want to find a concept that influences a single player or both players. 
Therefore, we consider two different ways of using rollouts. For a rollout  $\{z_{t} \}_{t=0}^T$, we use every other latent representation, i.e., $\{z_{2t} \}_{t=0}^{\text{floor}(T/2)}$, to find concepts for a single player (i.e., for the player to move, or their opponent). These concepts are referred to as `single'. We use every latent representation to find concepts for both players, i.e. $\{z_{t} \}_{t=0}^T$. These concepts will be referred to as `both'.
For the rollout depth used in our dynamic concept formulations, we consider $T=5$ and $T=10$. 
For the subpar variations, we required AZ to estimate a minimum value difference of $0.20$ and/or a visit count difference of $10\%$ (of the most visited move). 

\subsubsection{Datasets}\label{appendix:dataset}
We use labelled and unlabelled datasets to construct concept constraints for the convex optimisation formulation. We use a different convex optimisation formulation for each concept we want to discover. Therefore, for each concept, we need a set of chess positions $\Xset^+$ that contains a concept.
We leverage educational resources designed to teach humans chess, which contains themed chess puzzles: positions designed to encapsulate single important concepts and test the degree to which a chess player can deploy them in realistic situations. We go beyond chess puzzles, delving into chess positions arising from different openings and searching for AZ-specific concepts to find new ones.

\paragraph{Factors in datasets.} The datasets we use vary in the following ways:
\begin{itemize}
    \item \textbf{Concept type} e.g., the concepts can be strategic or tactical (see \cite{wiki:chess_strategy} and \cite{wiki:chess_tactic} for a further explanation on strategy and tactics), or correspond to different periods of the game (opening/middle game/endgame). 
    \item \textbf{Degree of human knowledge} some datasets contain human games while others contain AZ's games. The degree of human knowledge (or style of play) may vary across the datasets.
    \item \textbf{Complexity} some concepts are elementary (i.e., can be learned by beginner-level chess players), whereas others are highly complex (i.e., can only be understood by top-level grandmasters).
\end{itemize}

Below, we briefly describe the different data sets used
\begin{enumerate}
\item \textbf{Piece dataset} We construct a labelled dataset that contains paired chess positions that either contain or do not contain a chess piece. These simple concepts indicate the presence of a piece - queen, rook, bishop, knight, or pawn. We exclude the concept of `king' as this piece is always present in a chess game.
    \item \textbf{Stockfish dataset} We construct a labelled dataset for each concept in the Stockfish engine~\citep{stockfish}. Each dataset contains two sets of chess positions -  with and without the concept. Humans use Stockfish concepts  in chess position evaluation (such as piece placement, open files, etc.) 
    \item \textbf{STS puzzles dataset} This is a labelled dataset containing 15 different categories (see \cite{sts} for further details). These puzzles capture different types of strategic themes.
    \item \textbf{Chess openings} (e.g., concepts in Grünfeld vs. Najdorf vs. Queens Gambit) Openings (i.e., the first couple of moves) determine the pawn structure that acts as the backbone of a chess position -- it determines crucial aspects of the game, such as optimal piece placement, square weaknesses and strengths, and more generally, plans. We use the Encyclopedia of Chess Openings (ECO) to create a set of labelled chess opening positions~\citep{eco}. For each chess position, we use AZ's MCTS rollouts to construct sequences of the opening moves. 
    \item \textbf{AZ self-play games} We construct an unlabelled dataset. We sample $30,000$ chess positions from AZ's games. To ensure that the chess positions we analyze contain complex concepts,
    we only select chess positions where two versions of AZ at different points in training select a different move. These versions differ by 75 Elo points. 
\end{enumerate}
Table~\ref{table:summary_dataset} briefly summarises the different datasets. 

\begin{longtable}{l c c c c} 
\caption{Datasets Summaries: from concepts more known to humans (top rows) to AZ (bottom rows)}
\label{table:summary_dataset} \\ 
\toprule
\textbf{Name} &  \multicolumn{2}{c}{\textbf{Concept Type}} &  \textbf{Type of}  & \textbf{Complexity} \\  
& \textbf{Strategic vs. Tactical} & \textbf{Game Phase} &\textbf{Knowledge} &  \\ 
\midrule
\endhead
\midrule
\multicolumn{4}{r}{{Continued on next page}} \\
\midrule
\endfoot
\bottomrule
\endlastfoot
Piece  & N/A & All & Human & Low \\ 
Stockfish  & Both & All & Human & Varies \\ 
STS &  S & Middle/End & Human & Medium \\ 
Opening  & S & Begin & Human/AZ & Varies \\
AZ  & Both & All & AZ & High \\ 
\end{longtable}

The human crafted datasets are used to (1) validate the convex optimisation framework and (2) explain the novel concepts by relating discovered concepts to something humans know (by learning a graph). 

\subsubsection{Convex formulation for different datasets} \label{appendix:cp_formulations}

In this section, we provide further details on the convex optimisation formulations for each dataset. 
Unless otherwise stated, we use a 90-10 train-test split for the supervised datasets. 
In our analysis, we generally consider layers $19$, $20$, $21$ and $23$. However, as layers $19$ and $23$ showed the largest difference in rank compared to human data in our analysis \S\ref{sec:rank}, we use layer these layers to discover the concepts shown to grandmasters.

\paragraph{Piece.}
We artificially created a piece dataset by sampling chess positions from grandmaster games sampled from \cite{chessbase}. For each concept, we sampled $100$ chess positions with the concept to create the set $\Xset^{+}$ and then created $100$ chess positions without the concept to create the set $\Xset^{-}$(by removing the piece). 

We formulated the convex optimisation problem as follows. For pair of chess positions, $x_{i}^+ \in \Xset^{+}$ and $x_{i}^- \in \Xset^{-}$, we find the corresponding latent representations in layer $l$ to create $\Zset_l^{+}$ and $\Zset_l^{-}$, respectively. Using these latent representations, we can search for the piece presence concept using the following formulation
\begin{align}
 \min & \quad ||\vcl||_1 \\
 \st  & \quad \vcld z_{i,l}^+ \geq \vcld z_{i,l}^- \quad  \forall z_{i,l}^+ \in \Zset_l^+, z_{i,l}^- \in \Zset_l^-.
\end{align}
Note that this is the same formulation as for static concepts more generally (see \S\ref{sec:cp_static_concept}). 

\paragraph{Stockfish.}
Following \cite{mcgrath2022acquisition}, we extract the concepts encoded in the Stockfish engine. We use the Stockfish engine code to extract a concept value for each position.  We sampled $30,000$ chess positions from \cite{chessbase}. For each of these chess positions, we found the chess positions that were in the top 5th percentile (assumed to contain the concept) to construct $\Xset^+$ and the bottom 95th percentile (assumed not to contain the concept) for the concept score to construct $\Xset^-$. 
We randomly paired the chess positions that contained the concept, $x_{i}^+$, with chess positions that scored low for the concept $x_j^{-}$. 
Similarly to before, we extracted the latent representations for each concept. 
Then, using the formulation for static concepts, we found the concept vector using the following formulation
\begin{align}
 \min & \quad ||\vcl||_1 \\
 \st & \quad \vcld z_{i, l}^+ \geq \vcld z_{j, l}^- \ \forall i:x_i \in \Xset^+, j:x_j \in \Xset^-.
\end{align}

\paragraph{Strategic.}
We use the strategic test suite to extract strategic concepts~\citep{sts}.  
In this dataset, there are 15 different concepts. We omit the 12th concept due to irregular data formatting. The remaining concepts are undermine, open files, knight outposts, square vacancy, bishop vs. knight, recapture, offer of simplification, fgh-pawn, abc-pawn, simplification, king activity, pawn push center, 7th rank, avoiding an exchange~\citep[see][for further details]{sts}.

Each concept has a set of $100$ chess positions, $\Xset$, and the solution (move) requires applying a strategic concept in each chess position. 
In our analysis, we run  MCTS on the chess position and store the search statistics. 
We store the optimal trajectory for each chess position $x_i$, denoted as $\Xset^{+}_{i, \leq T}$, where $T$ is the maximum rollout depth.  
Similarly we select a subpar rollout $\Xset^{-}_{i, \leq T}$. To find the subpar rollout, we find a rollout in the MCTS tree with the most visits where (1) the estimated difference in value is at least $0.2$, and (2) the visit difference is at least $10 \%$. 
As in the main text, for $\Xset^{+}_{i, \leq T}$ and $\Xset^{-}_{i, \leq T}$, we find the corresponding latent representations in layer $l$: $\Zset^{+}_{i, \leq T}$ and $\Zset^{-}_{i, \leq T}$, respectively.
Then, we can find the concept vector using the dynamic concept formulations (see \S\ref{sec:cp_dynamic_concept}) as follows
\begin{align}
 \min \  & ||\vcl||_1 \\
 \st \ & \vcld z^{+}_{i,t} \geq \vcld z^{-}_{i,t} \forall t \leq T, i: x_i \in \Xset.
\end{align}
In our analysis, we use a maximum depth of $T=5$.

\paragraph{Openings.}
For the openings, we focus on the English, Dutch, Scandinavian, Najdorf, French, Tarrasch, Winawer, Ruy Lopez, Grünfeld, King's Indian, Queen's Gambit Declined and Queen's Gambit Accepted.
We consider a subset of all openings due to computational costs. 
For each opening, we use the encyclopedia of chess openings to find relevant starting chess positions to construct $\Xset^{+}$
(see \href{https://github.com/lichess-org/chess-openings }{the encyclopedia of chess openings}).  
For each chess position, we ran MCTS to obtain the search statistics and used the formula in Equation~\ref{eq:tree_formula} to find a concept for each opening.
Further, we create a concept set $\Xset^{+}$ for each ECO index belonging to one of the aforementioned openings.

\paragraph{AZ Games.}
We simulated $1,308$ games. Conditional on hardware, AZ's play is deterministic (after training).  To create diverse games, we sample different starting chess positions. We use the ECO to find starting chess positions
(see \href{https://github.com/lichess-org/chess-openings }{the encyclopedia of chess openings}) and we simulate games from these initial chess positions.  
For each chess position, we ran MCTS to obtain the search statistics.  

We leverage AZ's training history to find interesting chess positions. We select a version of AZ that is 75 Elo points weaker than the final model.
To construct $\Xset^{+}$, we run through each game and select chess positions where the two AZ versions choose a different move. 
Using this approach, we constructed a dataset with $3,974$ chess positions and used the formulation provided in Equation~\ref{eq:tree_formula} to find concept vectors.

\begin{longtable}{lrrrrrrrrr}
\caption{Hyperparameter selection for $\beta$}
\label{table:beta_hypertuning} \\ 
\toprule
  &  $0.05$ & $0.01$ & $0.025$ & $0.05$ & $0.1$ & $0.25$  & $0.5$ & $1.0$ & $2.0$ \\ 
\midrule
\endhead
\midrule
\multicolumn{5}{r}{{Continued on next page}} \\
\midrule
\endfoot
\bottomrule
\endlastfoot
Layer 18 & 58.42 & 58.58 & 58.50 & 58.42 & 58.17 & 58.17 & 57.42 & 56.75 & 55.00 \\ 
Layer 19 & 60.17 & 60.42 & 60.42 & 60.17 & 60.08 & 60.17 & 59.75 & 58.00 & 55.83 \\
Layer 23 & 58.83 & 59.42 & 59.33 & 58.83 & 58.58 & 56.25 & 53.92 & 47.75 & 40.33\\ 
\end{longtable}

\paragraph{Other Implementation Details.}
We solve the convex optimisation problem using a standard solver in the package \texttt{cvxpy} \citep{cvxpy1, cvxpy2}.

\subsection{Beta hyperparameter tuning} \label{appx:beta}
In this section, we provide the validation values for the concept amplification experiments in \S\ref{res:concept_amplification}.
For the values in Table \ref{table:beta_hypertuning}, we randomly chose 2 concept sets from the STS dataset and estimated the amplification results for different values of $\beta$.
Overall, we observe that the results are not very sensitive to the value of $\beta$. 

\subsection{Teachability implementation} \label{appx:teachability}

In this section, we provide further details on the implementation of teachability. We assume that we have a concept vector $\vcl$ that was found in layer $l$. To construct prototypes, we sample $30,000$ chess positions from AZ games. For each concept, we find chess positions for which $\vcld z_{i,l}$ is in the top $2.5 \%$, and store these as prototypes $ \{ x_{1,l}, \dots x_{n,l} \}$. 
For dynamic concepts, we found prototypes using MCTS statistics. For each chess position, we ran MCTS. Next, we found the optimal and subpar line (similar to the convex optimisation formulation constraint). For a prototype $x_i$, we required that  $v_{c,l}\cdot z^{+}_{i,t,l} \geq v_{c,l}z^{-}_{i,t,l}, \ \forall t$.

We use AZ as the teacher network. For the student network, we want to find an agent that does not know the concept but does understand chess. As chess is a complex game, we cannot train an agent from scratch (using only the curriculum). Instead, we take a training checkpoint of AZ and estimate its knowledge of the concept using
\begin{equation} 
   T =  \sum_{x_i \in \Xset} \mathbbm{1} [ \text{argmax}(\pi_s(x_i)), \text{argmax}(\pi_t(x_i)) ],
\end{equation}
where $\pi_s()$ is the student policy and $\pi_t()$ is the teacher policy. This measures how often the teacher and student agree on the best move. We select the student as the latest checkpoint for which the top-1 policy overlap is less than $0.2$.  

We use the prototypes as a curriculum. We train the student network by minimizing the KL divergence between the policies, $\sum_{x_i \in \Xset_{train}} \text{KL}[\pi_t(x_i), \pi_s(x_i)]$. 
We use the Adam optimiser~\citep{kingma2014adam} with learning rate \num{1e-4}. As we have several concepts, we train each student for $5$ epochs, as we find that this is sufficiently indicative of performance if we train for longer. 

We benchmark the performance by comparing it to a student network trained on a random concept and evaluated on (1) the concept data and (2) the random data. 
A random concept is a vector with the same shape as the latent dimension and sampled from a standard normal distribution.  The vector is then re-scaled by a factor of (1/n) where $n$ is the number of hidden units in the layer. 
We use the random concept vector to find random prototypes. The randomly sampled prototypes are still informative as they are paired with AZ's policy. 

\subsection{Graph analysis} \label{appx:graph_analysis}
We build a directed graph between AZ's concept vectors and human-labelled concept vectors to improve our understanding of the new, unlabelled concepts.
In our work, we found a set of concept vectors $\mathbb{V}$ (see \S\ref{sec:discovering_concepts}). Assume we have two different concept vectors $v_{c,l}, v_{k,l} \in \mathbb{V}$, where $c \neq k$. 
Our idea is simple: if concepts $\vcl$ and $v_{k,l}$ are related, they will arise in the same chess positions.  
Let $s_{c,i,l}$ denote whether concept $\vcl$ is present in a chess position $i$ using the latent representation $z_{i,l}$ in layer $l$. 
For each concept $c$, we estimate the regression model (similar to \cite{meinshausen2006high})
\begin{equation} \label{graph regression}
  s_{c,i, l} =\sum_{l\in 19,23} \sum_{k \in \mathbb{V}\backslash c} \beta_{k,l} s_{k,i, l} + \lambda ||\beta_{k,l}||^2 \quad \forall i: x_i \in \Xset,
\end{equation}
where $\beta_{k,l}$ is the regression coefficient and $\Xset$ is a set of positions 
(we will elaborate on this later). If $\beta_{k,l}$ is significant at a $5\%$ level, we add a directed edge from $v_{k,l}$ to $\vcl$.
We use a regression as a concept $\vcl$ may be important for $v_{k,l}$ but not vice-versa.

The next question is: how do we define $s_{c, i, l}$? Following the ideas in \S\ref{sec:discovering_concepts}, we assume that product $\vcld z_{i,l}$ is higher for $z_{i,l}$, corresponding to positions that contain concept $c$.  
Therefore, for static concepts, we measure the concept presence as 
\begin{equation}
    s_{c,i,l}^{\text{static}}= \vcl \cdot z_{i,l},
\end{equation}
where $z_{i,l} \in \Zset_l^s$, and $\Zset_l^s$ is a set of latent representations in layer $l$ corresponding to a set of chess positions $\Xset^s$.
We sample $2,000$ positions from human games to create $\Xset^s$ as static concepts are found using human-labelled positions.

For dynamics concepts, we estimate the concept relevance as
\begin{equation}
    s_{c, i,l}^{\text{dynamic}}= \sum_{z_{i,l,t} \in \Zset_{i,l, \leq T}} \vcld z_{i,l,t}
\end{equation}
where $z_{i,l, t} \in \Zset_{i,l, \leq T}$ are the latent representations corresponding to the AZ's chosen MCTS rollout for a starting chess position $x_i$ and maximum depth $T$, and $x_i \in \Xset^d$.
We sample $2,000$ positions from AZ's games to construct $\Xset^d$, as the dynamic concepts are predominantly found in AZ's games. 

As we want to analyze the relationship between static and dynamic concepts, we estimate $s_{c,i,l}^{\text{static}}$ and $s_{c,i,l}^{\text{dynamic}}$ for each $\vcl$. 

In our analysis, we only keep human concepts of high quality, i.e., that have a concept constraint score of higher than $0.90$. Further, for our analysis to be stable, we removed highly correlated variables -- i.e., variables that had a correlation coefficient higher than $0.99$~\citep{robinson1974fitting, snee1973some, snee1984comment} dropping $580$ concepts (out of $1,371$).

\paragraph{Graph Summary.}
The graph is dense -- the graph contains $29\% \ (180,609/625,681)$ of the possible directed edges.
As the graph is very dense, we do not provide the entire graph. Instead, we provide parts of the graph in \S\ref{sec:human} when presenting examples of AZ's concepts. 

\paragraph{Graph Verification.}
To verify the graph, we run an experiment to test whether two concepts with an edge contain related knowledge.
If a model learns a concept $c$, this should improve the model's performance on another related concept $c_e$ more than on an unrelated concept $c_n$. 
For a concept $c$, let $\mathbb{C}_e$ denote the set of concepts with an edge and $\mathbb{C}_n$ denote the set of concepts with no edge in the graph.
Following the teachability procedure in \S\ref{sec:teachability}, we train a student model using prototypes of concept $c$.
Next, using Equation~\ref{eq:teachability_train}, we evaluated student's performance on:
\begin{itemize}
    \item concepts with an edge $\mathbb{C}_e$; we denote the performance by $T_{c, c_e}$ 
    \item concepts without an edge (to $\vcl$) $\mathbb{C}_n$; we denote the performance by  $T_{c, c_n}$.
\end{itemize}
If the graph correctly captures the relationships between concepts, then we expect that a model trained on $c$ performs better on $\mathbb{C}_e$ than $\mathbb{C}_n$, i.e. $T_{c, c_e} > T_{c, c_n}$. 

However, we must consider that the concepts in $\mathbb{C}_e$ may be inherently easier to learn than those in $\mathbb{C}_n$. 
To account for this, we train a model on a random set of data (as in \S\ref{sec:teachability}), which we use as a benchmark. 
We estimate the performance of this model on $\mathbb{C}_e$ and $\mathbb{C}_n$, which we denote by $T_{r, c_e}$ and $T_{r, c_n}$, respectively. 

\begin{figure}[!ht]
\caption{Teachability score on related and unrelated concepts}
\centering
\includegraphics[width=0.45\textwidth]{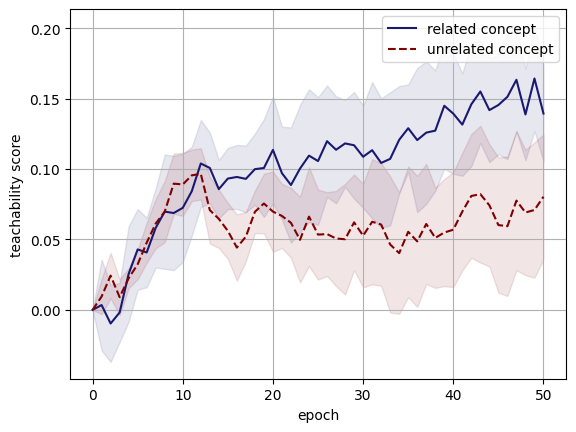}
\hspace{0.02\textwidth}
\includegraphics[width=0.45\textwidth]{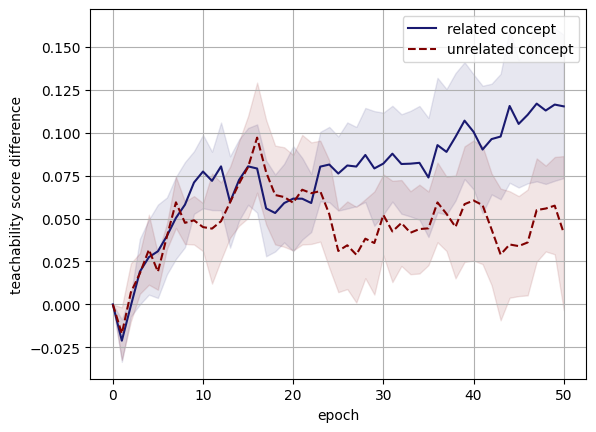}
\label{fig:retrain_concept}
\end{figure}

If our graph accurately captures the underlying relationships, we expect that $T_{c, c_e} - T_{r, c_e} > T_{c, c_n} -T_{r, c_n}$. 
Figure~\ref{fig:retrain_concept} shows training curves for
two concepts each of which with $5$ related concepts and $5$ unrelated concepts. 

\FloatBarrier

We find that the performance on related concepts is significantly better than unrelated concepts at a $5 \%$ significance level. This suggests that the graph structure may accurately capture the relationship between concepts. 

\subsection{Human experiments} \label{appx:human}

\paragraph{Recruitment.}
We recruited four chess players based on their Elo rating. All the participants hold the grandmaster title; one of our participants is rated 2600-2700, and three are rated 2700-2800. 

\paragraph{Experiment Instructions.}
Each grandmaster was asked to spend two hours on Phase 1, one hour on Phase 2 and two hours on Phase 3. We ask the grandmasters to provide (1) the move they would play or their ranked candidate moves and (2) a thought record -- the idea is to capture any thoughts about the chess position.
The grandmasters were sent the chess positions to solve at home, in their own time. 
We explained that the chess positions could vary in nature. The chess positions could be better, equal or worse for the player to move. Similarly, the continuation may require calculation or finding a general plan. 

\paragraph{Evaluation.}
We evaluated how often the grandmasters find the move selected by AZ. 
Note that if grandmasters made the right move but incorrect reasoning appeared in their free-form comments, we counted this as an incorrect answer.

\paragraph{Prototype Filtering.}
To ensure the quality of the prototype selection, we filter them  according to the following criteria:
\begin{itemize}
    \item 
    \textbf{Quality of the value estimate.} We ensure that the AZ value estimate is close to the correct assessment of the prototype by running self-play and computing the expected score. If the expected score and the value estimate are in concordance, the prototype chess position is kept, otherwise, it is discarded from consideration for the human study.
    \item 
    \textbf{Chess position complexity.} For the concepts to be sufficiently complex to be of interest to the top grandmasters, we use prototypes where the policies of the 512K step checkpoint and fully trained models disagree on their top move. The 512K checkpoint model is 75 Elo points weaker than the final model, and therefore, if the policies differ, AZ learned the continuation during a late stage of training when it was already strong.
    \item 
    \textbf{Solution complexity.} We manually remove \textit{trivial} chess positions where the solution is theoretically known (e.g., present as an entry in pre-computed tablebases such as the Syzygy tablebase~\citep{tablebase}). Tablebases are sets of chess positions where the ground truth evaluation (outcome with ideal play) is known.
    \item \textbf{Reliability.} We reject chess positions where AZ's limited compute budget may lead to an unreliable chess position evaluation (i.e., where we observe abrupt changes in the predicted outcome). Therefore, we require that the evaluation stays approximately consistent (i.e., the predicted outcome (win/loss/draw) does not change) throughout the provided lines. 
\end{itemize}
We did not filter based on the difference in the value or the policy probability mass of the optimal move compared to other moves. 
The reasons are that (1) AZ's value estimate is noisy, and (2) either filter could remove potentially interesting chess positions. For example, requiring a small entropy and large value estimate difference (between moves) would result in predominantly tactical chess positions, thereby omitting interesting strategic puzzles.

\paragraph{AZ's calculations.}
In the second stage of the human experiment, we provide part of the MCTS statistics.
We ran MCTS without a depth limit for a maximum of $10,000$ simulations. 
We pruned the MCTS tree to avoid providing too many lines, or lines that were insufficiently explored. We provided the main line (most frequently visited), second and third moves, ranked according to visits. We did this for depth $t \leq 2$. 

\newpage 
\subsection{Extra results} \label{appx:more_results}

\subsubsection{Concept constraint satisfaction} \label{appx:res:cp}

\begin{longtable}[ht]{lrrrr}
\caption{Constraint Satisfaction Results for Piece Dataset}
\label{table:cp_pieces} \\ 
\toprule
Concept  &  layer 19 &  layer 20 &  layer 21 &  layer 23 \\
\midrule
\endhead
\midrule
\multicolumn{5}{r}{{Continued on next page}} \\
\midrule
\endfoot
\bottomrule
\endlastfoot
   rook &      1.0 &      1.0 &      1.0 &      1.0 \\
 knight &      1.0 &      1.0 &      1.0 &      1.0 \\
 bishop &      1.0 &      1.0 &      1.0 &      1.0 \\
  queen &      1.0 &      1.0 &      1.0 &      1.0 \\
   pawn &      1.0 &      1.0 &      1.0 &      1.0 \\
\end{longtable}

In the table below, the concepts are extracted from Stockfish 8's public API. Further details can be found in Appendix A of \cite{mcgrath2022acquisition}. 

\newpage 

\begin{longtable}{lcrrrr}
\caption{Concept constraint Satisfaction Results for Stockfish Dataset. w/b/t denotes White/Black/total difference; mg/eg/ph denotes middle game/ endgame/phased value, where the phased value is the weighted sum between the middle and endgame values, depending on the phase of the game.}
\label{table:cp_stockfish} \\ 
\toprule
Concept &  &  layer 19 &  layer 20 &  layer 21 &  layer 23 \\
\midrule
\endhead
\midrule
\multicolumn{5}{r}{{Continued on next page}} \\
\midrule
\endfoot
\bottomrule
\endlastfoot
bishop & [b,eg] &      1.0 &      1.0 &      1.0 &      1.0 \\
bishop & [b,mg] &      1.0 &      0.9 &      0.9 &      1.0 \\
bishop & [b,ph] &      0.9 &      0.9 &      0.9 &      1.0 \\
bishop & [t,eg] &      0.7 &      0.7 &      0.4 &      0.5 \\
bishop & [t,mg] &      0.6 &      0.7 &      0.8 &      0.6 \\
bishop & [t,ph] &      0.7 &      0.6 &      0.7 &      0.4 \\
bishop & [w,eg] &      1.0 &      1.0 &      0.9 &      1.0 \\
bishop & [w,mg] &      1.0 &      1.0 &      1.0 &      1.0 \\
bishop & [w,ph] &      1.0 &      1.0 &      0.5 &      1.0 \\
    imbalance & [t,eg] &      0.6 &      0.2 &      0.6 &      0.4 \\
    imbalance & [t,mg] &      0.6 &      0.2 &      0.6 &      0.4 \\
    imbalance & [t,ph] &      0.5 &      0.4 &      0.5 &      0.8 \\
  king safety & [b,eg] &      0.9 &      0.9 &      0.9 &      0.8 \\
  king safety & [b,mg] &      1.0 &      1.0 &      1.0 &      1.0 \\
  king safety & [b,ph] &      1.0 &      1.0 &      1.0 &      1.0 \\
  king safety & [t,eg] &      0.7 &      0.9 &      0.6 &      0.7 \\
  king safety & [t,mg] &      0.8 &      0.8 &      0.7 &      0.7 \\
  king safety & [t,ph] &      0.6 &      0.6 &      0.7 &      0.6 \\
  king safety & [w,eg] &      0.6 &      0.7 &      0.6 &      0.6 \\
  king safety & [w,mg] &      1.0 &      1.0 &      1.0 &      1.0 \\
  king safety & [w,ph] &      0.0 &      1.0 &      1.0 &      1.0 \\
      knights & [b,eg] &      0.0 &      0.0 &      0.9 &      0.0 \\
      knights & [b,mg] &      0.9 &      0.9 &      0.5 &      0.9 \\
      knights & [b,ph] &      0.0 &      0.0 &      1.0 &      0.8 \\
      knights & [t,eg] &      0.6 &      0.6 &      0.7 &      0.7 \\
      knights & [t,mg] &      0.9 &      0.5 &      0.9 &      0.9 \\
      knights & [t,ph] &      0.8 &      0.5 &      0.4 &      0.8 \\
      knights & [w,eg] &      0.9 &      0.8 &      0.8 &      0.9 \\
      knights & [w,mg] &      1.0 &      0.8 &      0.6 &      0.8 \\
      knights & [w,ph] &      0.8 &      0.6 &      0.3 &      0.8 \\
     material & [t,eg] &      0.8 &      0.7 &      0.3 &      0.7 \\
     material & [t,mg] &      0.6 &      0.8 &      0.5 &      0.6 \\
     material & [t,ph] &      0.6 &      0.6 &      0.6 &      0.7 \\
     mobility & [b,eg] &      1.0 &      1.0 &      1.0 &      1.0 \\
     mobility & [b,mg] &      1.0 &      1.0 &      1.0 &      1.0 \\
     mobility & [b,ph] &      1.0 &      1.0 &      1.0 &      1.0 \\
     mobility & [t,eg] &      0.8 &      0.9 &      0.8 &      0.7 \\
     mobility & [t,mg] &      0.7 &      0.4 &      0.6 &      0.6 \\
     mobility & [t,ph] &      0.8 &      0.7 &      0.7 &      0.8 \\
     mobility & [w,eg] &      1.0 &      1.0 &      1.0 &      1.0 \\
     mobility & [w,mg] &      1.0 &      1.0 &      1.0 &      1.0 \\
     mobility & [w,ph] &      1.0 &      1.0 &      1.0 &      1.0 \\
 passed pawns & [b,eg] &      0.8 &      0.8 &      0.9 &      0.8 \\
 passed pawns & [b,mg] &      0.8 &      0.9 &      0.7 &      0.7 \\
 passed pawns & [b,ph] &      0.9 &      1.0 &      0.9 &      0.9 \\
 passed pawns & [t,eg] &      0.8 &      0.9 &      0.7 &      0.7 \\
 passed pawns & [t,mg] &      0.6 &      0.7 &      0.3 &      0.8 \\
 passed pawns & [t,ph] &      0.6 &      0.5 &      0.6 &      0.7 \\
 passed pawns & [w,eg] &      1.0 &      1.0 &      1.0 &      1.0 \\
 passed pawns & [w,mg] &      0.6 &      0.8 &      0.6 &      0.7 \\
 passed pawns & [w,ph] &      1.0 &      1.0 &      0.9 &      0.9 \\
 pawns & [t,eg] &      0.6 &      0.5 &      0.4 &      0.5 \\
 pawns & [t,mg] &      0.3 &      0.7 &      0.5 &      0.4 \\
 pawns & [t,ph] &      0.8 &      0.7 &      0.7 &      0.7 \\
 phase &   &   1.0 &      1.0 &      1.0 &      1.0 \\
queens & [b,eg] &      0.3 &      0.3 &      0.4 &      0.3 \\
queens & [b,mg] &      0.9 &      0.9 &      0.8 &      0.8 \\
queens & [b,ph] &      0.9 &      0.9 &      0.7 &      0.9 \\
queens & [t,eg] &      0.3 &      0.3 &      0.4 &      0.3 \\
queens & [t,mg] &      0.6 &      0.4 &      0.2 &      0.7 \\
queens & [t,ph] &      0.6 &      0.5 &      0.6 &      0.6 \\
queens & [w,eg] &      0.3 &      0.3 &      0.4 &      0.3 \\
queens & [w,mg] &      0.8 &      0.7 &      0.6 &      0.9 \\
queens & [w,ph] &      0.9 &      0.8 &      0.7 &      0.7 \\
 rooks & [b,eg] &      1.0 &      0.9 &      1.0 &      0.9 \\
 rooks & [b,mg] &      1.0 &      1.0 &      0.9 &      1.0 \\
 rooks & [b,ph] &      0.9 &      1.0 &      1.0 &      0.9 \\
 rooks & [t,eg] &      0.6 &      0.5 &      0.3 &      0.7 \\
 rooks & [t,mg] &      0.7 &      0.8 &      0.7 &      0.7 \\
 rooks & [t,ph] &      0.7 &      0.5 &      0.5 &      0.8 \\
 rooks & [w,eg] &      1.0 &      1.0 &      0.9 &      1.0 \\
 rooks & [w,mg] &      1.0 &      1.0 &      1.0 &      1.0 \\
 rooks & [w,ph] &      0.9 &      0.9 &      0.8 &      0.9 \\
scale factor &   &   0.7 &      0.8 &      0.8 &      0.6 \\
 space & [b,eg] &      0.3 &      0.3 &      0.4 &      0.3 \\
 space & [b,mg] &      1.0 &      1.0 &      1.0 &      1.0 \\
 space & [b,ph] &      1.0 &      1.0 &      0.9 &      1.0 \\
 space & [t,eg] &      0.3 &      0.3 &      0.4 &      0.3 \\
 space & [t,mg] &      0.8 &      0.8 &      1.0 &      0.9 \\
 space & [t,ph] &      0.9 &      0.9 &      0.7 &      0.9 \\
 space & [w,eg] &      0.3 &      0.3 &      0.4 &      0.3 \\
 space & [w,mg] &      1.0 &      1.0 &      0.9 &      1.0 \\
 space & [w,ph] &      1.0 &      1.0 &      1.0 &      1.0 \\
threats & [b,eg] &      1.0 &      1.0 &      0.9 &      0.9 \\
threats & [b,mg] &      1.0 &      0.7 &      0.8 &      1.0 \\
threats & [b,ph] &      1.0 &      0.8 &      0.9 &      1.0 \\
threats & [t,eg] &      0.7 &      0.9 &      0.7 &      0.7 \\
threats & [t,mg] &      0.4 &      0.7 &      0.4 &      0.3 \\
threats & [t,ph] &      0.4 &      0.6 &      0.7 &      0.4 \\
threats & [w,eg] &      0.9 &      0.9 &      0.8 &      0.9 \\
threats & [w,mg] &      1.0 &      0.8 &      0.9 &      0.9 \\
threats & [w,ph] &      1.0 &      0.9 &      1.0 &      1.0 \\
total score &  &     0.7 &      0.8 &      0.4 &      0.6 \\
 total & [t,eg] &      0.5 &      0.7 &      0.5 &      0.6 \\
 total & [t,mg] &      0.8 &      0.6 &      0.6 &      0.8 \\
 total & [t,ph] &      0.8 &      0.7 &      0.9 &      0.8 \\
\end{longtable}

Further descriptions of the STS concepts can be found at \href{https://sites.google.com/site/strategictestsuite/about}{this website}.

\begin{longtable}{lrrrr}
\caption{Concept constraint Satisfaction Results for STS Dataset}
\label{table:sts} \\ 
\toprule
                 Concept  &  layer 19 &  layer 20 &  layer 21 &  layer 23 \\
\midrule
\endhead
\midrule
\multicolumn{5}{r}{{Continued on next page}} \\
\midrule
\endfoot
\bottomrule
\endlastfoot
               Undermine &      1.0 &      1.0 &      1.0 &      0.9 \\
              Open files &      1.0 &      1.0 &      1.0 &      1.0 \\
         Knight outposts &      1.0 &      1.0 &      1.0 &      0.9 \\
          Square vacancy &      1.0 &      1.0 &      1.0 &      1.0 \\
       Bishop vs. knight &      1.0 &      0.9 &      0.9 &      1.0 \\
               Recapture &      1.0 &      1.0 &      1.0 &      1.0 \\
 Offer of simplification &      0.9 &      1.0 &      1.0 &      0.8 \\
                fgh-pawn &      1.0 &      1.0 &      1.0 &      1.0 \\
                abc-pawn &      1.0 &      1.0 &      1.0 &      1.0 \\
          Simplification &      0.9 &      1.0 &      1.0 &      0.9 \\
           King activity &      1.0 &      1.0 &      1.0 &      1.0 \\
               Pawn push center &      1.0 &      1.0 &      1.0 &      1.0 \\
                7th rank &      1.0 &      1.0 &      1.0 &      1.0 \\
          Avoid exchange &      1.0 &      1.0 &      1.0 &      1.0 \\
\end{longtable}

\begin{longtable}{lrrrr}
\caption{Concept constraint Satisfaction Results for Opening Concept Configurations}
\label{table:opening_hyper} \\ 
\toprule
Hyperparamters &  layer 19 &  layer 20 &  layer 21 &  layer 23 \\
\midrule
\endhead
\midrule
\multicolumn{5}{r}{{Continued on next page}} \\
\midrule
\endfoot
\bottomrule
\endlastfoot
        single &     1.00 (0.00) &     1.00 (0.00) &     1.00 (0.00) &     1.00 (0.00) \\
   both (k=10) &     1.00 (0.00) &     0.99 (0.00) &     0.99 (0.01) &     1.00 (0.00) \\
    both (k=5) &     1.00 (0.00) &     1.00 (0.00) &     1.00 (0.00) &     1.00 (0.00) \\
\end{longtable}

\begin{longtable}{lrrrr}
\caption{Concep constraint Satisfaction Results for Opening Concepts}
\label{table:opening} \\ 
\toprule
      Concept &  layer 19 &  layer 20 &  layer 21 &  layer 23 \\
\midrule
\endhead
\midrule
\multicolumn{5}{r}{{Continued on next page}} \\
\midrule
\endfoot

\bottomrule
\endlastfoot
      English &      1.0 &      1.0 &      1.0 &      1.0 \\
        Dutch &      1.0 &      1.0 &      1.0 &      1.0 \\
 Scandinavian &      1.0 &      1.0 &      1.0 &      1.0 \\
     Sicilian &      1.0 &      1.0 &      1.0 &      1.0 \\
      Najdorf &      1.0 &      1.0 &      1.0 &      1.0 \\
       French &      1.0 &      1.0 &      1.0 &      1.0 \\
    Tarrasch &      1.0 &      1.0 &      1.0 &      1.0 \\
      Winawer &      1.0 &      1.0 &      1.0 &      1.0 \\
     Grünfeld &      1.0 &      1.0 &      1.0 &      1.0 \\
\end{longtable}

\subsection{Sample Efficiency Results} \label{appx:sample_efficiency}

\begin{figure}[ht] 
\centering
\caption{Sample efficiency of convex optimisation  framework across $10$ seeds for the bottleneck layer ($19$), value head ($20$) and policy head ($21, 23$).}
\includegraphics[width=0.8\textwidth]{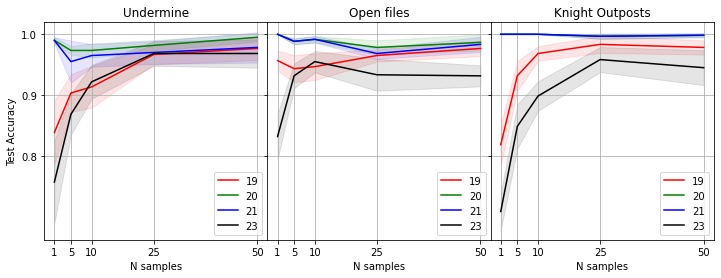}
\includegraphics[width=0.8\textwidth]{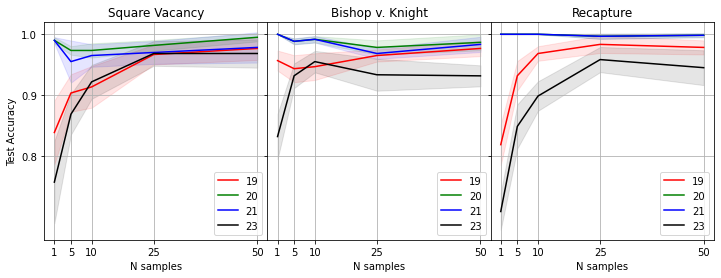}
\label{fig:sample_efficiency_dataset2}
\end{figure}

Figure~\ref{fig:sample_efficiency_dataset2} shows the result for sample efficiency experiment (\S\ref{res:sample_eff}) on the STS dataset.
As with the piece dataset, we find that the convex optimisation method reaches close to full-set accuracy with as little as $10$ or $25$ data points.
However, contrary to the piece dataset, we find that the performance is relatively similar across all layers, and slightly lower for the policy head (layer $23)$. 
One reason may be that these concepts are more important for the value than the policy. 

\end{document}